\documentclass[10pt,twocolumn,letterpaper]{article}

\usepackage[pagenumbers]{cvpr} 

\usepackage{graphicx}
\usepackage{multirow}
\usepackage{array} 
\usepackage{makecell}
\usepackage{adjustbox}
\usepackage{colortbl}
\usepackage{xcolor}
\usepackage{lipsum}
\usepackage{float}
\usepackage{pgf}
\usepackage{amsmath}
\usepackage{caption}
\usepackage{subcaption}
\usepackage{placeins}
\usepackage{tikz}

\usepackage{threeparttable}
\usepackage[table,xcdraw]{xcolor}
\definecolor{customRed}{RGB}{204, 0, 0}       
\definecolor{customPurple}{RGB}{153, 50, 204} 
\definecolor{customYellow}{RGB}{255, 223, 0}  
\definecolor{customLightYellow}{RGB}{255, 250, 205} 

\usepackage[dvipsnames]{xcolor}

\newcommand{\bc}{\mathbf{c}}
\newcommand{\bd}{\mathbf{d}}

\newcommand{\bp}{\mathbf{p}}

\newcommand{\figref}[1]{Figure~\ref{#1}}
\newcommand{\secref}[1]{Section~\ref{#1}}

\newcommand{\eqnref}[1]{Eq.~\ref{#1}}
\newcommand{\tabref}[1]{Table~\ref{#1}}

\makeatletter
\DeclareRobustCommand\onedot{\futurelet\@let@token\@onedot}
\def\@onedot{\ifx\@let@token.\else.\null\fi\xspace}
 
\def\ie{i.e\onedot}

\makeatother

\definecolor{yellow}{rgb}{1, 1, 0.7}
\definecolor{orange}{rgb}{1, 0.85, 0.7}
\definecolor{tablered}{rgb}{1, 0.7, 0.7}
\definecolor{red}{rgb}{1, 0, 0}

\definecolor{wincolor}{rgb}{0.85, 0.0, 0.0}

\definecolor{darkyellow}{rgb}{0.8, 0.8, 0.5}
\definecolor{darkred}{rgb}{0.7, 0.3, 0.3}
\definecolor{darkgreen}{rgb}{0.3, 0.7, 0.3}
\definecolor{blue}{rgb}{0.251, 0.498, 0.824}
\definecolor{green}{rgb}{0, 1.0, 0}
\definecolor{pink}{rgb}{1, 0.4, 0.7}

\definecolor{realred}{rgb}{0.95, 0.1, 0.0}

\newcommand{\boldparagraph}[1]{\vspace{0.2cm}\noindent{\bf #1:} }

\definecolor{tabfirst}{rgb}{1, 0.7, 0.7} 
\definecolor{tabsecond}{rgb}{1, 0.85, 0.7} 
\definecolor{tabthird}{rgb}{1, 1, 0.7} 
\definecolor{tabrealtime}{rgb}{0.8, 1, 0.8} 

\definecolor{red}{rgb}{1, 0.7, 0.7} 
\definecolor{blue}{rgb}{0.498, 0.749, 1} 

\newcommand{\beginsupplement}{%
    \setcounter{table}{0}
    \renewcommand{\thetable}{A\arabic{table}}%
    \setcounter{figure}{0}
    \renewcommand{\thefigure}{A\arabic{figure}}%
    \setcounter{section}{0}
    \renewcommand{\thesection}{A\arabic{section}}%
    \setcounter{equation}{0}%
    \renewcommand{\theequation}{A\arabic{equation}}%
}

\makeatletter
\pdfpxdimen=\dimexpr 1in/216\relax
\makeatother

\definecolor{cvprblue}{rgb}{0.21,0.49,0.74}
\usepackage[pagebackref,breaklinks,colorlinks,citecolor=cvprblue]{hyperref}

\def\paperID{355} 
\def\confName{3DV\xspace}
\def\confYear{2026\xspace}

\title{ConeGS: Error-Guided Densification Using Pixel Cones for Improved Reconstruction With Fewer Primitives}

\author{
Bartłomiej Baranowski
\quad
Stefano Esposito
\quad
Patricia Gschoßmann
\\
Anpei Chen$^{\dagger}$
\quad
Andreas Geiger
\\
University of Tübingen, Tübingen AI Center\\
{\tt\small \href{https://baranowskibrt.github.io/conegs/}{baranowskibrt.github.io/conegs}}
}

\begin{document}
\twocolumn[{
    \renewcommand\twocolumn[1][]{#1}%
    \maketitle
    \vspace{-6mm}
    \begin{minipage}{0.33\linewidth}
        \centering
        \resizebox{\linewidth}{!}{\input{images/teaser_jpg/teaser_psnr.pgf}\hspace{3mm}} 
        \includegraphics[width=0.94\linewidth]{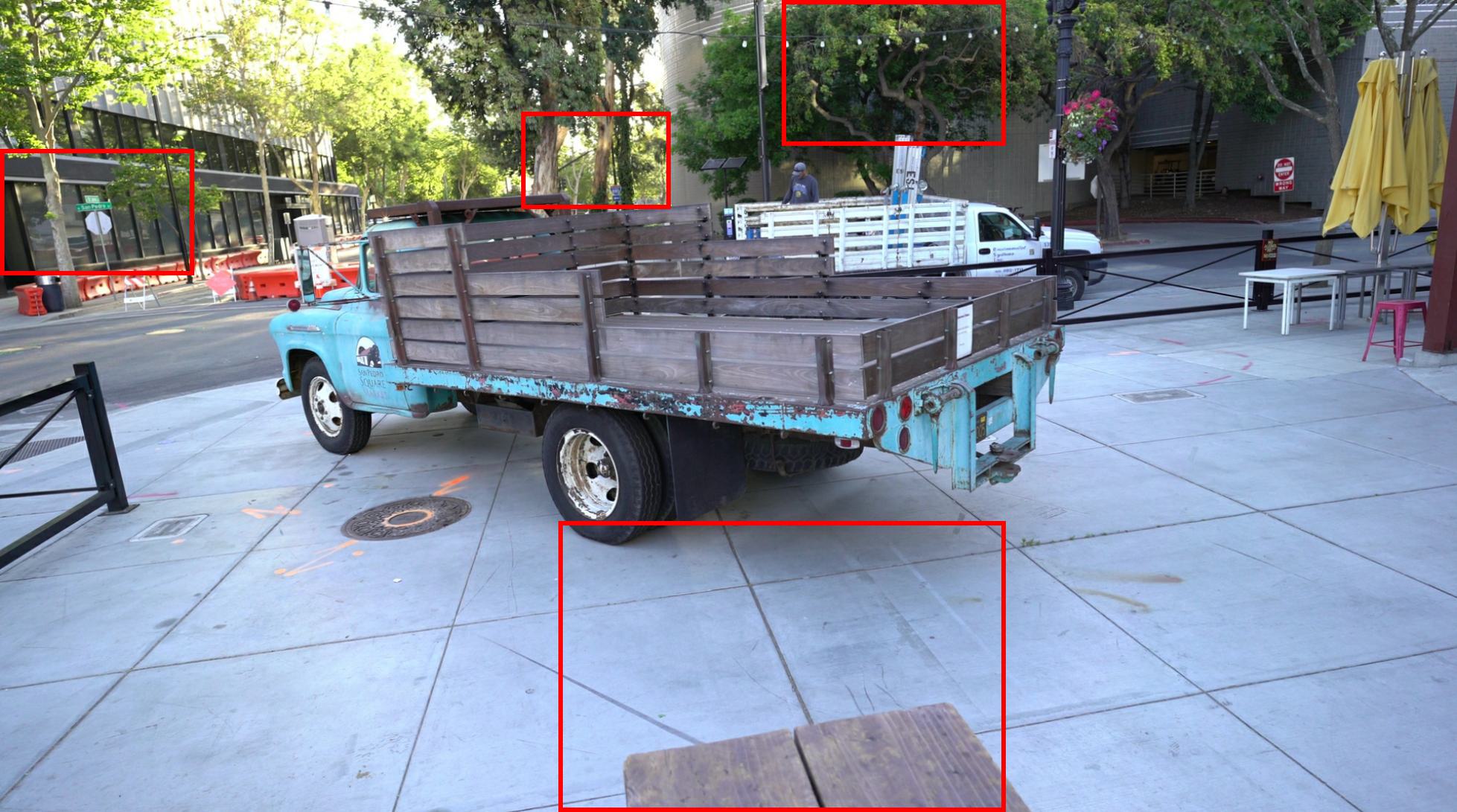}%
    \end{minipage}%
    \begin{minipage}{0.66\linewidth}
    \centering
    \begin{tabular}{@{}c@{\hspace{0.5em}}c@{\hspace{0.5em}}c@{\hspace{0.5em}}c@{}}
        \includegraphics[width=0.03\textwidth]{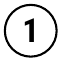} & \includegraphics[width=0.03\textwidth]{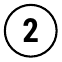} & \includegraphics[width=0.03\textwidth]{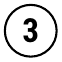} 
        \\
    \begin{minipage}[t]{0.31\linewidth}
        \centering
        \includegraphics[width=0.5\linewidth]{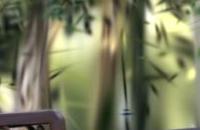}%
        \includegraphics[width=0.5\linewidth]{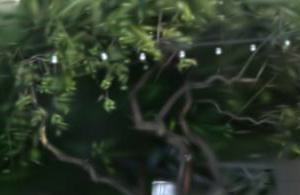}%
        \par\vspace{-2pt}
        \includegraphics[width=0.5\linewidth]{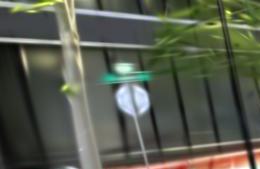}%
        \includegraphics[width=0.5\linewidth]{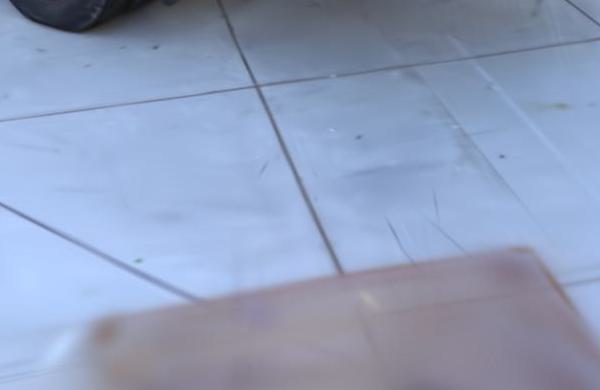}%
    \end{minipage} &
    \begin{minipage}[t]{0.31\linewidth}
        \centering
        \includegraphics[width=0.5\linewidth]{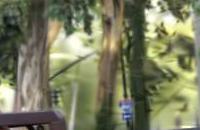}%
        \includegraphics[width=0.5\linewidth]{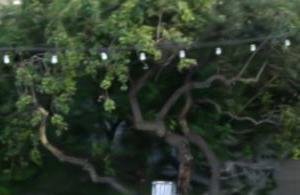}%
        \par\vspace{-2pt}
        \includegraphics[width=0.5\linewidth]{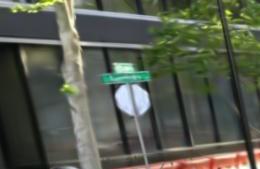}%
        \includegraphics[width=0.5\linewidth]{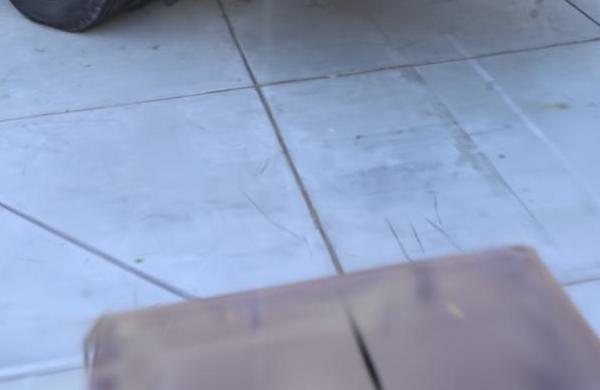}%
    \end{minipage} &
    \begin{minipage}[t]{0.31\linewidth}
        \centering
        \includegraphics[width=0.5\linewidth]{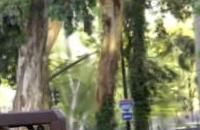}%
        \includegraphics[width=0.5\linewidth]{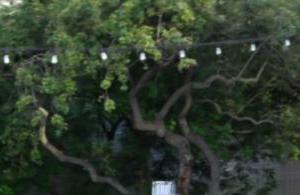}%
        \par\vspace{-2pt}
        \includegraphics[width=0.5\linewidth]{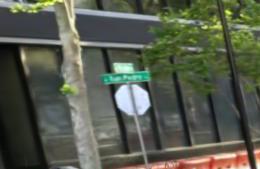}%
        \includegraphics[width=0.5\linewidth]{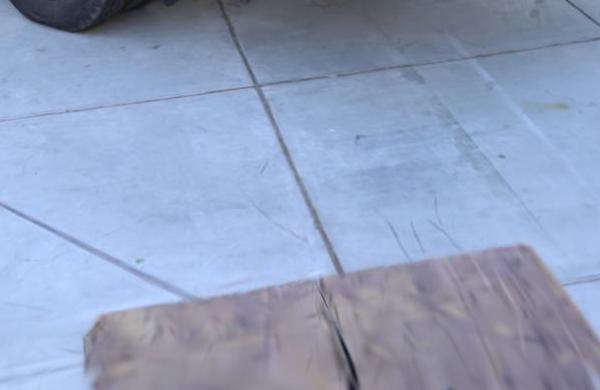}%
    \end{minipage} &
    \rotatebox[origin=c]{90}{\textsf{Ours}}
    \\[3.3em]
    
    \begin{minipage}[t]{0.31\linewidth}
        \centering
        \includegraphics[width=0.5\linewidth]{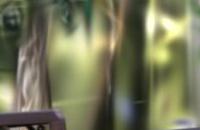}%
        \includegraphics[width=0.5\linewidth]{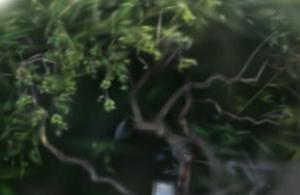}%
        \par\vspace{-2pt}
        \includegraphics[width=0.5\linewidth]{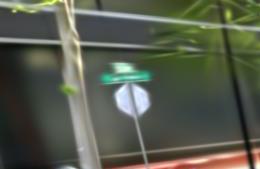}%
        \includegraphics[width=0.5\linewidth]{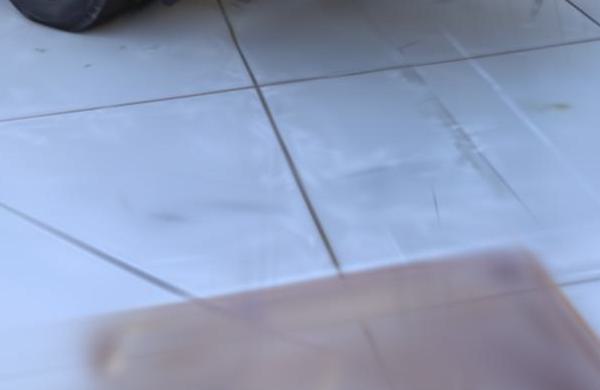}%
    \end{minipage} &
    \begin{minipage}[t]{0.31\linewidth}
        \centering
        \includegraphics[width=0.5\linewidth]{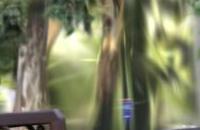}%
        \includegraphics[width=0.5\linewidth]{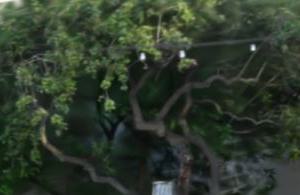}%
        \par\vspace{-2pt}
        \includegraphics[width=0.5\linewidth]{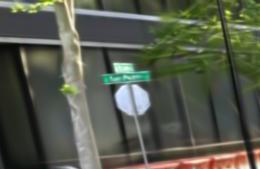}%
        \includegraphics[width=0.5\linewidth]{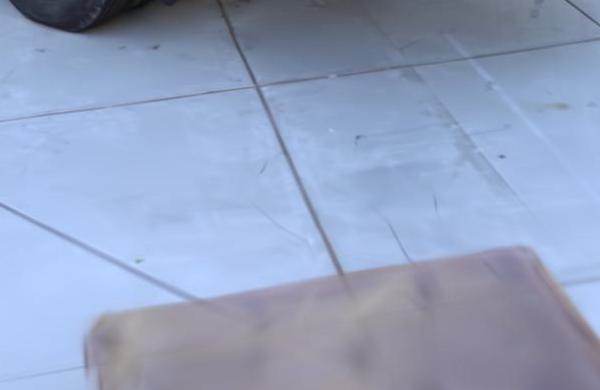}%
    \end{minipage} &
    \begin{minipage}[t]{0.31\linewidth}
        \centering
        \includegraphics[width=0.5\linewidth]{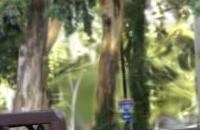}%
        \includegraphics[width=0.5\linewidth]{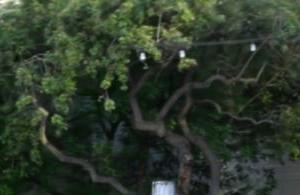}%
        \par\vspace{-2pt}
        \includegraphics[width=0.5\linewidth]{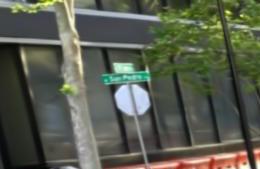}%
        \includegraphics[width=0.5\linewidth]{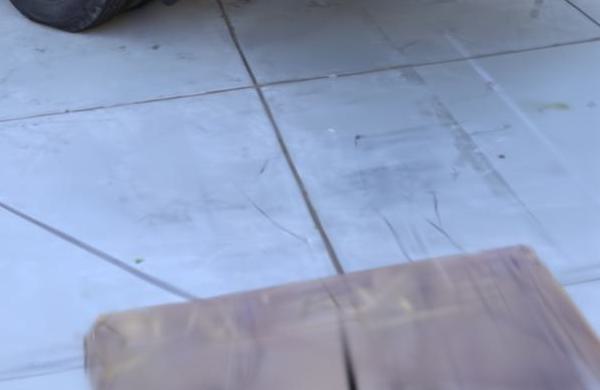}%
    \end{minipage} &
    \rotatebox[origin=c]{90}{\textsf{MCMC}~\cite{kheradmand20243d}}
    \\[3.3em]
    
    \begin{minipage}[t]{0.31\linewidth}
        \centering
        \includegraphics[width=0.5\linewidth]{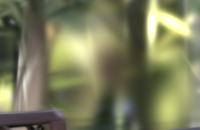}%
        \includegraphics[width=0.5\linewidth]{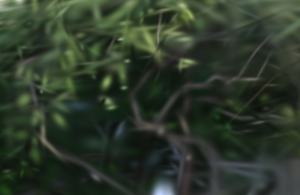}%
        \par\vspace{-2pt}
        \includegraphics[width=0.5\linewidth]{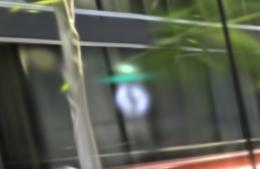}%
        \includegraphics[width=0.5\linewidth]{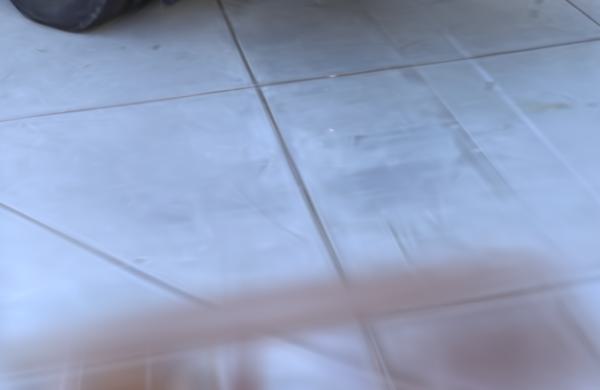}%
    \end{minipage} &
    \begin{minipage}[t]{0.31\linewidth}
        \centering
        \includegraphics[width=0.5\linewidth]{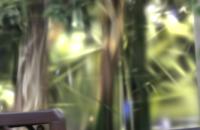}%
        \includegraphics[width=0.5\linewidth]{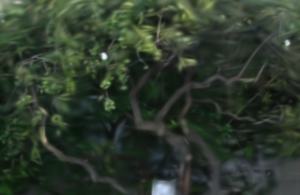}%
        \par\vspace{-2pt}
        \includegraphics[width=0.5\linewidth]{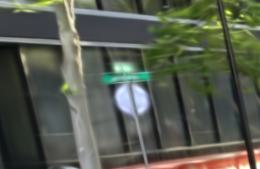}%
        \includegraphics[width=0.5\linewidth]{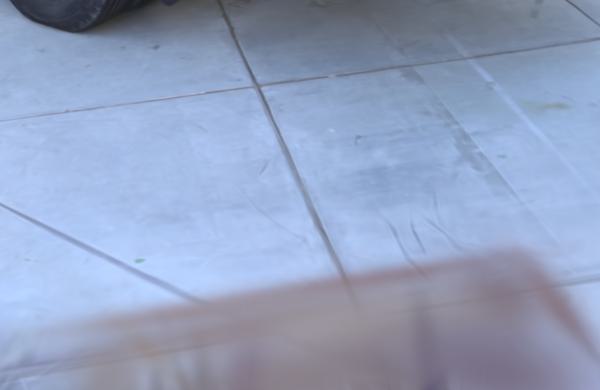}%
    \end{minipage} &
    \begin{minipage}[t]{0.31\linewidth}
        \centering
        \includegraphics[width=0.5\linewidth]{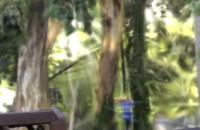}%
        \includegraphics[width=0.5\linewidth]{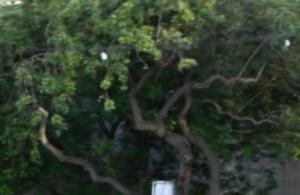}%
        \par\vspace{-2pt}
        \includegraphics[width=0.5\linewidth]{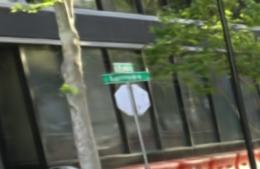}%
        \includegraphics[width=0.5\linewidth]{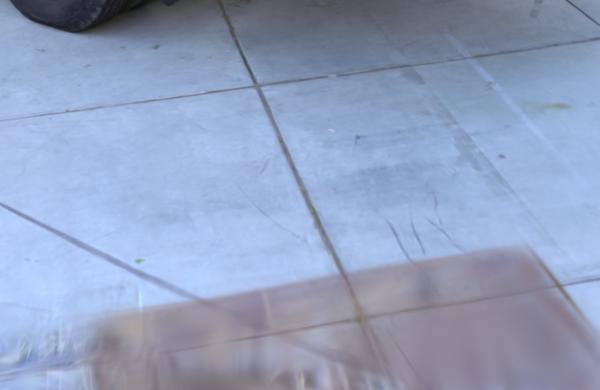}%
    \end{minipage} &

    \rotatebox[origin=c]{90}{\textsf{EDGS}~\cite{Kotovenko2025ARXIV_EDGS_Eliminating_Densification}}
    
    \end{tabular}
    
    \end{minipage}

    \captionof{figure}{%
    \textbf{ConeGS} replaces cloning-based densification with a novel method that generates pixel-cone-sized primitives in regions of high image-space error. By improving placement and removing reliance on existing scene structure thanks to a flexible iNGP-based exploration, it achieves higher reconstruction quality than baselines using the same number of primitives. Results are averaged over Mip-NeRF 360~\cite{barron2022mipnerf360} and OMMO~\cite{lu2023largescaleoutdoormultimodaldataset} datasets, with a visual comparison on the \texttt{truck} scene from Tanks \& Temples~\cite{Knapitsch2017}.
        \vspace{1.5em}
    }
    \label{fig:teaser}
}]

\renewcommand{\thefootnote}{\fnsymbol{footnote}}
\footnotetext[2]{ Corresponding author.}
\renewcommand{\thefootnote}{\arabic{footnote}}


\begin{abstract}

3D Gaussian Splatting (3DGS) achieves state-of-the-art image quality and real-time performance in novel view synthesis but often suffers from a suboptimal spatial distribution of primitives. This issue stems from cloning-based densification, which propagates Gaussians along existing geometry, limiting exploration and requiring many primitives to adequately cover the scene. We present \textit{ConeGS}, an image-space-informed densification framework that is independent of existing scene geometry state. ConeGS first creates a fast Instant Neural Graphics Primitives (iNGP) reconstruction as a geometric proxy to estimate per-pixel depth. During the subsequent 3DGS optimization, it identifies high-error pixels and inserts new Gaussians along the corresponding viewing cones at the predicted depth values, initializing their size according to the cone diameter. A pre-activation opacity penalty rapidly removes redundant Gaussians, while a primitive budgeting strategy controls the total number of primitives, either by a fixed budget or by adapting to scene complexity, ensuring high reconstruction quality. Experiments show that ConeGS consistently enhances reconstruction quality and rendering performance across Gaussian budgets, with especially strong gains under tight primitive constraints where efficient placement is crucial.

\end{abstract}
\section{Introduction}
\label{sec:introduction}

\begin{figure}
    \centering
    \includegraphics[width=1\columnwidth]{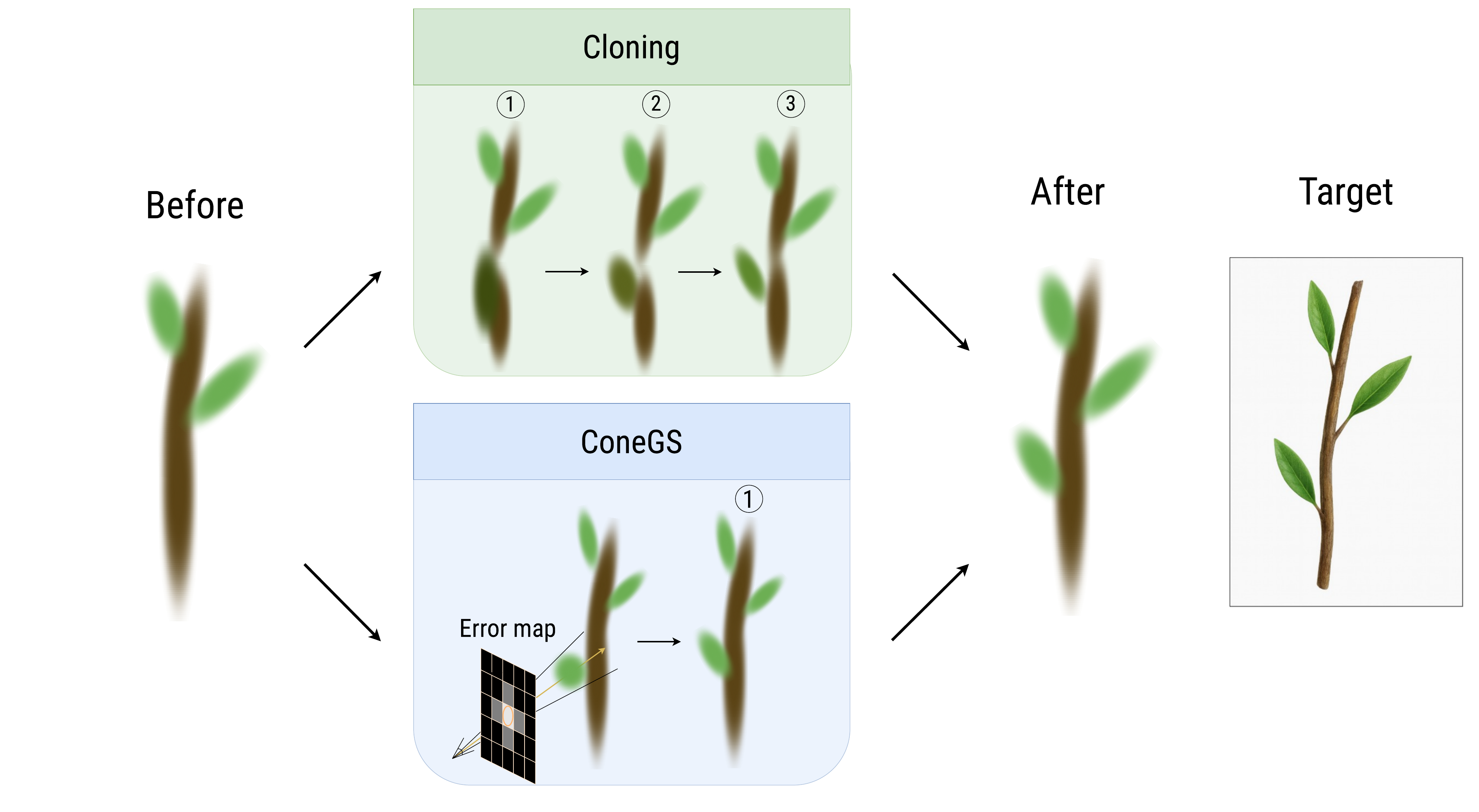}
    \caption{\textbf{Densification comparison.} Cloning-based methods are difficult to tune, and the resulting primitives may require many iterations to fit correctly into the scene. \textbf{ConeGS}, by contrast, places primitives precisely using the pixel viewing cone size, enabling faster scene integration without reliance on the existing geometry.}
    \label{fig:densification_comparison}
\end{figure}

Neural Radiance Fields (NeRF)~\cite{mildenhall2020nerf} have significantly advanced novel view synthesis, achieving remarkable fidelity in scene reconstruction. However, representing scenes with neural networks makes NeRF slow to train and render, though it provides smooth parameterization and flexibility to handle changes in scene structure. Recently, 3D Gaussian Splatting (3DGS)~\cite{kerbl3Dgaussians} has gained attention as a faster, more practical alternative to NeRF, explicitly modeling scenes with sets of 3D Gaussians to achieve interactive rendering speeds while maintaining competitive visual fidelity. 
However, 3DGS increases expressiveness through cloning and splitting, which offer limited exploration, rely on hard-to-define densification rules, and generate many unnecessary primitives. As a result, primitives often accumulate in suboptimal regions, leaving large parts of the scene underrepresented or mispredicted.

To address these issues, we propose \textbf{ConeGS}, which replaces cloning-based densification with a novel strategy that targets pixels exhibiting high photometric error.
By sampling these pixels and using depth estimates from a fast Instant Neural Graphics Primitives (iNGP)~\cite{mueller2022instant} reconstruction, new Gaussians are placed precisely in regions where the current representation is insufficient.
This targeted placement increases expressiveness in areas requiring higher primitive density, improving reconstruction quality while avoiding redundant primitives.
To determine the size of new Gaussians, we draw inspiration from Mip-NeRF~\cite{barron2021mipnerf}.
During densification, each Gaussian is initialized according to the size of the viewing cone of the pixel from which it is generated at the specified depth. 
Their initial size is thus defined directly by their image-space coverage, eliminating the need for local size analysis or adjustments to reconstructed regions.
\figref{fig:densification_comparison} illustrates the effectiveness of the proposed approach.
Combined with a pre-activation opacity penalty that quickly removes redundant Gaussians, this enables scene representation with fewer primitives while preserving high reconstruction quality.
We further incorporate two primitive budgeting strategies to regulate the total number of primitives, either through a fixed budget or by adapting to scene complexity.
ConeGS outperforms baseline methods across diverse datasets and a wide range of primitive budgets. The advantage is most pronounced under tight primitive budgets. At higher budgets, it matches the quality of cloning-based methods, where efficient primitive placement is less critical, while still rendering faster than the baselines.
In summary, our contributions are:
\begin{itemize}
    \item A densification strategy that places new Gaussians in regions of high photometric error in image space, guided by depth estimates from an iNGP-based geometric proxy.
    \item An approach that determines the size of new Gaussians from the viewing cones of the pixels from which they are generated.
    \item An improved opacity penalty that promptly removes low-opacity Gaussians, combined with a budgeting strategy that balances scene complexity and primitive count.
\end{itemize}

Finally, our method is also compatible with other 3DGS improvements, making it straightforward to integrate with existing approaches for greater efficiency, or with methods where cloning strategies are ambiguous or hard to formalize~\cite{scaffoldgs, Held20243DConvex, Ltzow2025LinPrimLP}. 

\section{Related work}

\boldparagraph{Neural Radiance Fields} 
NeRFs~\cite{mildenhall2020nerf} represent scenes as continuous volumetric radiance fields, enabling high-quality novel view synthesis. This is achieved by parameterizing the scene with a neural network (typically an MLP), whose weights encode the scene globally.
Despite producing photorealistic results, these methods rely on costly volumetric rendering and remain computationally inefficient.
Extensions such as Mip-NeRF~\cite{barron2021mipnerf} and Mip-NeRF360~\cite{barron2022mipnerf360} reduce aliasing via conical frustum integration, while Zip-NeRF~\cite{barron2023zipnerf} improves view consistency with hierarchical sampling and multi-scale supervision. 
Hybrid approaches~\cite{yu_and_fridovichkeil2021plenoxels,yu2021plenoctrees,SunSC22,Chen2022ECCV} mitigate this by combining explicit data structures with compact neural representations, enabling faster optimization and real-time rendering. 
Instant Neural Graphics Primitives (iNGP)~\cite{mueller2022instant} further accelerate training through multi-resolution hash-grid encoding and shallow MLPs. 

\boldparagraph{Primitive-based Differentiable Rendering} 
3D Gaussian Splatting (3DGS)~\cite{kerbl3Dgaussians} has emerged as an efficient alternative to Neural Radiance Fields (NeRF)~\cite{mildenhall2020nerf}. 
Rather than modeling the scene as a global volume, 3DGS represents it with local explicit 3D Gaussians and uses differentiable rasterization, resulting in significantly faster rendering.
Its balance of fidelity and efficiency has attracted significant attention and spurred a wide range of follow-up research.
Prior works have focused on tackling anti-aliasing~\cite{Yu2023MipSplattingA3,Yan2024CVPR}, reconstructing dynamic scenes~\cite{Wu_2024_CVPR,yang2023deformable3dgs}, enabling generative content creation~\cite{tang2023dreamgaussian,Zou2023}, reducing rendering artifacts~\cite{radl2024stopthepop}, substituting alpha composition with volumetric rendering~\cite{mai2024everexactvolumetricellipsoid,talegaonkar2025volumetricallyconsistent3dgaussian}, extracting geometry~\cite{Huang2DGS2024,Yu2024GOF,guedon2023sugar}, level-of-detail reconstruction~\cite{Ren2024OctreeGSTC},  frequency-based regularization~\cite{Zhang2024FreGS3G, Zeng2025FrequencyAwareDC}, and introducing new primitives or kernel functions~\cite{Hamdi_2024_CVPR,Held20243DConvex,liu2025deformablebetasplatting,Ltzow2025LinPrimLP}. 
Recent efforts have also targeted reducing computational and memory costs, often through feature quantization or code-book encoding~\cite{Niedermayr_2024_CVPR,girish2024eaglesefficientaccelerated3d,fan2023lightgaussian,scaffoldgs}, or scene simplification~\cite{Zhang2024GaussianSpaA}.
\cite{Fang2024MiniSplatting2B3} reduces computation by lowering the number of primitives through an aggressive densification and pruning strategy, while~\cite{fang2024minisplattingrepresentingscenesconstrained, Fang2024MiniSplatting2B3, Zhou2025PerceptualGSSP} insert new Gaussians at the currently estimated depths using re-initialization.
Unlike our method, this approach overwrites existing structures instead of adding new points, and further depends on the scene already being well reconstructed.
\cite{kheradmand20243d} improves primitive distribution and exploration by incorporating positional errors and applying penalties to opacity and scaling.
Closely related to our approach, several works focus on improving densification to reduce redundancy and better capture fine details.
Strategies include refining cloning heuristics~\cite{kheradmand20243d,Bul2024RevisingDI,huang2025decomposingdensificationgaussiansplatting}, per-Gaussian property- or saliency-based cues~\cite{10.1145/3680528.3687694}, geometry- and volume-aware criteria~\cite{jiang2024geotexdensifiergeometrytextureawaredensificationhighquality,refining_gaussian_splatting_2025, Zhou2025GradientDirectionAwareDC}, addressing gradient collision~\cite{ye2024absgs}, perceptual sensitivity~\cite{Zhou2025PerceptualGSSP}, learnable schemes~\cite{GenerativeDensification, Liu2025QuickSplatF3}, and based on Gaussian Processes~\cite{Guo2025GPGSGP}. Some works target densification in challenging settings~\cite{mohamad2024denser3dgaussianssplatting}, filling holes in the representation~\cite{Cheng2024ICML_GaussianPro_3D_Gaussian,li2025densesplatdensifyinggaussiansplatting}, though typically adding only a few primitives.  
PixelSplat~\cite{charatan23pixelsplat} models dense probability distributions for more robust Gaussian placement, influencing later approaches~\cite{chen2024mvsplat}. Recent work~\cite{Kotovenko2025ARXIV_EDGS_Eliminating_Densification} suggests that densification may be unnecessary for high-quality reconstruction given strong initialization. Like our method, they start by estimating scene geometry, but rely only on correspondences from a pretrained dense matching network, without enhancing densification, and at higher GPU memory cost than our approach.
Concurrent work \cite{zhang2025nest}, employs Gaussians with spatially varying texture colors, improving fine-detail reconstruction and reducing the number of primitives needed.
Other methods use neural radiance fields for depth supervision~\cite{Foroutan2024EvaluatingAT, li2024dngaussian} or point cloud extraction~\cite{niemeyer2024radsplat, Foroutan2024EvaluatingAT, wang2024pygslargescalescenerepresentation} to initialize a scene, but not to improve densification directly. Concurrent work~\cite{fang2025nerfvaluableassistant3d} applies NeRF for initialization and limited densification, constrained by existing Gaussian locations, and does not explore varying Gaussian sizes, which we find beneficial for reconstruction quality.
\section{Preliminaries}
\label{sec:preliminaries}

\boldparagraph{3D Gaussian Splatting}
\label{subsec:gaussian_splatting}
3DGS~\cite{kerbl3Dgaussians} represents a scene as an unordered set of 3D Gaussian primitives $\{\mathcal{G}_i | i = 1, \ldots, M\}$.
Each primitive $\mathcal{G}_i = (\mathbf{p}_i, \mathbf{s}_i, \mathbf{R}_i, o_i, \mathbf{c}_i)$ is defined by its position $\mathbf{p}_i\in\mathbb{R}^3$, scaling vector $\mathbf{s}_i\in\mathbb{R}^3$, rotation matrix $\mathbf{R}_i\in\mathbb{R}^{3\times3}$, opacity $o_i\in\mathbb{R}$, and view-dependent color $\mathbf{c}_i \in \mathbb{R}^3$.
The color $\mathbf{c}_i$ is represented by spherical harmonics (SH) coefficients $\mathbf{k}_i \in \mathbb{R}^{3L}$, where $L$ is the number of coefficients determined by the chosen SH order.
The 3D covariance matrix is given by $\mathbf{\Sigma}_i = \mathbf{R}_i\mathbf{S}_i\mathbf{S}_i^T\mathbf{R}_i^T$, where $\mathbf{S}_i = \text{diag}(\mathbf{s}_i)$ is the scaling matrix.
The color $\hat{C}$ of a pixel is computed by $\alpha$-blending over a set of $N$ Gaussians, sorted by depth, whose projections overlap the pixel:
\begin{align}
  \hat{C} &= \sum\nolimits_{i \in N} \mathbf{c}_i \alpha_i \prod\nolimits_{j=1}^{i-1}(1 - \alpha_j) ,\label{eq:3dgs_rendering} \\
  \alpha_i &= o_i K(\mathbf{p}_c, \boldsymbol{\mu}^{\text{2D}}_i, \mathbf{\Sigma}^{\text{2D}}_i) , \label{eq:3dgs_alpha}
\end{align}

where $\alpha_i$ is the blending weight of the $i$-th Gaussian, $\mathbf{p}_c$ is the pixel center in image coordinates, $\boldsymbol{\mu}_i^{\text{2D}}$ and $\mathbf{\Sigma}_i^{\text{2D}}$ are the 2D projected mean and covariance of $\mathcal{G}_i$, and $K(\cdot)$ is a Gaussian filter response in screen space.
The exact form of $K$ depends on the chosen filter~\cite{kerbl3Dgaussians, 964490, Yu2023MipSplattingA3}. 
Gaussians are traditionally initialized from an SfM point cloud, with each component of $\mathbf{s}_i$ set equal to the mean Euclidean distance to the three nearest neighbors $\mathcal{N}_3(i)$ of Gaussian $i$:
\begin{equation}
    \mathbf{s}_i = (s_{i}, s_{i}, s_{i}), \quad
    s_{i} = \frac{1}{3} \sum_{k \in \mathcal{N}_3(i)} \| \mathbf{p}_k - \mathbf{p}_i \|
    \quad .
    \label{eq:distance_initialization}
\end{equation}

During training, the Gaussian parameters are optimized with the loss:
\begin{equation}
    \mathcal{L}_{\text{GS}} = (1 - \lambda)\,\text{MAE}(I, I^*) + \lambda\,\mathcal{L}_{D\text{-}SSIM} , 
    \label{eq:3dgs_loss}
\end{equation}
where $\lambda=0.2$, $\text{MAE}$ is the mean absolute error between the rendered image $I$ and the ground-truth image $I^*$, and $\mathcal{L}_{D\text{-}SSIM} = 1 - \text{SSIM}(I, I^*)$~\cite{1284395}.

\boldparagraph{Neural Radiance Fields}
\label{subsec:neural_radiance_fields}
NeRFs~\cite{mildenhall2020nerf} model a scene as a continuous 3D field that maps a 3D location along a camera ray and the viewing direction of the corresponding pixel to a density $\sigma \in \mathbb{R}$ and color $\mathbf{c} \in \mathbb{R}^3$.
A camera ray is parameterized as $\mathbf{r}(t) = \bp_{\text{cam}} + t \bd$, where $\bp_{\text{cam}}$ is the camera position and $\bd$ is a unit direction vector pointing toward the center of a pixel. 
Each ray is discretized into $N$ intervals defined by distances $\{t_i, t_{i+1}\}_{i=1}^N$.
For each sample position $\mathbf{r}(t_i)$ along the ray, the NeRF is queried to predict the sample's color $\bc_i$ and density $\sigma_i$.
Using volumetric rendering~\cite{mildenhall2020nerf}, the corresponding pixel color is approximated as:

\begin{align}
  \hat{C} &= \sum\nolimits_{i=1}^N \mathbf{c}_i \alpha_i \prod\nolimits_{j=1}^{i-1}(1 - \alpha_j)\label{eq:nerf_rendering} , \\
  \alpha_i &= 1 - \exp(-\sigma_i \delta_i) \quad \text{with} \enspace \delta_i = t_{i+1} - t_i .
  \label{eq:opacity_formula}
\end{align}

Here, $\alpha_i$ is the opacity of the $i$-th sample, $\delta_i$ is the length of its ray segment, and the product term represents the transmittance $\tau_i = \prod_{j=1}^{i-1}(1 - \alpha_j)$.

Sampling only a single ray per pixel can lead to blur and aliasing.
Mip-NeRF~\cite{barron2021mipnerf} addresses this by replacing the ray with a cone that models the pixel footprint, \ie the 3D volume a pixel covers in world space. 
The cone is divided into frustums, and integration is performed over these volumes rather than along a 1D line.
The cone's radius $r_{\mathrm{cone}}(t)$ defines the cross-section of the pixel cone at distance $t$ and is computed from the directions of rays passing through the pixel and its neighbors:

\begin{equation}
\label{eq:pixel_footprint_radius}
r_{\mathrm{cone}}(t) = t \; \frac{\| \mathbf{d}_x - \mathbf{d} \| + \| \mathbf{d}_y - \mathbf{d} \|}{2} ,
\end{equation}

where $\mathbf{d}$ is the direction of the ray through the center of the pixel, and $\mathbf{d}_x, \mathbf{d}_y$ are the directions of rays through the neighboring pixels in the $x$ and $y$ directions, respectively.

\begin{figure*}[t]
    \centering
    \renewcommand{\arraystretch}{0.5} 
    \setlength{\tabcolsep}{1pt}     
    \includegraphics[width=0.9\linewidth, keepaspectratio,
                   trim={20pt 10pt 20pt 0pt}, clip]%
                  {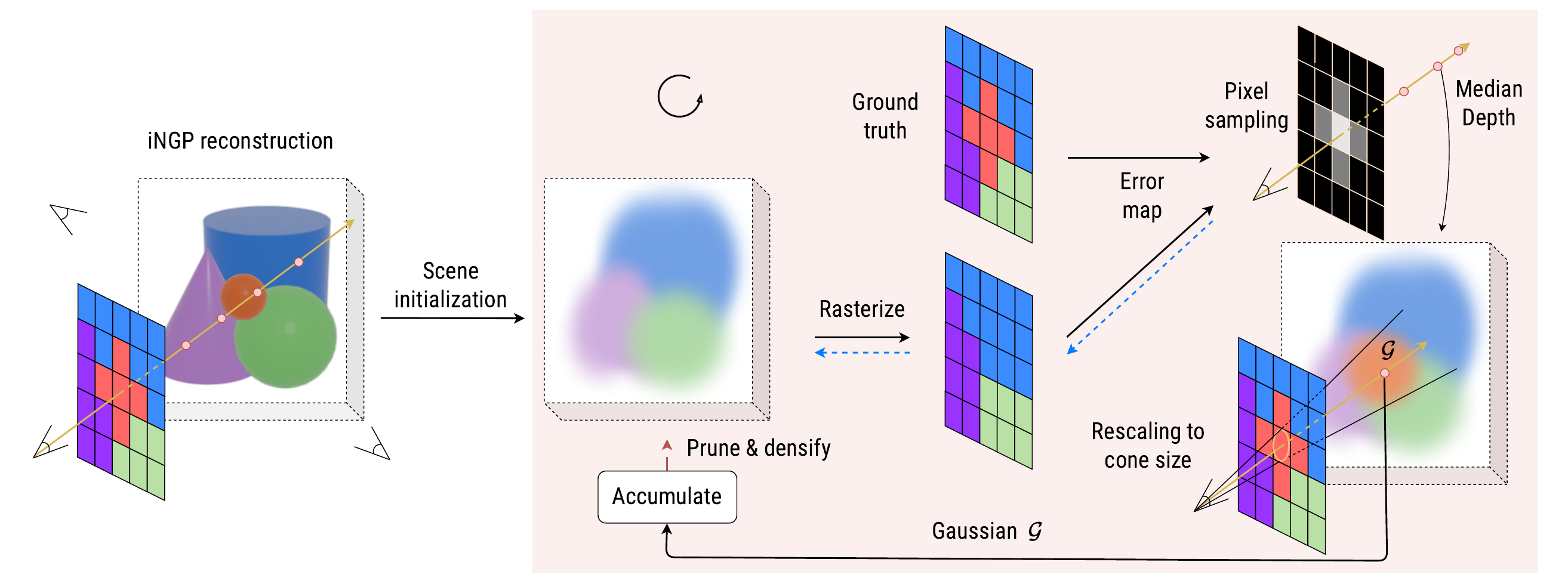}
  \hspace*{-2.6cm}
  \subcaptionbox{Initialization (\secref{subsec:initialization})%
                 \label{fig:pipeline:a}}[0.48\linewidth]{}
  \subcaptionbox{Optimization (\secref{subsec:optimization})%
                 \label{fig:pipeline:b}}[0.48\linewidth]{}
    
    \caption{
    \textbf{Overview of the ConeGS pipeline.} 
($a$) First, an iNGP reconstruction is obtained to serve as a geometric proxy for object surfaces, guiding the placement of Gaussians both during scene initialization and throughout the 3DGS optimization process.
($b$) During 3DGS optimization, ConeGS performs error-guided densification by sampling a subset of pixels with high $L_1$ error.
For each sampled pixel, a new Gaussian $\mathcal{G}$ is created along the pixel's viewing cone at the depth estimated by iNGP and scaled to match the cone's size. 
New Gaussians are accumulated and, every 100 iterations, inserted into the scene after pruning those with low opacity.
Blue arrows indicate \textcolor{blue}{gradient updates} to Gaussian parameters, and the red arrow marks \textcolor{red}{scene updates}.
}
    \label{fig:pipeline}
\end{figure*}

\section{Method}
\label{sec:method}

This section outlines our ray-based densification approach for 3DGS.
First, we explain how we use an iNGP model as a geometric proxy to initialize the 3D Gaussian scene (\secref{subsec:initialization}).
Next, we detail our ray-based densification strategy, which uses the iNGP to place pixel-cone-sized Gaussians in high-error regions, along with associated optimization changes (\secref{subsec:optimization}).
Finally, we provide additional implementation details in \secref{subsec:implementation_details}.
An overview of the complete pipeline is shown in \figref{fig:pipeline}.

\subsection{Initialization}
\label{subsec:initialization}

We use a trained iNGP model~\cite{mueller2022instant} as a geometric proxy to initialize the 3DGS scene and guide densification. Trained briefly on input images, it provides accurate depth estimates, that position both the initial Gaussian primitives and those added later during densification, with minimal impact on training time. Additionally, the depths can be evaluated on the fly during optimization, reducing both memory usage and computation compared to precomputing all depth maps.
We initialize the scene with $\mathcal{P}_{\text{init}}$ Gaussians, set to one million as in~\cite{niemeyer2024radsplat}, or fewer if a smaller budget is specified (see~\secref{subsec:primitive_budgeting}).
To construct this set, we uniformly sample $\mathcal{P}_{\text{init}}$ image-pixel pairs $(I, u, v)$ from the training set pixel domain \(\mathcal{I}_{\text{train}}\).
Each sampled image-pixel pair defines exactly one Gaussian in the initialized scene.
For each sample, we define its associated camera ray
\begin{equation}
    \mathbf{r}_I(u, v, t) = \mathbf{p}_I + t \, \mathbf{d}_I(u, v) , 
\end{equation}
where \(I\) is an image index,  \((u, v)\) are pixel coordinates, \(\mathbf{p}_I \in \mathbb{R}^3\) is the camera center, and \(\mathbf{d}_I(u, v) \in \mathbb{R}^3\) is the normalized ray direction. 
We query the iNGP along this ray to obtain discrete transmittance values $\{\tau_k\}$, from which the median depth $t_{\text{med}}$ is computed as:
\begin{equation}
    t_{\text{med}} = t_k \quad \text{where} \quad \tau_{k-1} > 0.5 \geq \tau_k .
\end{equation}
The center of the $j$-th Gaussian primitive is then set to:
\begin{equation}
    \mathbf{p}_j = \mathbf{p}_I + t_{\text{med}} \, \mathbf{d}_I(u, v),
\end{equation}
yielding the set of initial centers $\{\mathbf{p}_j\}_{j \in \mathcal{I}{\text{sample}}}$ with $\mathcal{I}_{\text{sample}} \subset \mathcal{I}_{\text{train}}$.
The scale $\mathbf{s}_j$ is initialized isotropically using the average distance to the three nearest neighbors $\mathcal{N}_3(j)$, following Eq.~\eqref{eq:distance_initialization}.
The rotation is set to identity $\mathbf{R}_j = \mathbf{I}$, the opacity to $o_j = 0.1$, and the SH coefficients to:
\begin{equation}
    \mathbf{k}_j = \left( \mathbf{k}_{1:3, j}, \mathbf{k}_{4:L, j} \right), \quad
\mathbf{k}_{1:3, j} = \mathbf{c}_{j, 0}, \quad
\mathbf{k}_{4:L, j} = \mathbf{0}, 
\end{equation}
where $\mathbf{c}_{j,0}$ is the RGB color for the sampled pixel $(I, u, v)$ rendered with iNGP using a zeroed-out view direction.
Although our densification strategy can achieve high-quality reconstructions without scene initialization, we retain this step to ensure consistently strong performance across all metrics (see \secref{subsec:ablations}).

\subsection{Optimization}
\label{subsec:optimization}
We fully replace the standard 3DGS cloning-based densification with an error-guided strategy that adapts the iNGP-based ray-depth rendering procedure from \secref{subsec:initialization} to position new Gaussian primitives.
Below, we outline the sampling, scaling, budgeting, and pruning stages of our densification pipeline.

\boldparagraph{Error-Weighted Gaussian Densification}
To limit the number of primitives while targeting poorly reconstructed regions, we add new Gaussians at pixels with high photometric error.
At iteration \(j\), we render an image \(I_j\) and compute the per-pixel absolute error ($L_1$ loss) \(E(\mathbf{p}) = |I_j(\mathbf{p}) - I^*(\mathbf{p})|\) with respect to the ground-truth image \(I^*\). 
We then sample $N_{\text{sample}}$ pixels without replacement according to a multinomial distribution $\mathcal{M}$ with probabilities proportional to the normalized error map:

\begin{equation}
\{\mathbf{p}_s\}_{s=1}^{N_{\text{sample}}} \sim \mathcal{M} \left(N_{\text{sample}}, \frac{E(\mathbf{p})}{\sum_{\mathbf{p}' \in I_j} E(\mathbf{p}')}\right) , 
\end{equation}

While $L_1$ loss does not always indicate possible improvements and can also arise from noise or difficult-to-optimize reflections, we found it to be a reliable indication of lacking expressiveness, especially at low primitive budgets. For each sampled pixel \(\mathbf{p}_s\), a new Gaussian \(\mathcal{G}_s\) is created, with its center placed along the corresponding ray at the median depth $t_{\text{med}, s}$ given by the iNGP. 
Newly spawned Gaussians are appended to an accumulation set:

\begin{equation}
    \mathcal{G}_{\text{accum}} \gets \mathcal{G}_{\text{accum}} \cup \left\{ \mathcal{G}_s \right\}_{s=1}^{N_{\text{sample}}}    .
\end{equation}

Every 100 iterations, low-opacity Gaussians in the scene \(\mathcal{G}_{\text{scene}}\) are pruned, and Gaussians in $\mathcal{G}_{\text{accum}}$ are merged into the scene:

\begin{equation}
\mathcal{G}_{\text{scene}} \gets \mathcal{G}_{\text{scene}} \cup \mathcal{G}_{\text{accum}}, \quad \mathcal{G}_{\text{accum}} \gets \emptyset .
\end{equation}

\begin{figure}
    \centering
    \includegraphics[width=1\columnwidth]{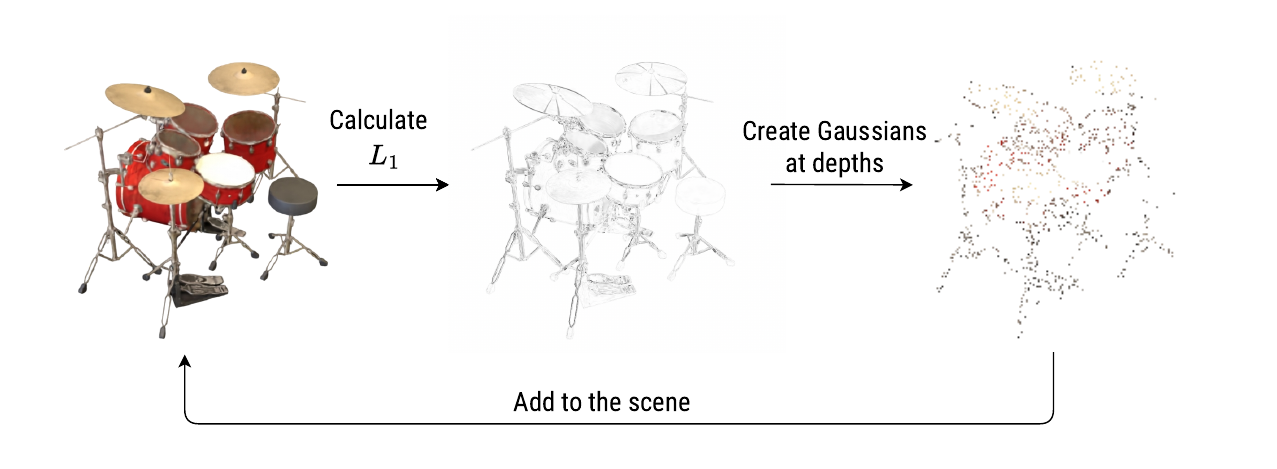}
    \caption{\textbf{Densification overview.} 
    Illustration of the proposed error-guided strategy.
    We render an image with 3DGS, compute the per-pixel $L_1$ error, sample pixels proportionally to their error magnitude, and place new Gaussians at the iNGP-predicted depth along the corresponding viewing rays.
    }
    \label{fig:ray_densification}
\end{figure}

The newly inserted Gaussians are then jointly optimized with existing primitives.
Unlike cloning-based methods, which constrain new primitives to the vicinity of existing ones and thus hinder exploration of unseen regions, our approach places Gaussians directly in high-error areas, enabling effective scene coverage even far from existing geometry.
Moreover, we preserve the integrity of well-reconstructed areas, since new Gaussians are added on top of the existing structure rather than created through splitting or cloning.
Although occasional iNGP depth inaccuracies may introduce misplaced Gaussians, diverse viewpoint coverage ensures that inconsistent ones are quickly corrected or pruned, while multiview-consistent ones are retained.
The full densification pipeline is shown in \figref{fig:ray_densification}.

\boldparagraph{Pixel-Footprint-Aligned Scaling}
Selecting an appropriate scale for newly added Gaussians during densification is crucial.
If primitives are too large, they may obscure fine details and be pruned prematurely, whereas overly small ones contribute little to the rendered image, yielding weak gradients and slowing convergence.
Although $k$-NN–based scaling is effective for initialization, recomputing nearest-neighbor distances at every densification step is computationally expensive and sensitive to outliers. Large distances can produce inflated scales, causing new Gaussians to overlap well-reconstructed regions and hinder further optimization (see \secref{subsec:ablations}).
To avoid these issues, we set the initial scale of each newly added Gaussian directly from the pixel footprint at the median depth $t_{\text{med}, i}$ along its corresponding camera ray, as defined in Eq.~\ref{eq:pixel_footprint_radius}:
\begin{equation}
\label{eq:gaussian_scale_isotropic}
\mathbf{s}_i = \lambda_{\text{scale}} \, r_{\mathrm{cone}}(t_{\mathrm{med}, i}) \, (1, 1, 1) , 
\end{equation}
where \(\lambda_{\text{scale}}=2\) converts the cone radius \(r_{\mathrm{cone}}(t_{\mathrm{med}, i})\) to the diameter of its cross-section.
This ensures that, from the spawning viewpoint, the Gaussian’s projection onto the image plane approximately matches the pixel width, independent of scene depth.
The assigned scale is only an initial value, with subsequent optimization steps jointly updating all primitives to allow newly added Gaussians to adjust to the existing scene.
Our pixel-aligned, depth-aware scaling provides three key benefits:
(1) it is independent of the current primitive distribution, avoiding the structural biases of cloning-based methods that replicate and reinforce local geometry,
(2) pixel size allows Gaussians to contribute to optimization immediately and efficiently fit fine details while minimizing overlap with existing structure, and
(3) its isotropic shape promotes stable multi-view integration, without shapes that the cloned Gaussians inherit.

\boldparagraph{Primitive Budgeting}
\label{subsec:primitive_budgeting}
We consider two budgeting strategies for controlling the number of Gaussians in the scene.
The first enforces a hard upper bound, as in~\cite{kheradmand20243d}, ensuring that densification never exceeds the prescribed budget.
This constraint regulates memory and computation while preventing uncontrolled growth of the primitive set.
At each densification step, we set the number of sampled pixels \(N_{\text{sample}}\) so that newly added Gaussians replace those pruned, avoiding excess primitives that would otherwise be discarded under the budget.
This is computed as:
\begin{equation}
    N_{\text{sample}} = \frac{\max(0.2 N_{\text{GS}},\ 1.2 N_{\text{last}})}{100} , 
    \label{eq:accumulated_budget}
\end{equation}
where \(N_{\text{GS}}\) is the current total number of Gaussians, and \(N_{\text{last}}\) is the number of primitives inserted in the previous densification step, and the division by 100 reflects the densification interval.
This formulation keeps $N_{\text{GS}}$ close to the budget limit even under aggressive pruning, maintaining consistent scene coverage throughout optimization.

The second strategy adapts the number of primitives to the scene's complexity, enabling controlled growth without imposing a fixed upper bound:
\begin{equation}
    N_{\text{sample}} = \frac{\beta N_{\text{GS}}}{100} .
    \label{eq:accumulated_no_budget}
\end{equation}
Here, $\beta$ controls the growth rate of the primitive set.
Smaller values balance the number of Gaussians added with those pruned, maintaining a relatively stable primitive count, whereas larger values yield higher primitive counts, increasing geometric detail at the cost of memory and computing power. 
With scene initialization at 1M primitives and the application of the opacity penalty, we balance Gaussians added and pruned, unlike~\cite{Bul2024RevisingDI}, which requires a predefined upper limit on primitives.

\boldparagraph{Opacity-Regularized Pruning}
Following~\cite{kerbl3Dgaussians, kheradmand20243d}, Gaussians with opacity below $0.005$ are pruned every 100 iterations to remove primitives with negligible contribution to the rendered image.
Earlier work has promoted sparsity through different strategies: periodically resetting opacities~\cite{kerbl3Dgaussians}, which can destabilize training~\cite{Bul2024RevisingDI}, reducing opacities by a constant amount after each densification~\cite{Bul2024RevisingDI}, or introducing a post-activation opacity penalty 
\(\mathcal{L}_{\mathbf{o}}^{\text{post}} = \|\sigma(\mathbf{o}_{\text{pre}})\|_{1}\)~\cite{kheradmand20243d}, where \(\mathbf{o}_{\text{pre}}\) denotes the opacity logits before the sigmoid and \(\sigma\) is the sigmoid function. This applies the strongest constraint around \(0.5\) and only a weak penalty near the pruning threshold.
In contrast, we employ a pre-activation opacity penalty \(\mathcal{L}_{\mathbf{o}}^{\text{pre}} = \|\mathbf{o}_{\text{pre}}\|_{1}\). It provides a steady constraint across the full opacity range, including very low values, gradually reducing under-contributing primitives. The penalty acts throughout training, and our densification strategy can freely add new primitives, allowing any structure lost through pruning to be recovered more easily than with cloning-based approaches. In all experiments, this loss is scaled by \(\lambda_{\mathbf{o}} = 0.0002\).

\subsection{Implementation Details}
\label{subsec:implementation_details}
For the iNGP model, we use the proposal-based implementation from NerfAcc~\cite{li2023nerfacc}, trained for 20k iterations with the original setup and architecture. Gaussian optimization runs for 30k iterations, with our densification active for the first 25k. Unlike 3DGS, all SH components are optimized from the start, enabled by the stable initialization that removes the need for gradual SH introduction.

%
%
\begin{table*}[!t]
\centering
\resizebox{0.95\linewidth}{!}{%
\begin{threeparttable}
\begin{tabular}{c|ccc|ccc|ccc|ccc}
\toprule
 & \multicolumn{3}{c|}{Mip-NeRF360 \cite{barron2022mipnerf360}} & \multicolumn{3}{c|}{OMMO \cite{lu2023largescaleoutdoormultimodaldataset}} & \multicolumn{3}{c|}{Tanks \& Temples \cite{Knapitsch2017}} & \multicolumn{3}{c}{DeepBlending \cite{DeepBlending2018}} \\
 & \textbf{PSNR $\uparrow$} & \textbf{SSIM $\uparrow$} & \textbf{LPIPS $\downarrow$} & \textbf{PSNR $\uparrow$} & \textbf{SSIM $\uparrow$} & \textbf{LPIPS $\downarrow$} & \textbf{PSNR $\uparrow$} & \textbf{SSIM $\uparrow$} & \textbf{LPIPS $\downarrow$} & \textbf{PSNR $\uparrow$} & \textbf{SSIM $\uparrow$} & \textbf{LPIPS $\downarrow$} \\
\midrule
\multicolumn{13}{c}{Number of Gaussians limited to $100$k} \\
\midrule
3DGS~\cite{kerbl3Dgaussians} (SfM init.) & 23.61 & 0.693 & 0.413 & 26.45 & 0.820 & 0.296 & 22.38 & 0.774 & 0.333 & 24.65 & 0.827 & 0.412 \\
Foroutan et al.~\cite{Foroutan2024EvaluatingAT}$^\dagger$ & 26.64 & 0.781 & 0.318 & 26.89 & 0.829 & 0.276 & 22.48 & 0.766 & 0.341 & 25.31 & 0.822 & 0.418 \\
MCMC~\cite{kheradmand20243d} (rand. init.) & 25.72 & 0.730 & 0.369 & 25.92 & 0.808 & 0.313 & 21.45 & 0.750 & 0.365 & 27.94 & 0.859 & 0.369 \\
MCMC~\cite{kheradmand20243d} (SfM init.) & 27.06 & \colorbox{tabsecond}{0.800} & 0.303 & \colorbox{tabsecond}{27.01} & \colorbox{tabsecond}{0.841} & 0.266 & \colorbox{tabthird}{22.50} & \colorbox{tabsecond}{0.780} & 0.332 & \colorbox{tabthird}{28.94} & \colorbox{tabsecond}{0.876} & \colorbox{tabsecond}{0.333}  \\
MCMC~\cite{kheradmand20243d} (iNGP init.) & \colorbox{tabsecond}{27.35} & 0.797 & \colorbox{tabthird}{0.299} & 26.95 & 0.837 & \colorbox{tabthird}{0.265} & \colorbox{tabsecond}{22.69} & 0.775 & \colorbox{tabthird}{0.326} & \colorbox{tabsecond}{29.02} & 0.872 & \colorbox{tabthird}{0.337} \\
GaussianPro~\cite{Cheng2024ICML_GaussianPro_3D_Gaussian} & 25.57 & 0.766 & 0.338 & 26.14 & 0.822 & 0.289 & 20.59 & 0.757 & 0.348 & 28.15 & 0.870 & 0.342 \\
Perceptual-GS~\cite{Zhou2025PerceptualGSSP} & 25.66 & 0.774 & 0.320 & 26.13 & 0.819 & 0.292 & 20.76 & 0.759 & 0.347 & 28.32 & \colorbox{tabthird}{0.874} & 0.338 \\
EDGS~\cite{Kotovenko2025ARXIV_EDGS_Eliminating_Densification} & \colorbox{tabthird}{27.09} & \colorbox{tabthird}{0.798} & \colorbox{tabsecond}{0.296} & \colorbox{tabthird}{26.99} & \colorbox{tabthird}{0.838} & \colorbox{tabsecond}{0.261} & 22.32 & \colorbox{tabthird}{0.777} & \colorbox{tabsecond}{0.324} & 28.43 & 0.872 & \colorbox{tabthird}{0.337} \\
Ours & \colorbox{tabfirst}{27.74} & \colorbox{tabfirst}{0.809} & \colorbox{tabfirst}{0.285} & \colorbox{tabfirst}{27.59} & \colorbox{tabfirst}{0.852} & \colorbox{tabfirst}{0.243} & \colorbox{tabfirst}{23.12} & \colorbox{tabfirst}{0.791} & \colorbox{tabfirst}{0.310} & \colorbox{tabfirst}{29.44} & \colorbox{tabfirst}{0.880} & \colorbox{tabfirst}{0.328} \\

\midrule
\multicolumn{13}{c}{Number of Gaussians limited to $500$k} \\
\midrule
3DGS~\cite{kerbl3Dgaussians} (SfM init.) & 28.22 & 0.821 & 0.260 & \colorbox{tabthird}{28.96} & 0.883 & 0.196 & 23.54 & 0.816 & 0.265 & 29.25 & 0.882 & 0.302  \\
Foroutan et al.~\cite{Foroutan2024EvaluatingAT}$^\dagger$ & \colorbox{tabthird}{28.88} & 0.862 & 0.204 & 28.80 & 0.884 & \colorbox{tabthird}{0.185} & 23.52 & 0.816 & \colorbox{tabthird}{0.257} & \colorbox{tabsecond}{29.61} & 0.886 & 0.297  \\
MCMC~\cite{kheradmand20243d} (rand. init.) & 28.38 & 0.844 & 0.237 & 28.23 & 0.874 & 0.212 & 22.96 & 0.808 & 0.279 & 28.84 & 0.875 & 0.315  \\
MCMC~\cite{kheradmand20243d} (SfM init.) & 28.82 & 0.861 & 0.214 & 28.72 & \colorbox{tabthird}{0.885} & 0.194 & \colorbox{tabsecond}{23.68} & \colorbox{tabthird}{0.825} & 0.259 & \colorbox{tabthird}{29.44} & 0.887 & 0.308 \\
MCMC~\cite{kheradmand20243d} (iNGP init.) & \colorbox{tabsecond}{29.02} & \colorbox{tabsecond}{0.867} & \colorbox{tabthird}{0.198} & 28.75 & \colorbox{tabthird}{0.885} & 0.186 & \colorbox{tabthird}{23.57} & 0.823 & \colorbox{tabsecond}{0.243} & \colorbox{tabsecond}{29.61} & 0.885 & \colorbox{tabthird}{0.288} \\
GaussianPro~\cite{Cheng2024ICML_GaussianPro_3D_Gaussian} & 28.02 & 0.827 & 0.253 & 28.32 & 0.876 & 0.206 & 22.31 & 0.802 & 0.286 & 29.33 & 0.886 & 0.301 \\
Perceptual-GS~\cite{Zhou2025PerceptualGSSP} & 28.66 & 0.856 & 0.211 & 28.42 & 0.876 & 0.203 & 22.77 & 0.813 & 0.267 & \colorbox{tabthird}{29.44} & \colorbox{tabthird}{0.888} & 0.296 \\
EDGS~\cite{Kotovenko2025ARXIV_EDGS_Eliminating_Densification} & 28.82 & \colorbox{tabthird}{0.865} & \colorbox{tabsecond}{0.193} & \colorbox{tabsecond}{29.01} & \colorbox{tabfirst}{0.893} & \colorbox{tabfirst}{0.168} & 23.47 & \colorbox{tabfirst}{0.831} & \colorbox{tabfirst}{0.229} & 29.33 & \colorbox{tabsecond}{0.889} & \colorbox{tabsecond}{0.286}  \\
Ours & \colorbox{tabfirst}{29.08} & \colorbox{tabfirst}{0.870} & \colorbox{tabfirst}{0.190} & \colorbox{tabfirst}{29.14} & \colorbox{tabsecond}{0.892} & \colorbox{tabsecond}{0.170} & \colorbox{tabfirst}{23.69} & \colorbox{tabsecond}{0.829} & \colorbox{tabfirst}{0.229} & \colorbox{tabfirst}{29.86} & \colorbox{tabfirst}{0.891} & \colorbox{tabfirst}{0.285} \\

\midrule
\multicolumn{13}{c}{Number of Gaussians bounded by Mini-Splatting2~\cite{Fang2024MiniSplatting2B3}} \\ 
\midrule
Mini-Splatting2~\cite{Fang2024MiniSplatting2B3} & \colorbox{tabsecond}{28.89} & \colorbox{tabfirst}{0.875} & \colorbox{tabsecond}{0.183} & \colorbox{tabsecond}{28.06} & \colorbox{tabsecond}{0.875} & \colorbox{tabsecond}{0.198} & \colorbox{tabsecond}{22.79} & \colorbox{tabsecond}{0.823} & \colorbox{tabsecond}{0.239} & \colorbox{tabfirst}{29.99} & \colorbox{tabfirst}{0.898} & \colorbox{tabfirst}{0.279} \\
Ours & \colorbox{tabfirst}{29.26} & \colorbox{tabfirst}{0.875} & \colorbox{tabfirst}{0.179} & \colorbox{tabfirst}{28.90} & \colorbox{tabfirst}{0.887} & \colorbox{tabfirst}{0.179} & \colorbox{tabfirst}{23.66} & \colorbox{tabfirst}{0.829} & \colorbox{tabfirst}{0.231} & \colorbox{tabsecond}{29.82} & \colorbox{tabsecond}{0.891} & \colorbox{tabsecond}{0.280} \\
\bottomrule
\end{tabular}
\begin{tablenotes}
\small
\item[$\dagger$] Original code was not publicly available. Our implementation uses iNGP initialization and does not include the additional depth-based loss.
\end{tablenotes}
\end{threeparttable}
}
\caption{\textbf{Quantitative results} with a Gaussian number limit of 100k and 500k. Mini-Splatting2~\cite{Fang2024MiniSplatting2B3} does not support constraining the number of Gaussians during reconstruction, so we match its Gaussian count. We highlight the \colorbox{tabfirst}{best}, \colorbox{tabsecond}{second best} and \colorbox{tabthird}{third best} results among methods with the same Gaussian counts. Per-scene metrics for selected methods are provided in the supplementary material.}
\label{tab:all_results}
\end{table*}

\section{Evaluation}
\label{sec:experiments}

\begin{figure*}
    \centering
    \renewcommand{\arraystretch}{0.5} 
    \setlength{\tabcolsep}{1pt}     
    \resizebox{0.95\linewidth}{!}{%
    \begin{tabular}{%
        >{\centering\arraybackslash}m{0.05\linewidth}
        >{\centering\arraybackslash}m{0.23\linewidth}
        >{\centering\arraybackslash}m{0.23\linewidth}
        >{\centering\arraybackslash}m{0.23\linewidth}
        >{\centering\arraybackslash}m{0.23\linewidth}
    }
         & EDGS~\cite{Kotovenko2025ARXIV_EDGS_Eliminating_Densification} & MCMC~\cite{kheradmand20243d} & Ours & GT \\[5pt]
        \rotatebox{90}{\makecell{Room~\cite{barron2022mipnerf360} \\ (50k)}} &
        \includegraphics[width=\linewidth]{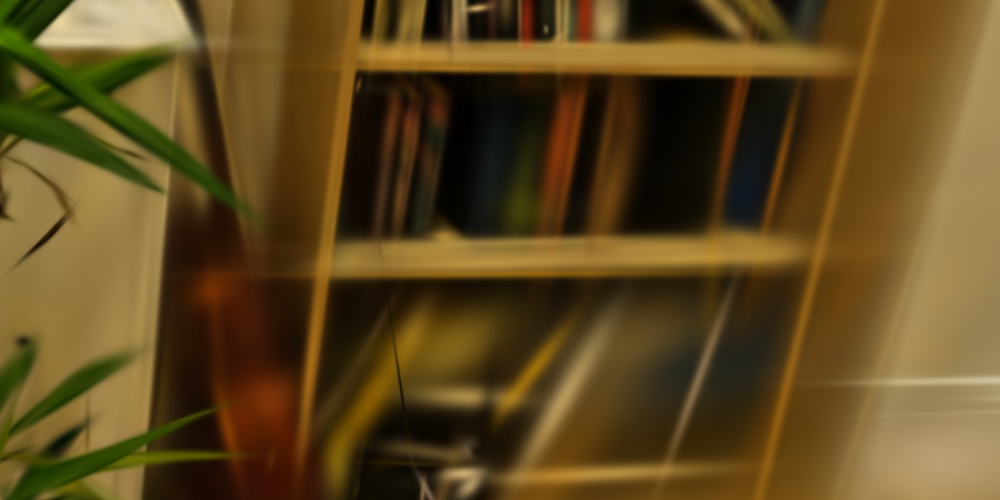} &
        \includegraphics[width=\linewidth]{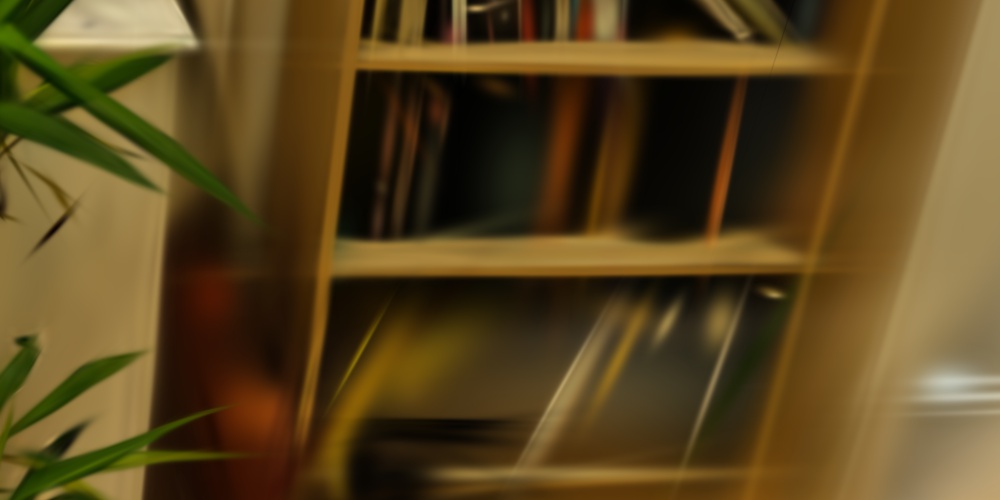} &
        \includegraphics[width=\linewidth]{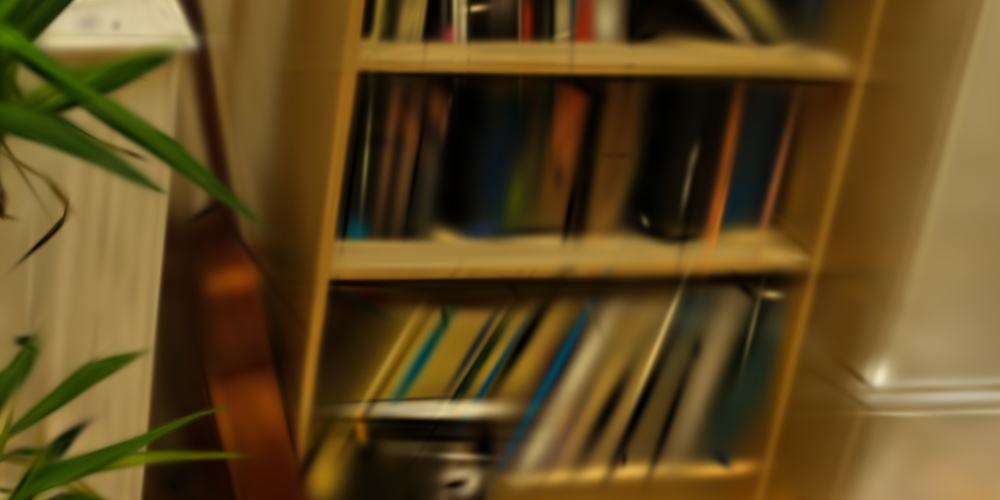} &
        \includegraphics[width=\linewidth]{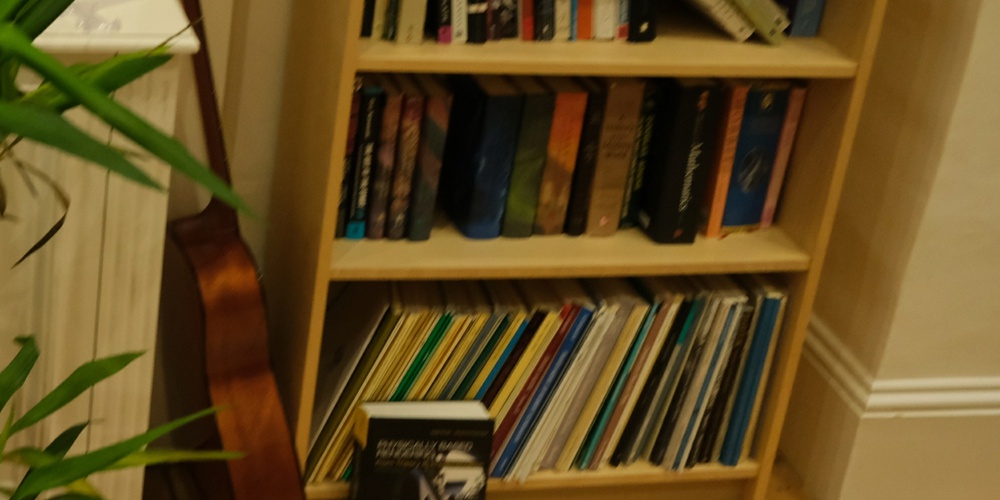} \\
        \rotatebox{90}{\makecell{Dr Johnson \\ \cite{DeepBlending2018} (100k)}} &
        \includegraphics[width=\linewidth]{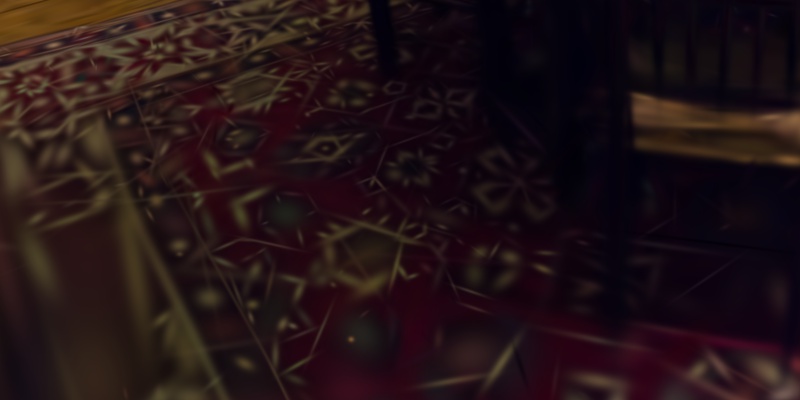} &
        \includegraphics[width=\linewidth]{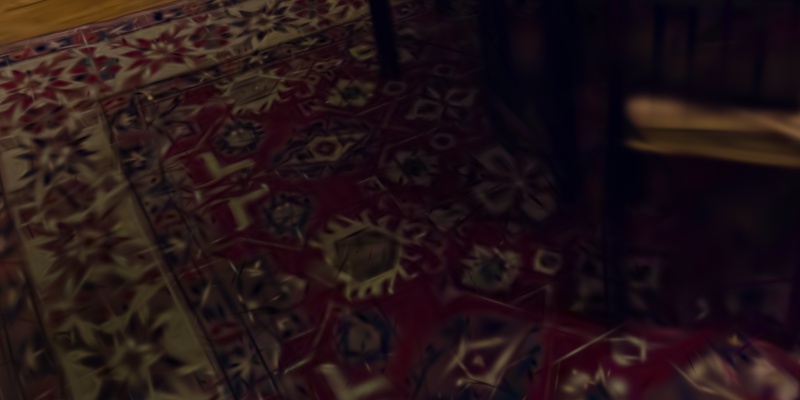} &
        \includegraphics[width=\linewidth]{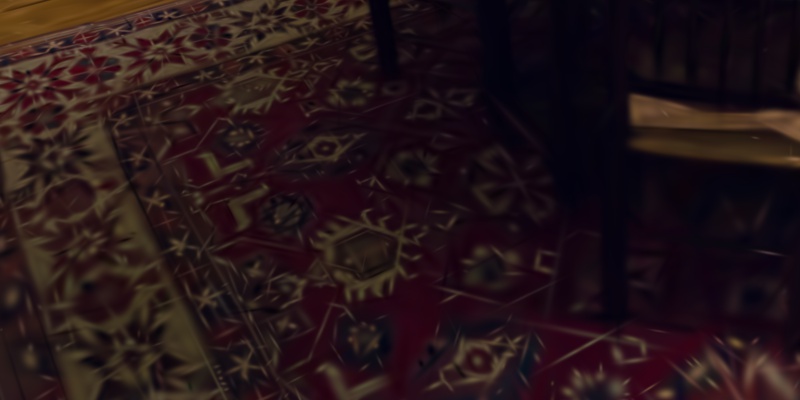} &
        \includegraphics[width=\linewidth]{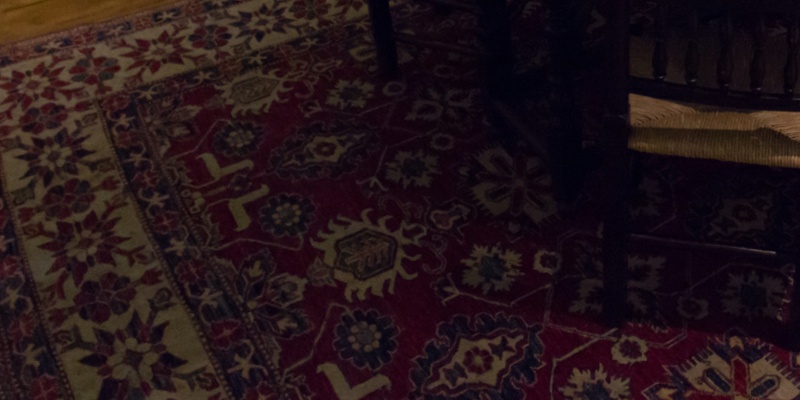} \\
        \rotatebox{90}{\makecell{05~\cite{lu2023largescaleoutdoormultimodaldataset} \\ (200k)}} &
        \includegraphics[width=\linewidth]{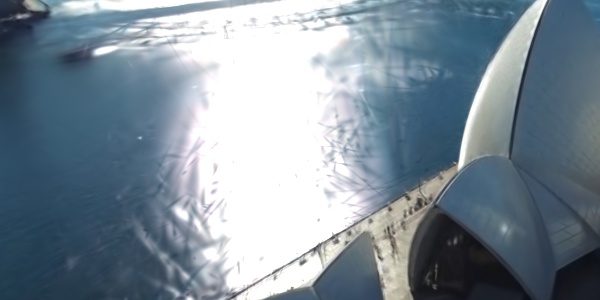} &
        \includegraphics[width=\linewidth]{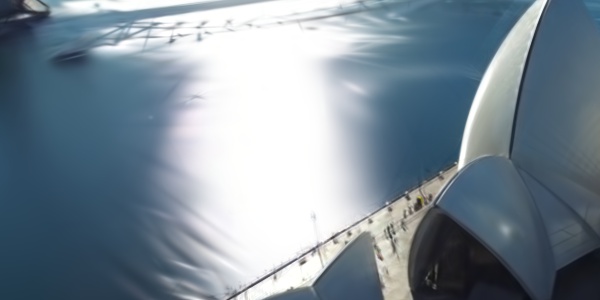} &
        \includegraphics[width=\linewidth]{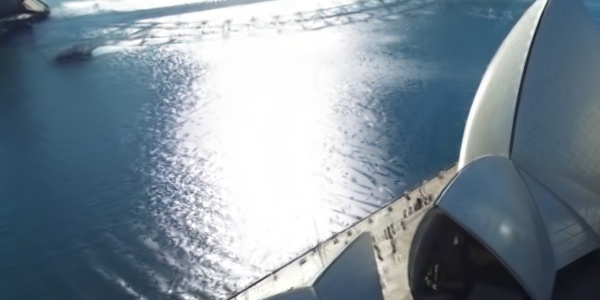} &
       \includegraphics[width=\linewidth]{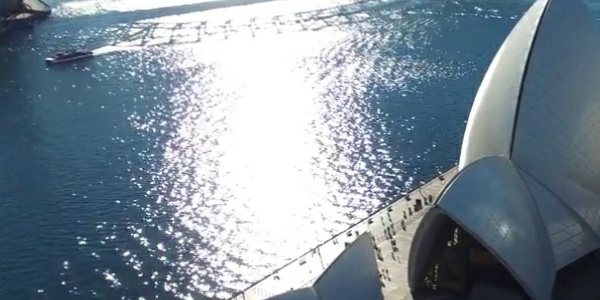} \\
        \rotatebox{90}{\makecell{13~\cite{lu2023largescaleoutdoormultimodaldataset} \\ (500k)}} &
        \includegraphics[width=\linewidth]{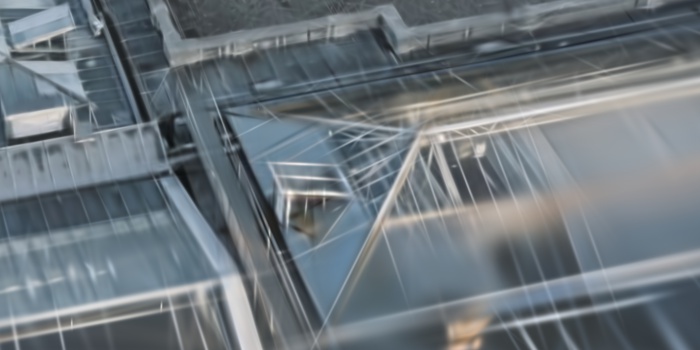} &
        \includegraphics[width=\linewidth]{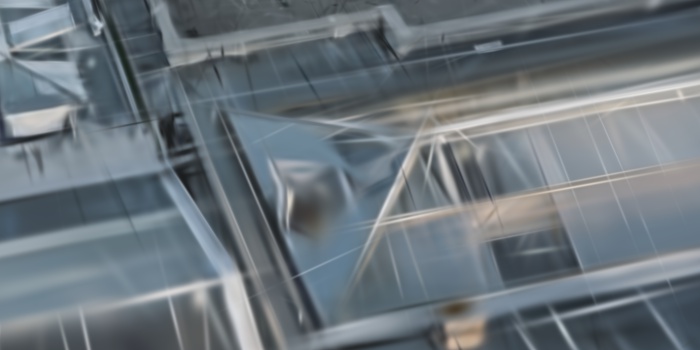} &
        \includegraphics[width=\linewidth]{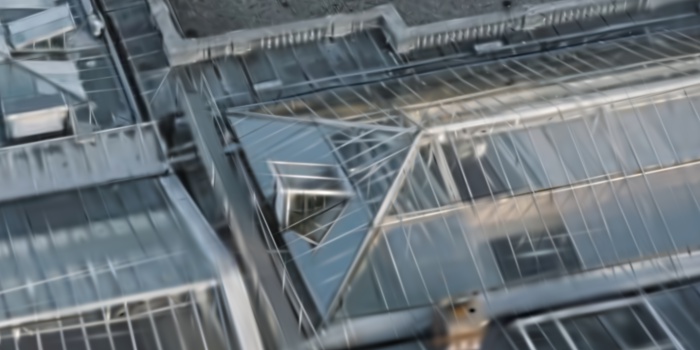} &
        \includegraphics[width=\linewidth]{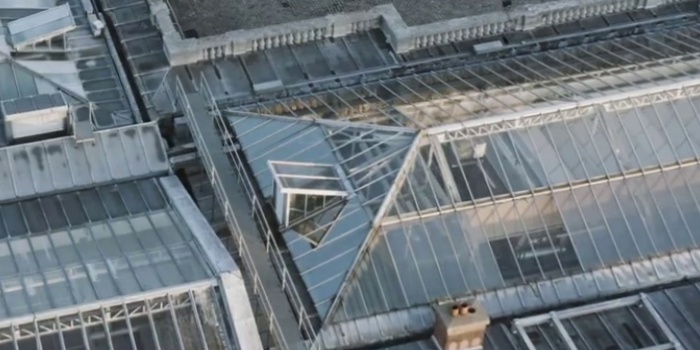} \\
    \end{tabular}
     }
    \caption{\textbf{Qualitative results} comparing our method with MCMC~\cite{kheradmand20243d} (with SfM point cloud initialization) and EDGS~\cite{Kotovenko2025ARXIV_EDGS_Eliminating_Densification} on the Mip-NeRF 360~\cite{barron2022mipnerf360}, OMMO~\cite{lu2023largescaleoutdoormultimodaldataset}, and DeepBlending~\cite{DeepBlending2018} datasets, with varying Gaussian budgets (given in parentheses).}
    \label{fig:image_comparison}
\end{figure*}

\begin{figure}
        \centering
        \resizebox{1\columnwidth}{!}{%
            \input{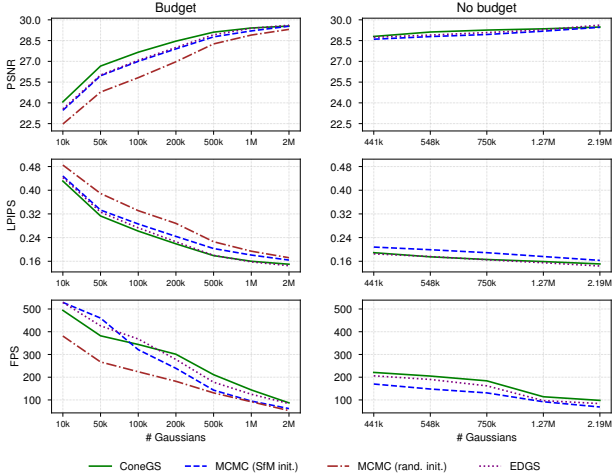}
        }
        \caption{\textbf{PSNR, LPIPS and FPS plots} with (left) and without (right) a primitive budget, where counts correspond to $\beta$ values. Averaged across Mip-NeRF360~\cite{barron2022mipnerf360} and OMMO~\cite{lu2023largescaleoutdoormultimodaldataset}. Numerical results are provided in the appendix.
        }
        \label{fig:lpips_psnr_fps_plots_combined}
\end{figure}

\boldparagraph{Dataset and Metrics}
We evaluate our method on publicly available scenes from Mip-NeRF360~\cite{barron2022mipnerf360} and OMMO dataset~\cite{lu2023largescaleoutdoormultimodaldataset}, with \texttt{01} scene from OMMO resized to have 1600 pixels width. Following~\cite{kerbl3Dgaussians, kheradmand20243d}, we also include the \texttt{train} and \texttt{truck} scene from Tanks \& Temples~\cite{Knapitsch2017}, as well as \texttt{Dr Johnson} and \texttt{playroom} from the DeepBlending~\cite{DeepBlending2018} dataset.
We report PSNR, SSIM~\cite{1284395}, and LPIPS~\cite{8578166}, with rendering speeds averaged across the full test set. All FPS measurements were recorded on an NVIDIA RTX 2080 Ti, whereas training speeds are reported on an NVIDIA A100, since EDGS requires more memory. These training speeds do not include later-added speed improvements~\cite{10.1145/3680528.3687694}.

\boldparagraph{Baselines}
We primarily compare our method against 3DGS~\cite{kerbl3Dgaussians}, it's extension with iNGP point cloud initialization~\cite{Foroutan2024EvaluatingAT}, MCMC~\cite{kheradmand20243d} using different initialization types (random, SfM, iNGP point clouds), as well as, GaussianPro~\cite{Cheng2024ICML_GaussianPro_3D_Gaussian}, Perceptual-GS~\cite{Zhou2025PerceptualGSSP}, EDGS (with densification)~\cite{Kotovenko2025ARXIV_EDGS_Eliminating_Densification}, with the densification stopped for all of them if the primitive budget is reached. If the number of primitives at initialization would be higher than the specified budget, the number of primitives is sampled uniformly to fit below it. In the random initialization settings we follow the process described in MCMC~\cite{kheradmand20243d}. We additionally test on Mini-Splatting2~\cite{Fang2024MiniSplatting2B3} by matching their final number of primitives instead of a specific budget, due to their method relying on generating a high number of initial Gaussians.

\begin{table}
\centering
\vspace{2pt}
\resizebox{1.0\columnwidth}{!}{
\begin{tabular}{l|ccc|cccc}
\toprule
 & \makecell{Ours \\ 10k iters} & \makecell{Ours \\ 20k iters} & \makecell{Ours \\ 40k iters} & \makecell{EDGS~\cite{Kotovenko2025ARXIV_EDGS_Eliminating_Densification}} & \makecell{3DGS~\cite{kerbl3Dgaussians} \\ (SfM init.)} & \makecell{MCMC~\cite{kheradmand20243d} \\ (rand. init.)} & \makecell{MCMC~\cite{kheradmand20243d} \\ (SfM init.)} \\
\midrule
\textbf{PSNR} $\uparrow$ & \colorbox{tabsecond}{29.33} & \colorbox{tabfirst}{29.37} & \colorbox{tabfirst}{29.37} & 29.18 & 28.71 & 28.98 & \colorbox{tabthird}{29.23 }\\
\textbf{SSIM} $\uparrow$ & \colorbox{tabthird}{0.877} & \colorbox{tabsecond}{0.880} & \colorbox{tabfirst}{0.881} & \colorbox{tabthird}{0.877} & 0.846 & 0.865 & 0.875 \\
\textbf{LPIPS} $\downarrow$ & \colorbox{tabthird}{0.171} & \colorbox{tabsecond}{0.168} & \colorbox{tabfirst}{0.166} & \colorbox{tabsecond}{0.168} & 0.222 & 0.204 & 0.190 \\
\textbf{FPS} $\uparrow$ & \colorbox{tabsecond}{134} & \colorbox{tabfirst}{137} & \colorbox{tabfirst}{137} & \colorbox{tabthird}{131} & 112 & 92 & 95 \\
\textbf{3DGS time} $\downarrow$ & \colorbox{tabthird}{20.7} & \colorbox{tabsecond}{20.5} & \colorbox{tabsecond}{20.5} & \colorbox{tabfirst}{19.4} & 22.6 & 26.3 & 25.1 \\
\textbf{Init. time} $\downarrow$ & 1.6 & 3.1 & 6.1 & 2.3 & - & - & - \\
\textbf{Overall time} $\downarrow$ & \colorbox{tabsecond}{22.3} & 23.6 & 26.6 & \colorbox{tabfirst}{21.7} & \colorbox{tabthird}{22.6} & 26.3 & 25.1 \\
\bottomrule
\end{tabular}
}
\caption{\textbf{Quantitative results} averaged over Mip-NeRF360~\cite{barron2022mipnerf360} showing reconstruction timings for 3DGS~\cite{kerbl3Dgaussians}, MCMC~\cite{kheradmand20243d}, EDGS~\cite{Kotovenko2025ARXIV_EDGS_Eliminating_Densification}, and our method (with varying iNGP durations), capped at 1M Gaussians. 3DGS time reports target scene optimization and densification, while init. time shows iNGP reconstruction for our method and initial matching for EDGS~\cite{Kotovenko2025ARXIV_EDGS_Eliminating_Densification}.}
\label{tab:timings}
\end{table}

\subsection{Results}
We observe improvements over the baselines across a wide range of specified budgets in \tabref{tab:all_results}, with plots comparing the most important methods on the budget and no-budget scenario in \figref{fig:lpips_psnr_fps_plots_combined}.
For a lower limit on Gaussians, we outperform the benchmarks across all datasets and metrics, while providing a competitive reconstruction quality compared to the best performing baselines on the high budget scenarios. In \tabref{tab:timings} we additionally show that even on a high number of primitives of 1M and including the iNGP training, our method provides competitive speed to other methods.
The qualitative results in \figref{fig:image_comparison} show significant improvement using a wide range of primitive budgets, demonstrating that for the same limit of primitives our method is able to produce a much better reconstruction, especially in areas that are challenging to properly capture on a low budget, such as isolated or high frequency structures. We provide additional qualitative and quantitative results, along with further scene analysis, in the appendix.

\subsection{Ablations}
\label{subsec:ablations}
We analyze each component's impact through ablation studies in \tabref{table:ablation}. \textbf{(a), (b)} Longer iNGP reconstruction only slightly improves reconstruction quality. \textbf{(c)} continuing training the iNGP model also during optimization, \textbf{(d)} initializing Gaussians with the ground truth pixel color, or \textbf{(e)} predicting spherical harmonics with iNGP, leads to marginally worse results. Demonstrating the strength of our densification method, using \textbf{(f)} pixel-cone-sized primitives during initialization, or even not using any initialization \textbf{(g)}, results in worse PSNR but maintains low LPIPS and considerably improves rendering speed, thanks to less overlap between primitives, reducing blending.
\textbf{(h)} Sampling pixels uniformly instead of guiding Gaussian creation using the $L_1$ loss from the training set produces lower reconstruction quality, although due to uniform sampling in image space still focusing more on parts of the scene seen most across views, the drop is not drastic. Densifying with 3DGS depth \textbf{(i)}, also without using iNGP even for initialization \textbf{(j)}, strongly affects the results. Similarly, k-NN sizing of newly added primitives based on their closest neighbors \textbf{(k)}, or changing the opacity penalty \textbf{(l), (m), (n)}, has a large effect on reconstruction quality, reinforcing the benefits of our densification approach.
\textbf{(o), (p)} Altering Gaussian sizes from their default pixel-width cone size during densification slightly reduces quality, yet the small difference suggests Gaussians quickly resize to fit the scene.

\begin{table}
        \centering
        \vspace{10pt}
        \resizebox{\columnwidth}{!}{
        \begin{tabular}{l|cccc}
        \toprule
         \textit{Ablation} & \textbf{PSNR} $\uparrow$ & \textbf{SSIM} $\uparrow$ & \textbf{LPIPS} $\downarrow$ & \textbf{FPS} $\uparrow$ \\
        \midrule
Ours & \colorbox{tabfirst}{27.74} & \colorbox{tabsecond}{0.810} & \colorbox{tabsecond}{0.285} & 328 \\
\midrule
(a) 10k iNGP iter. & \colorbox{tabfirst}{27.74} & 0.808 & 0.287 & 313 \\
(b) 40k iNGP iter. & \colorbox{tabthird}{27.72} & \colorbox{tabfirst}{0.811} & \colorbox{tabfirst}{0.284} & \colorbox{tabthird}{333} \\
(c) Train iNGP during 3DGS & \colorbox{tabsecond}{27.73} & \colorbox{tabthird}{0.809} & \colorbox{tabsecond}{0.285} & 310 \\
(d) Color from GT image & \colorbox{tabsecond}{27.73} & 0.807 & 0.287 & 320 \\
(e) Prediction of SH with iNGP & \colorbox{tabfirst}{27.74} & 0.805 & 0.290 & 320 \\
\midrule
(f) Cone-sized initialization & 27.38 & 0.806 & 0.287 & \colorbox{tabfirst}{437} \\
(g) Without initialization & 27.46 & \colorbox{tabfirst}{0.811} & \colorbox{tabsecond}{0.285} & \colorbox{tabsecond}{415} \\
\midrule
(h) Uniform image-space sampling & 27.51 & 0.806 & \colorbox{tabthird}{0.286} & 307 \\
(i) Densify with 3DGS depth & 27.43 & 0.797 & 0.296 & 332 \\
(j) SfM initialization + 3DGS depth dens. & 27.15 & 0.790 & 0.302 & 329 \\
(k) Densify with k-NN scaling & 27.54 & 0.802 & 0.295 & 299 \\
(l) No opacity penalty & 27.31 & 0.794 & 0.301 & 294 \\
(m) Post-densification opacity decrease~\cite{Bul2024RevisingDI} & 27.46 & 0.798 & 0.297 & 256 \\
(n) MCMC-style opacity penalty~\cite{kheradmand20243d} & 27.49 & 0.803 & 0.293 & 239 \\
\midrule
(o) \(\lambda_{\text{scale}}=1\) & 27.70 & \colorbox{tabsecond}{0.810} & \colorbox{tabsecond}{0.285} & 329 \\
(p) \(\lambda_{\text{scale}}=4\) & 27.63 & 0.808 & \colorbox{tabsecond}{0.285} & 329 \\
        \bottomrule
        \end{tabular}
        }
        \caption{\textbf{Ablation study} of our method with $100$k Gaussians, averaged over the Mip-NeRF360~\cite{barron2022mipnerf360} dataset. We highlight the \colorbox{tabfirst}{best}, \colorbox{tabsecond}{second best}, and \colorbox{tabthird}{third best} results.}
        \label{table:ablation}
\end{table}

\section{Conclusion}
We introduce \textbf{ConeGS}, a reconstruction pipeline replacing cloning-based densification with a method guided by photometric error and a coarse iNGP proxy, where new primitives are sized by pixel cones. Together with an improved opacity penalty, this allows creating primitives independently of existing structures through more flexible exploration. ConeGS consistently improves reconstruction quality and rendering performance across Gaussian budgets, with strong gains under tight primitive constraints. It achieves up to 0.6 PSNR increase and 20\% speedup over cloning-based baselines.

\boldparagraph{Limitations}
ConeGS performs well on standard scenes but may struggle with large-scale environments, inaccurate poses, or sparse views, occasionally creating floaters (see appendix). Benefits are also reduced at high Gaussian budgets, where dense coverage limits the impact of error-guided placement, primarily offering faster rendering.

\section*{Acknowledgments}

Stefano Esposito acknowledges travel support from the European Union’s Horizon 2020 research and innovation program under ELISE Grant Agreement No. 951847.


{
    \small
    \bibliographystyle{ieeenat_fullname}
    \bibliography{main}

\begin{thebibliography}{67}
\providecommand{\natexlab}[1]{#1}
\providecommand{\url}[1]{\texttt{#1}}
\expandafter\ifx\csname urlstyle\endcsname\relax
  \providecommand{\doi}[1]{doi: #1}\else
  \providecommand{\doi}{doi: \begingroup \urlstyle{rm}\Url}\fi

\bibitem[Abdul~Gafoor et~al.(2025)Abdul~Gafoor, Preda, and Zaharia]{refining_gaussian_splatting_2025}
Mohamed Abdul~Gafoor, Marius Preda, and Titus Zaharia.
\newblock Refining gaussian splatting: A volumetric densification approach.
\newblock In \emph{Computer Science Research Notes}. University of West Bohemia, Czech Republic, 2025.

\bibitem[Barron et~al.(2021)Barron, Mildenhall, Tancik, Hedman, Martin-Brualla, and Srinivasan]{barron2021mipnerf}
Jonathan~T. Barron, Ben Mildenhall, Matthew Tancik, Peter Hedman, Ricardo Martin-Brualla, and Pratul~P. Srinivasan.
\newblock Mip-nerf: A multiscale representation for anti-aliasing neural radiance fields.
\newblock \emph{ICCV}, 2021.

\bibitem[Barron et~al.(2022)Barron, Mildenhall, Verbin, Srinivasan, and Hedman]{barron2022mipnerf360}
Jonathan~T. Barron, Ben Mildenhall, Dor Verbin, Pratul~P. Srinivasan, and Peter Hedman.
\newblock Mip-nerf 360: Unbounded anti-aliased neural radiance fields.
\newblock \emph{CVPR}, 2022.

\bibitem[Barron et~al.(2023)Barron, Mildenhall, Verbin, Srinivasan, and Hedman]{barron2023zipnerf}
Jonathan~T. Barron, Ben Mildenhall, Dor Verbin, Pratul~P. Srinivasan, and Peter Hedman.
\newblock Zip-nerf: Anti-aliased grid-based neural radiance fields.
\newblock \emph{ICCV}, 2023.

\bibitem[Bul{\`o} et~al.(2024)Bul{\`o}, Porzi, and Kontschieder]{Bul2024RevisingDI}
Samuel~Rota Bul{\`o}, Lorenzo Porzi, and Peter Kontschieder.
\newblock Revising densification in gaussian splatting.
\newblock \emph{ArXiv}, abs/2404.06109, 2024.

\bibitem[Charatan et~al.(2024)Charatan, Li, Tagliasacchi, and Sitzmann]{charatan23pixelsplat}
David Charatan, Sizhe Li, Andrea Tagliasacchi, and Vincent Sitzmann.
\newblock pixelsplat: 3d gaussian splats from image pairs for scalable generalizable 3d reconstruction.
\newblock In \emph{CVPR}, 2024.

\bibitem[Chen et~al.(2022)Chen, Xu, Geiger, Yu, and Su]{Chen2022ECCV}
Anpei Chen, Zexiang Xu, Andreas Geiger, Jingyi Yu, and Hao Su.
\newblock Tensorf: Tensorial radiance fields.
\newblock In \emph{ECCV}, 2022.

\bibitem[Chen et~al.(2024)Chen, Xu, Zheng, Zhuang, Pollefeys, Geiger, Cham, and Cai]{chen2024mvsplat}
Yuedong Chen, Haofei Xu, Chuanxia Zheng, Bohan Zhuang, Marc Pollefeys, Andreas Geiger, Tat-Jen Cham, and Jianfei Cai.
\newblock Mvsplat: Efficient 3d gaussian splatting from sparse multi-view images.
\newblock \emph{arXiv preprint arXiv:2403.14627}, 2024.

\bibitem[Cheng et~al.(2024)Cheng, Long, Yang, Yao, Yin, Ma, Wang, and Chen]{Cheng2024ICML_GaussianPro_3D_Gaussian}
Kai Cheng, Xiaoxiao Long, Kaizhi Yang, Yao Yao, Wei Yin, Yuexin Ma, Wenping Wang, and Xuejin Chen.
\newblock {GaussianPro:} 3d gaussian splatting with progressive propagation.
\newblock In \emph{International Conference on Machine Learning (ICML)}, 2024.

\bibitem[Fan et~al.(2023)Fan, Wang, Wen, Zhu, Xu, and Wang]{fan2023lightgaussian}
Zhiwen Fan, Kevin Wang, Kairun Wen, Zehao Zhu, Dejia Xu, and Zhangyang Wang.
\newblock Lightgaussian: Unbounded 3d gaussian compression with 15x reduction and 200+ fps, 2023.

\bibitem[Fang and Wang(2024{\natexlab{a}})]{Fang2024MiniSplatting2B3}
Guangchi Fang and Bing Wang.
\newblock Mini-splatting2: Building 360 scenes within minutes via aggressive gaussian densification.
\newblock \emph{ArXiv}, abs/2411.12788, 2024{\natexlab{a}}.

\bibitem[Fang and Wang(2024{\natexlab{b}})]{fang2024minisplattingrepresentingscenesconstrained}
Guangchi Fang and Bing Wang.
\newblock Mini-splatting: Representing scenes with a constrained number of gaussians, 2024{\natexlab{b}}.

\bibitem[Fang et~al.(2025)Fang, Shen, Igarashi, Wang, Wang, Yang, Ding, and Zhou]{fang2025nerfvaluableassistant3d}
Shuangkang Fang, I-Chao Shen, Takeo Igarashi, Yufeng Wang, ZeSheng Wang, Yi Yang, Wenrui Ding, and Shuchang Zhou.
\newblock Nerf is a valuable assistant for 3d gaussian splatting, 2025.

\bibitem[Foroutan et~al.(2024)Foroutan, Rebain, Yi, and Tagliasacchi]{Foroutan2024EvaluatingAT}
Yalda Foroutan, Daniel Rebain, Kwang~Moo Yi, and Andrea Tagliasacchi.
\newblock Evaluating alternatives to sfm point cloud initialization for gaussian splatting.
\newblock 2024.

\bibitem[Girish et~al.(2024)Girish, Gupta, and Shrivastava]{girish2024eaglesefficientaccelerated3d}
Sharath Girish, Kamal Gupta, and Abhinav Shrivastava.
\newblock Eagles: Efficient accelerated 3d gaussians with lightweight encodings, 2024.

\bibitem[Gu{\'e}don and Lepetit(2024)]{guedon2023sugar}
Antoine Gu{\'e}don and Vincent Lepetit.
\newblock Sugar: Surface-aligned gaussian splatting for efficient 3d mesh reconstruction and high-quality mesh rendering.
\newblock \emph{CVPR}, 2024.

\bibitem[Guo et~al.(2025)Guo, Su, Wang, Fan, Zhang, Han, and Wang]{Guo2025GPGSGP}
Zhihao Guo, Jingxuan Su, Shenglin Wang, Jinlong Fan, Jing Zhang, Li~Hong Han, and Peng Wang.
\newblock Gp-gs: Gaussian processes for enhanced gaussian splatting.
\newblock \emph{ArXiv}, abs/2502.02283, 2025.

\bibitem[Hamdi et~al.(2024)Hamdi, Melas-Kyriazi, Mai, Qian, Liu, Vondrick, Ghanem, and Vedaldi]{Hamdi_2024_CVPR}
Abdullah Hamdi, Luke Melas-Kyriazi, Jinjie Mai, Guocheng Qian, Ruoshi Liu, Carl Vondrick, Bernard Ghanem, and Andrea Vedaldi.
\newblock Ges : Generalized exponential splatting for efficient radiance field rendering.
\newblock In \emph{Proceedings of the IEEE/CVF Conference on Computer Vision and Pattern Recognition (CVPR)}, pages 19812--19822, 2024.

\bibitem[Hedman et~al.(2018)Hedman, Philip, Price, Frahm, Drettakis, and Brostow]{DeepBlending2018}
Peter Hedman, Julien Philip, True Price, Jan-Michael Frahm, George Drettakis, and Gabriel Brostow.
\newblock Deep blending for free-viewpoint image-based rendering.
\newblock 2018.

\bibitem[Held et~al.(2025)Held, Vandeghen, Hamdi, Deliege, Cioppa, Giancola, Vedaldi, Ghanem, and Van~Droogenbroeck]{Held20243DConvex}
Jan Held, Renaud Vandeghen, Abdullah Hamdi, Adrien Deliege, Anthony Cioppa, Silvio Giancola, Andrea Vedaldi, Bernard Ghanem, and Marc Van~Droogenbroeck.
\newblock {3D} convex splatting: Radiance field rendering with {3D} smooth convexes.
\newblock In \emph{Proceedings of the IEEE/CVF Conference on Computer Vision and Pattern Recognition (CVPR)}, 2025.

\bibitem[Huang et~al.(2024)Huang, Yu, Chen, Geiger, and Gao]{Huang2DGS2024}
Binbin Huang, Zehao Yu, Anpei Chen, Andreas Geiger, and Shenghua Gao.
\newblock 2d gaussian splatting for geometrically accurate radiance fields.
\newblock 2024.

\bibitem[Huang et~al.(2025)Huang, Liu, and Wong]{huang2025decomposingdensificationgaussiansplatting}
Binxiao Huang, Zhengwu Liu, and Ngai Wong.
\newblock Decomposing densification in gaussian splatting for faster 3d scene reconstruction, 2025.

\bibitem[Jiang et~al.(2024)Jiang, Xiang, Sun, Li, Zhou, Zhang, and Zhang]{jiang2024geotexdensifiergeometrytextureawaredensificationhighquality}
Hanqing Jiang, Xiaojun Xiang, Han Sun, Hongjie Li, Liyang Zhou, Xiaoyu Zhang, and Guofeng Zhang.
\newblock Geotexdensifier: Geometry-texture-aware densification for high-quality photorealistic 3d gaussian splatting, 2024.

\bibitem[Kerbl et~al.(2023)Kerbl, Kopanas, Leimk{\"u}hler, and Drettakis]{kerbl3Dgaussians}
Bernhard Kerbl, Georgios Kopanas, Thomas Leimk{\"u}hler, and George Drettakis.
\newblock 3d gaussian splatting for real-time radiance field rendering.
\newblock \emph{ACM Transactions on Graphics}, 42\penalty0 (4), 2023.

\bibitem[Kheradmand et~al.(2024)Kheradmand, Rebain, Sharma, Sun, Tseng, Isack, Kar, Tagliasacchi, and Yi]{kheradmand20243d}
Shakiba Kheradmand, Daniel Rebain, Gopal Sharma, Weiwei Sun, Yang-Che Tseng, Hossam Isack, Abhishek Kar, Andrea Tagliasacchi, and Kwang~Moo Yi.
\newblock 3d gaussian splatting as markov chain monte carlo.
\newblock In \emph{Conference on Neural Information Processing Systems (NeurIPS)}, 2024.

\bibitem[Knapitsch et~al.(2017)Knapitsch, Park, Zhou, and Koltun]{Knapitsch2017}
Arno Knapitsch, Jaesik Park, Qian-Yi Zhou, and Vladlen Koltun.
\newblock Tanks and temples: Benchmarking large-scale scene reconstruction.
\newblock \emph{ACM Transactions on Graphics}, 36\penalty0 (4), 2017.

\bibitem[Kotovenko et~al.(2025)Kotovenko, Grebenkova, and Ommer]{Kotovenko2025ARXIV_EDGS_Eliminating_Densification}
Dmytro Kotovenko, Olga Grebenkova, and Björn Ommer.
\newblock {EDGS:} eliminating densification for efficient convergence of 3dgs.
\newblock \emph{arXiv}, 2504.13204, 2025.

\bibitem[Li et~al.(2024)Li, Zhang, Bai, Zheng, Ning, Zhou, and Gu]{li2024dngaussian}
Jiahe Li, Jiawei Zhang, Xiao Bai, Jin Zheng, Xin Ning, Jun Zhou, and Lin Gu.
\newblock Dngaussian: Optimizing sparse-view 3d gaussian radiance fields with global-local depth normalization.
\newblock \emph{arXiv preprint arXiv:2403.06912}, 2024.

\bibitem[Li et~al.(2025)Li, Liu, Deng, and Wang]{li2025densesplatdensifyinggaussiansplatting}
Mingrui Li, Shuhong Liu, Tianchen Deng, and Hongyu Wang.
\newblock Densesplat: Densifying gaussian splatting slam with neural radiance prior, 2025.

\bibitem[Li et~al.(2023)Li, Gao, Tancik, and Kanazawa]{li2023nerfacc}
Ruilong Li, Hang Gao, Matthew Tancik, and Angjoo Kanazawa.
\newblock Nerfacc: Efficient sampling accelerates nerfs.
\newblock \emph{arXiv preprint arXiv:2305.04966}, 2023.

\bibitem[Liu et~al.(2025{\natexlab{a}})Liu, Sun, Chen, Wang, and Feng]{liu2025deformablebetasplatting}
Rong Liu, Dylan Sun, Meida Chen, Yue Wang, and Andrew Feng.
\newblock Deformable beta splatting, 2025{\natexlab{a}}.

\bibitem[Liu et~al.(2025{\natexlab{b}})Liu, H{\"o}llein, Nie{\ss}ner, and Dai]{Liu2025QuickSplatF3}
Yueh-Cheng Liu, Lukas H{\"o}llein, Matthias Nie{\ss}ner, and Angela Dai.
\newblock Quicksplat: Fast 3d surface reconstruction via learned gaussian initialization.
\newblock \emph{ArXiv}, abs/2505.05591, 2025{\natexlab{b}}.

\bibitem[Lu et~al.(2023)Lu, Yin, Chen, Chen, YU, and Fan]{lu2023largescaleoutdoormultimodaldataset}
Chongshan Lu, Fukun Yin, Xin Chen, Tao Chen, Gang YU, and Jiayuan Fan.
\newblock A large-scale outdoor multi-modal dataset and benchmark for novel view synthesis and implicit scene reconstruction, 2023.

\bibitem[Lu et~al.(2024)Lu, Yu, Xu, Xiangli, Wang, Lin, and Dai]{scaffoldgs}
Tao Lu, Mulin Yu, Linning Xu, Yuanbo Xiangli, Limin Wang, Dahua Lin, and Bo Dai.
\newblock Scaffold-gs: Structured 3d gaussians for view-adaptive rendering.
\newblock In \emph{CVPR}, 2024.

\bibitem[Mai et~al.(2024)Mai, Hedman, Kopanas, Verbin, Futschik, Xu, Kuester, Barron, and Zhang]{mai2024everexactvolumetricellipsoid}
Alexander Mai, Peter Hedman, George Kopanas, Dor Verbin, David Futschik, Qiangeng Xu, Falko Kuester, Jon Barron, and Yinda Zhang.
\newblock Ever: Exact volumetric ellipsoid rendering for real-time view synthesis, 2024.

\bibitem[Mallick et~al.(2024)Mallick, Goel, Kerbl, Steinberger, Carrasco, and De~La~Torre]{10.1145/3680528.3687694}
Saswat~Subhajyoti Mallick, Rahul Goel, Bernhard Kerbl, Markus Steinberger, Francisco~Vicente Carrasco, and Fernando De~La~Torre.
\newblock Taming 3dgs: High-quality radiance fields with limited resources.
\newblock In \emph{SIGGRAPH Asia 2024 Conference Papers}, New York, NY, USA, 2024. Association for Computing Machinery.

\bibitem[Mildenhall et~al.(2020)Mildenhall, Srinivasan, Tancik, Barron, Ramamoorthi, and Ng]{mildenhall2020nerf}
Ben Mildenhall, Pratul~P. Srinivasan, Matthew Tancik, Jonathan~T. Barron, Ravi Ramamoorthi, and Ren Ng.
\newblock Nerf: Representing scenes as neural radiance fields for view synthesis.
\newblock In \emph{ECCV}, 2020.

\bibitem[Mohamad et~al.(2024)Mohamad, Elghazaly, Hubert, and Frank]{mohamad2024denser3dgaussianssplatting}
Mahmud~A. Mohamad, Gamal Elghazaly, Arthur Hubert, and Raphael Frank.
\newblock Denser: 3d gaussians splatting for scene reconstruction of dynamic urban environments, 2024.

\bibitem[M\"uller et~al.(2022)M\"uller, Evans, Schied, and Keller]{mueller2022instant}
Thomas M\"uller, Alex Evans, Christoph Schied, and Alexander Keller.
\newblock Instant neural graphics primitives with a multiresolution hash encoding.
\newblock \emph{ACM TOG}, 2022.

\bibitem[Nam et~al.(2024)Nam, Sun, Kang, Lee, Oh, and Park]{GenerativeDensification}
Seungtae Nam, Xiangyu Sun, Gyeongjin Kang, Younggeun Lee, Seungjun Oh, and Eunbyung Park.
\newblock Generative densification: Learning to densify gaussians for high-fidelity generalizable 3d reconstruction.
\newblock \emph{arXiv preprint arXiv:2412.06234}, 2024.

\bibitem[Niedermayr et~al.(2024)Niedermayr, Stumpfegger, and Westermann]{Niedermayr_2024_CVPR}
Simon Niedermayr, Josef Stumpfegger, and R\"udiger Westermann.
\newblock Compressed 3d gaussian splatting for accelerated novel view synthesis.
\newblock In \emph{CVPR}, 2024.

\bibitem[Niemeyer et~al.(2024)Niemeyer, Manhardt, Rakotosaona, Oechsle, Duckworth, Gosula, Tateno, Bates, Kaeser, and Tombari]{niemeyer2024radsplat}
Michael Niemeyer, Fabian Manhardt, Marie-Julie Rakotosaona, Michael Oechsle, Daniel Duckworth, Rama Gosula, Keisuke Tateno, John Bates, Dominik Kaeser, and Federico Tombari.
\newblock Radsplat: Radiance field-informed gaussian splatting for robust real-time rendering with 900+ fps.
\newblock \emph{arXiv.org}, 2024.

\bibitem[Radl et~al.(2024)Radl, Steiner, Parger, Weinrauch, Kerbl, and Steinberger]{radl2024stopthepop}
Lukas Radl, Michael Steiner, Mathias Parger, Alexander Weinrauch, Bernhard Kerbl, and Markus Steinberger.
\newblock {StopThePop: Sorted Gaussian Splatting for View-Consistent Real-time Rendering}.
\newblock \emph{ACM Transactions on Graphics}, 4\penalty0 (43), 2024.

\bibitem[Ren et~al.(2024)Ren, Jiang, Lu, Yu, Xu, Ni, and Dai]{Ren2024OctreeGSTC}
Kerui Ren, Lihan Jiang, Tao Lu, Mulin Yu, Linning Xu, Zhangkai Ni, and Bo Dai.
\newblock Octree-gs: Towards consistent real-time rendering with lod-structured 3d gaussians.
\newblock \emph{IEEE transactions on pattern analysis and machine intelligence}, PP, 2024.

\bibitem[{Sara Fridovich-Keil and Alex Yu} et~al.(2022){Sara Fridovich-Keil and Alex Yu}, Tancik, Chen, Recht, and Kanazawa]{yu_and_fridovichkeil2021plenoxels}
{Sara Fridovich-Keil and Alex Yu}, Matthew Tancik, Qinhong Chen, Benjamin Recht, and Angjoo Kanazawa.
\newblock Plenoxels: Radiance fields without neural networks.
\newblock In \emph{CVPR}, 2022.

\bibitem[Sun et~al.(2022)Sun, Sun, and Chen]{SunSC22}
Cheng Sun, Min Sun, and Hwann{-}Tzong Chen.
\newblock Direct voxel grid optimization: Super-fast convergence for radiance fields reconstruction.
\newblock In \emph{CVPR}, 2022.

\bibitem[Talegaonkar et~al.(2025)Talegaonkar, Belhe, Ramamoorthi, and Antipa]{talegaonkar2025volumetricallyconsistent3dgaussian}
Chinmay Talegaonkar, Yash Belhe, Ravi Ramamoorthi, and Nicholas Antipa.
\newblock Volumetrically consistent 3d gaussian rasterization, 2025.

\bibitem[Tang et~al.(2023)Tang, Ren, Zhou, Liu, and Zeng]{tang2023dreamgaussian}
Jiaxiang Tang, Jiawei Ren, Hang Zhou, Ziwei Liu, and Gang Zeng.
\newblock Dreamgaussian: Generative gaussian splatting for efficient 3d content creation.
\newblock \emph{arXiv preprint arXiv:2309.16653}, 2023.

\bibitem[von L{\"u}tzow and Nie{\ss}ner(2025)]{Ltzow2025LinPrimLP}
Nicolas von L{\"u}tzow and Matthias Nie{\ss}ner.
\newblock Linprim: Linear primitives for differentiable volumetric rendering.
\newblock \emph{ArXiv}, abs/2501.16312, 2025.

\bibitem[Wang and Xu(2024)]{wang2024pygslargescalescenerepresentation}
Zipeng Wang and Dan Xu.
\newblock Pygs: Large-scale scene representation with pyramidal 3d gaussian splatting, 2024.

\bibitem[Wang et~al.(2004)Wang, Bovik, Sheikh, and Simoncelli]{1284395}
Zhou Wang, A.C. Bovik, H.R. Sheikh, and E.P. Simoncelli.
\newblock Image quality assessment: from error visibility to structural similarity.
\newblock \emph{IEEE Transactions on Image Processing}, 2004.

\bibitem[Wu et~al.(2024)Wu, Yi, Fang, Xie, Zhang, Wei, Liu, Tian, and Wang]{Wu_2024_CVPR}
Guanjun Wu, Taoran Yi, Jiemin Fang, Lingxi Xie, Xiaopeng Zhang, Wei Wei, Wenyu Liu, Qi Tian, and Xinggang Wang.
\newblock 4d gaussian splatting for real-time dynamic scene rendering.
\newblock In \emph{CVPR}, 2024.

\bibitem[Yan et~al.(2024)Yan, Low, Chen, and Lee]{Yan2024CVPR}
Zhiwen Yan, Weng~Fei Low, Yu Chen, and Gim~Hee Lee.
\newblock Multi-scale 3d gaussian splatting for anti-aliased rendering.
\newblock In \emph{CVPR}, 2024.

\bibitem[Yang et~al.(2023)Yang, Gao, Zhou, Jiao, Zhang, and Jin]{yang2023deformable3dgs}
Ziyi Yang, Xinyu Gao, Wen Zhou, Shaohui Jiao, Yuqing Zhang, and Xiaogang Jin.
\newblock Deformable 3d gaussians for high-fidelity monocular dynamic scene reconstruction.
\newblock \emph{arXiv preprint arXiv:2309.13101}, 2023.

\bibitem[Ye et~al.(2024)Ye, Li, Liu, Qiao, and Dou]{ye2024absgs}
Zongxin Ye, Wenyu Li, Sidun Liu, Peng Qiao, and Yong Dou.
\newblock Abs{GS}: Recovering fine details in 3d gaussian splatting.
\newblock In \emph{ACM MM}, 2024.

\bibitem[Yu et~al.(2021)Yu, Li, Tancik, Li, Ng, and Kanazawa]{yu2021plenoctrees}
Alex Yu, Ruilong Li, Matthew Tancik, Hao Li, Ren Ng, and Angjoo Kanazawa.
\newblock {PlenOctrees} for real-time rendering of neural radiance fields.
\newblock In \emph{ICCV}, 2021.

\bibitem[Yu et~al.(2024{\natexlab{a}})Yu, Chen, Huang, Sattler, and Geiger]{Yu2023MipSplattingA3}
Zehao Yu, Anpei Chen, Binbin Huang, Torsten Sattler, and Andreas Geiger.
\newblock Mip-splatting: Alias-free 3d gaussian splatting.
\newblock \emph{CVPR}, 2024{\natexlab{a}}.

\bibitem[Yu et~al.(2024{\natexlab{b}})Yu, Sattler, and Geiger]{Yu2024GOF}
Zehao Yu, Torsten Sattler, and Andreas Geiger.
\newblock Gaussian opacity fields: Efficient high-quality compact surface reconstruction in unbounded scenes.
\newblock \emph{arXiv:2404.10772}, 2024{\natexlab{b}}.

\bibitem[Zeng et~al.(2025)Zeng, Wang, Ju, and Guan]{Zeng2025FrequencyAwareDC}
Zhaojie Zeng, Yuesong Wang, Lili Ju, and Tao Guan.
\newblock Frequency-aware density control via reparameterization for high-quality rendering of 3d gaussian splatting.
\newblock \emph{ArXiv}, abs/2503.07000, 2025.

\bibitem[Zhang et~al.(2024{\natexlab{a}})Zhang, Zhan, Xu, Lu, and Xing]{Zhang2024FreGS3G}
Jiahui Zhang, Fangneng Zhan, Muyu Xu, Shijian Lu, and Eric~P. Xing.
\newblock Fregs: 3d gaussian splatting with progressive frequency regularization.
\newblock \emph{2024 IEEE/CVF Conference on Computer Vision and Pattern Recognition (CVPR)}, pages 21424--21433, 2024{\natexlab{a}}.

\bibitem[Zhang et~al.(2018)Zhang, Isola, Efros, Shechtman, and Wang]{8578166}
Richard Zhang, Phillip Isola, Alexei~A. Efros, Eli Shechtman, and Oliver Wang.
\newblock The unreasonable effectiveness of deep features as a perceptual metric.
\newblock In \emph{2018 IEEE/CVF Conference on Computer Vision and Pattern Recognition}, 2018.

\bibitem[Zhang et~al.(2025)Zhang, Chen, Xiong, Dai, Shen, and Xu]{zhang2025nest}
Xin Zhang, Anpei Chen, Jincheng Xiong, Pinxuan Dai, Yujun Shen, and Weiwei Xu.
\newblock Neural shell texture splatting: More details and fewer primitives.
\newblock In \emph{Proceedings of the IEEE/CVF International Conference on Computer Vision (ICCV)}, 2025.

\bibitem[Zhang et~al.(2024{\natexlab{b}})Zhang, Jia, Niu, and Yin]{Zhang2024GaussianSpaA}
Yangming Zhang, Wenqi Jia, Wei Niu, and Miao Yin.
\newblock Gaussianspa: An "optimizing-sparsifying" simplification framework for compact and high-quality 3d gaussian splatting.
\newblock \emph{2025 IEEE/CVF Conference on Computer Vision and Pattern Recognition (CVPR)}, pages 26673--26682, 2024{\natexlab{b}}.

\bibitem[Zhou and Ni(2025)]{Zhou2025PerceptualGSSP}
Hongbi Zhou and Zhangkai Ni.
\newblock Perceptual-gs: Scene-adaptive perceptual densification for gaussian splatting.
\newblock \emph{ArXiv}, abs/2506.12400, 2025.

\bibitem[Zhou et~al.(2025)Zhou, Xiong, Xia, Zhang, and Zhan]{Zhou2025GradientDirectionAwareDC}
Zheng Zhou, Yu-Jie Xiong, Chun-Ming Xia, Jia-Chen Zhang, and Hong-Jian Zhan.
\newblock Gradient-direction-aware density control for 3d gaussian splatting.
\newblock 2025.

\bibitem[Zixin~Zou(2023)]{Zou2023}
Yuanchen Guo Yangguang Li Ding Liang Yanpei Cao Songhai~Zhang Zixin~Zou, Zhipeng~Yu.
\newblock Triplane meets gaussian splatting: Fast and generalizable single-view 3d reconstruction with transformers.
\newblock \emph{arXiv preprint arXiv:2312.09147}, 2023.

\bibitem[Zwicker et~al.(2001)Zwicker, Pfister, van Baar, and Gross]{964490}
M. Zwicker, H. Pfister, J. van Baar, and M. Gross.
\newblock Ewa volume splatting.
\newblock In \emph{Proceedings Visualization, 2001. VIS '01.}, 2001.

\end{thebibliography}
}


\clearpage
\appendix

\beginsupplement

\FloatBarrier
\section*{Appendix}

This appendix introduces additional results (\secref{sec:additional_results}) and ablations (\secref{sec:additional_ablations}). We then discuss our reconstructed scene structure (\secref{sec:scene_structure}), possible failure cases (\secref{sec:failure_cases}), and provide visualizations for the different types of initializations mentioned in the main paper (\secref{sec:initializations}). Finally, we discuss the parallels between 3DGS and NeRF rendering, which allows training a radiance field using the rendering from 3DGS (\secref{subsec:mapping}).

\section{Detailed results}
\label{sec:additional_results}

\begin{table}

        \centering
        \vspace{2pt}
        \resizebox{\columnwidth}{!}{
        \begin{tabular}{l|ccccc}
        \toprule
         \textit{Ablation} & \textbf{PSNR} $\uparrow$ & \textbf{SSIM} $\uparrow$ & \textbf{LPIPS} $\downarrow$ & \textbf{FPS} $\uparrow$ & \# Gaussians\\
        \midrule
        Ours ($\beta=0$) & 29.12 & 0.873 & 0.183 & 185 & \multirow{3}{*}{542k} \\
        MCMC (SfM) & 28.86 & 0.865 & 0.209 & 136 &  \\
        EDGS & 28.87 & 0.868 & 0.187 & 194 \\
        \midrule
        Ours ($\beta=0.01$) & 29.25 & 0.877 & 0.174 & 169 & \multirow{3}{*}{674k}  \\
        MCMC (SfM) & 29.06 & 0.870 & 0.199 & 119 & \\
        EDGS & 28.97 & 0.873 & 0.178 & 167 \\
        \midrule
        Ours ($\beta=0.02$) & 29.34 & 0.880 & 0.166 & 146 & \multirow{3}{*}{942k} \\
        MCMC (SfM) & 29.15 & 0.876 & 0.189 & 101 &  \\
        EDGS & 29.10 & 0.878 & 0.167 & 134 \\
        \midrule
        Ours ($\beta=0.04$) & 29.38 & 0.881 & 0.161 & 105 & \multirow{3}{*}{1.66M}  \\
        MCMC (SfM) & 29.32 & 0.882 & 0.174 & 77 \\
        EDGS & 29.19 & 0.881 & 0.159 & 102 \\
        \midrule
        Ours & 29.35 & 0.879 & 0.159 & 80 & \multirow{4}{*}{2.57M}  \\
        MCMC (SfM) & 29.56 & 0.887 & 0.164 & 55 & \\
        EDGS & 29.37 & 0.884 & 0.153 & 67 \\
        3DGS & 29.03 & 0.870 & 0.184 & 69 & \\
        \bottomrule
        \end{tabular}
        }
        \caption{\textbf{No-budget Mip-NeRF360.} Comparison of reconstruction quality without a fixed limit on the number of primitives. For each cell, our method is run with a different $\beta$ parameter, which determines the number of Gaussians generated, and the other methods are limited to this number. In the last cell, the number of Gaussians is set to match the amount produced by 3DGS, with all other methods constrained accordingly. Results are averaged over the Mip-NeRF360~\cite{barron2022mipnerf360} dataset.
}
        \label{table:no_limit_benchmark_appendix_mip}   
\end{table}

\begin{table}
        \centering
        \vspace{2pt}
        \resizebox{\columnwidth}{!}{
        \begin{tabular}{l|ccccc}
        \toprule
         \textit{Ablation} & \textbf{PSNR} $\uparrow$ & \textbf{SSIM} $\uparrow$ & \textbf{LPIPS} $\downarrow$ & \textbf{FPS} $\uparrow$ & \# Gaussians\\
        \midrule
        Ours ($\beta=0$) & 28.54 & 0.879 & 0.195 & 253 & \multirow{3}{*}{352k} \\
        MCMC (SfM) & 28.37 & 0.877 & 0.207 & 199 &  \\
        EDGS & 28.62 & 0.885 & 0.183 & 219 \\
        \midrule
        Ours ($\beta=0.01$) & 29.02 & 0.890 & 0.176 & 237 & \multirow{3}{*}{438k} \\
        MCMC (SfM) & 28.55 & 0.882 & 0.199 & 175 &  \\
        EDGS & 28.84 & 0.890 & 0.173 & 211 \\
        \midrule
        Ours ($\beta=0.02$) & 29.19 & 0.894 & 0.166 & 217 & \multirow{3}{*}{581k} \\
        MCMC (SfM) & 28.79 & 0.888 & 0.190 & 157 &  \\
        EDGS & 29.07 & 0.896 & 0.163 & 188 \\
        \midrule
        Ours ($\beta=0.04$) & 29.32 & 0.898 & 0.157 & 116 & \multirow{3}{*}{927k} \\
        MCMC (SfM) & 29.05 & 0.895 & 0.177 & 105 \\
        EDGS & 29.35 & 0.902 & 0.151 & 134 \\
        \midrule
        Ours & 29.60 & 0.904 & 0.144 & 114 & \multirow{4}{*}{1.75M} \\
        MCMC & 29.46 & 0.904 & 0.157 & 89 &  \\
        EDGS & 29.85 & 0.909 & 0.137 & 99 \\
        3DGS & 29.30 & 0.896 & 0.171 & 88 &  \\
        \bottomrule
        \end{tabular}
        }
        \caption{\textbf{No-budget OMMO.} Comparison of reconstruction quality without a fixed limit on the number of primitives. For each cell, our method is run with a different $\beta$ parameter, which determines the number of Gaussians generated, and the other methods are limited to this number. In the last cell, the number of Gaussians is set to match the amount produced by 3DGS, with all other methods constrained accordingly. Results are averaged over the OMMO~\cite{lu2023largescaleoutdoormultimodaldataset} dataset.
}
        \label{table:no_limit_benchmark_appendix_ommo}
\end{table}

\begin{table*}
\centering
\resizebox{\linewidth}{!}{%
\begin{threeparttable}
\begin{tabular}{c|ccc|ccc|ccc|ccc}
\toprule
 & \multicolumn{3}{c|}{Mip-NeRF360 \cite{barron2022mipnerf360}} & \multicolumn{3}{c|}{OMMO \cite{lu2023largescaleoutdoormultimodaldataset}} & \multicolumn{3}{c|}{Tanks \& Temples \cite{Knapitsch2017}} & \multicolumn{3}{c}{DeepBlending \cite{DeepBlending2018}} \\
 & \textbf{PSNR $\uparrow$} & \textbf{SSIM $\uparrow$} & \textbf{LPIPS $\downarrow$} & \textbf{PSNR $\uparrow$} & \textbf{SSIM $\uparrow$} & \textbf{LPIPS $\downarrow$} & \textbf{PSNR $\uparrow$} & \textbf{SSIM $\uparrow$} & \textbf{LPIPS $\downarrow$} & \textbf{PSNR $\uparrow$} & \textbf{SSIM $\uparrow$} & \textbf{LPIPS $\downarrow$} \\
\midrule
\multicolumn{13}{c}{Number of Gaussians limited to $1$M} \\
\midrule
3DGS~\cite{kerbl3Dgaussians} (SfM init.) & 28.71 & 0.846 & 0.222 & \colorbox{tabthird}{29.40} & 0.894 & 0.178 & 23.60 & 0.826 & 0.245 & 28.93 & 0.881 & 0.292 \\
Foroutan et al.~\cite{Foroutan2024EvaluatingAT}$^\dagger$ & 29.22 & 0.876 & 0.177 & 29.32 & \colorbox{tabthird}{0.896} & \colorbox{tabthird}{0.163} & \colorbox{tabsecond}{23.75} & 0.829 & \colorbox{tabthird}{0.230} & 29.60 & 0.886 & 0.282  \\
MCMC~\cite{kheradmand20243d} (rand. init.) & 28.98 & 0.865 & 0.204 & 28.83 & 0.888 & 0.185 & 23.41 & 0.824 & 0.249 & 28.94 & 0.877 & 0.303 \\
MCMC~\cite{kheradmand20243d} (SfM init.) & 29.23 & 0.875 & 0.190 & 29.18 & 0.895 & 0.174 & \colorbox{tabfirst}{23.93} & \colorbox{tabsecond}{0.836} & 0.233 & \colorbox{tabthird}{29.67} & \colorbox{tabthird}{0.887}  & 0.288 \\
MCMC~\cite{kheradmand20243d} (iNGP init.) & \colorbox{tabsecond}{29.32} & \colorbox{tabsecond}{0.879} & \colorbox{tabsecond}{0.173} & 28.76 & 0.887 & 0.181 & 23.37 & 0.804 & 0.268 & \colorbox{tabfirst}{29.87} & \colorbox{tabfirst}{0.892} & \colorbox{tabthird}{0.276} \\
GaussianPro~\cite{Cheng2024ICML_GaussianPro_3D_Gaussian} & 28.56 & 0.845 & 0.226 & 28.79 & 0.885 & 0.189 & 23.01 & 0.818 & 0.256 & 29.27 & \colorbox{tabthird}{0.887} & 0.291 \\
Perceptual-GS~\cite{Zhou2025PerceptualGSSP} & \colorbox{tabthird}{29.27} & 0.876 & \colorbox{tabthird}{0.176} & 29.00 & 0.888 & 0.179 & 23.26 & 0.828 & 0.240 & 29.41 & \colorbox{tabsecond}{0.889} & 0.286 \\
EDGS~\cite{Kotovenko2025ARXIV_EDGS_Eliminating_Densification} & 29.18 & \colorbox{tabthird}{0.877} & \colorbox{tabfirst}{0.168} & \colorbox{tabfirst}{29.58} & \colorbox{tabfirst}{0.903} & \colorbox{tabfirst}{0.149} & 23.66 & \colorbox{tabfirst}{0.842} & \colorbox{tabfirst}{0.200} & 29.43 & \colorbox{tabsecond}{0.889} & \colorbox{tabfirst}{0.271} \\
Ours & \colorbox{tabfirst}{29.37} & \colorbox{tabfirst}{0.880} & \colorbox{tabfirst}{0.168} & \colorbox{tabsecond}{29.44} & \colorbox{tabsecond}{0.899} & \colorbox{tabsecond}{0.154} & \colorbox{tabthird}{23.70} & \colorbox{tabthird}{0.835} & \colorbox{tabsecond}{0.207} & \colorbox{tabsecond}{29.69} & \colorbox{tabsecond}{0.889} & \colorbox{tabsecond}{0.273} \\

\bottomrule
\end{tabular}
\begin{tablenotes}
\small
\item[$\dagger$] Original code was not publicly available. Our implementation uses iNGP initialization and does not include the additional depth-based loss.
\end{tablenotes}
\end{threeparttable}
}
\caption{\textbf{Quantitative results} with a Gaussian number limit of 1M. We highlight the \colorbox{tabfirst}{best}, \colorbox{tabsecond}{second best} and \colorbox{tabthird}{third best} results among methods with comparable numbers of Gaussians.}
\label{tab:1M_results}
\end{table*}

\begin{figure}
        \centering
            \begin{tabular}{@{}c@{\hskip 3pt}c@{}}
                \includegraphics[width=0.5\columnwidth, keepaspectratio, trim={0pt 85pt 0pt 0pt}, clip]{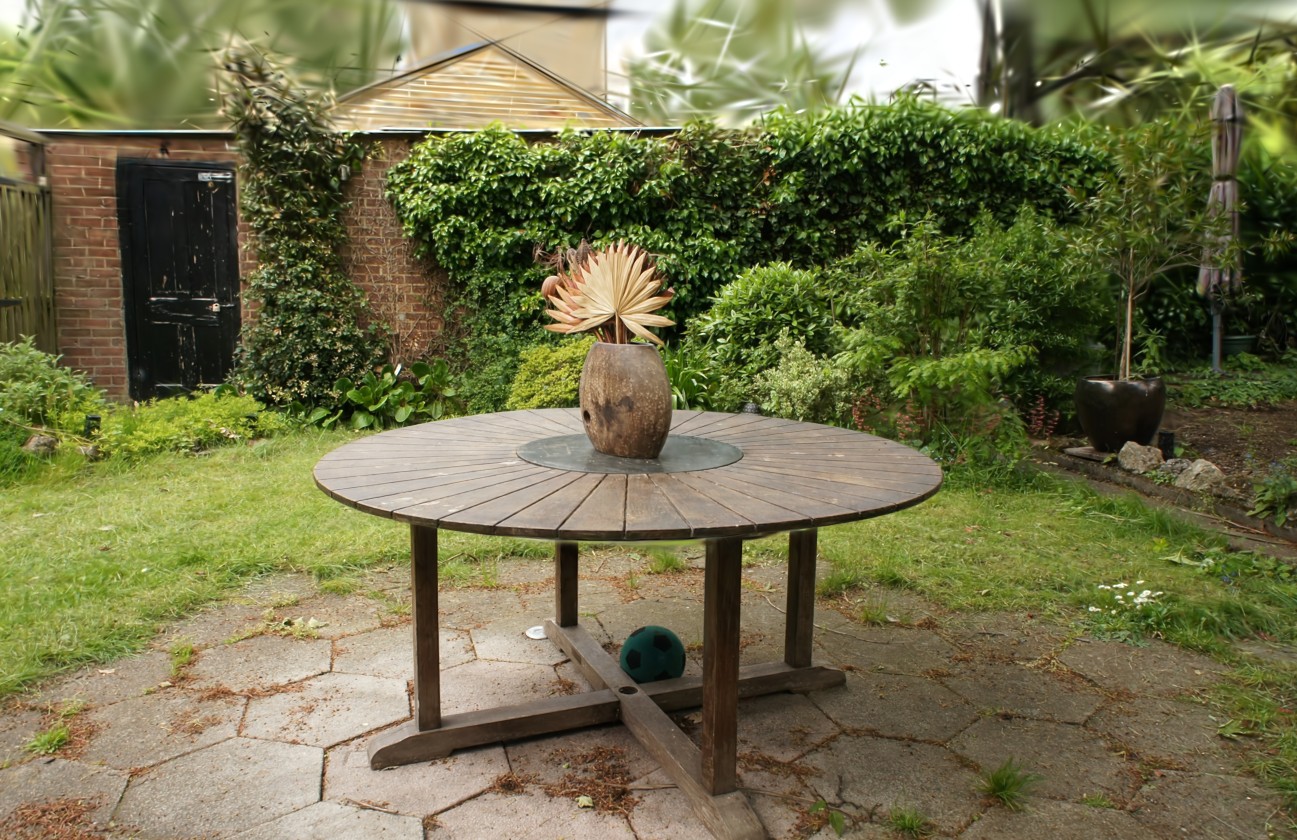} &
                \includegraphics[width=0.5\columnwidth, keepaspectratio, trim={0pt 112pt 0pt 0pt}, clip]{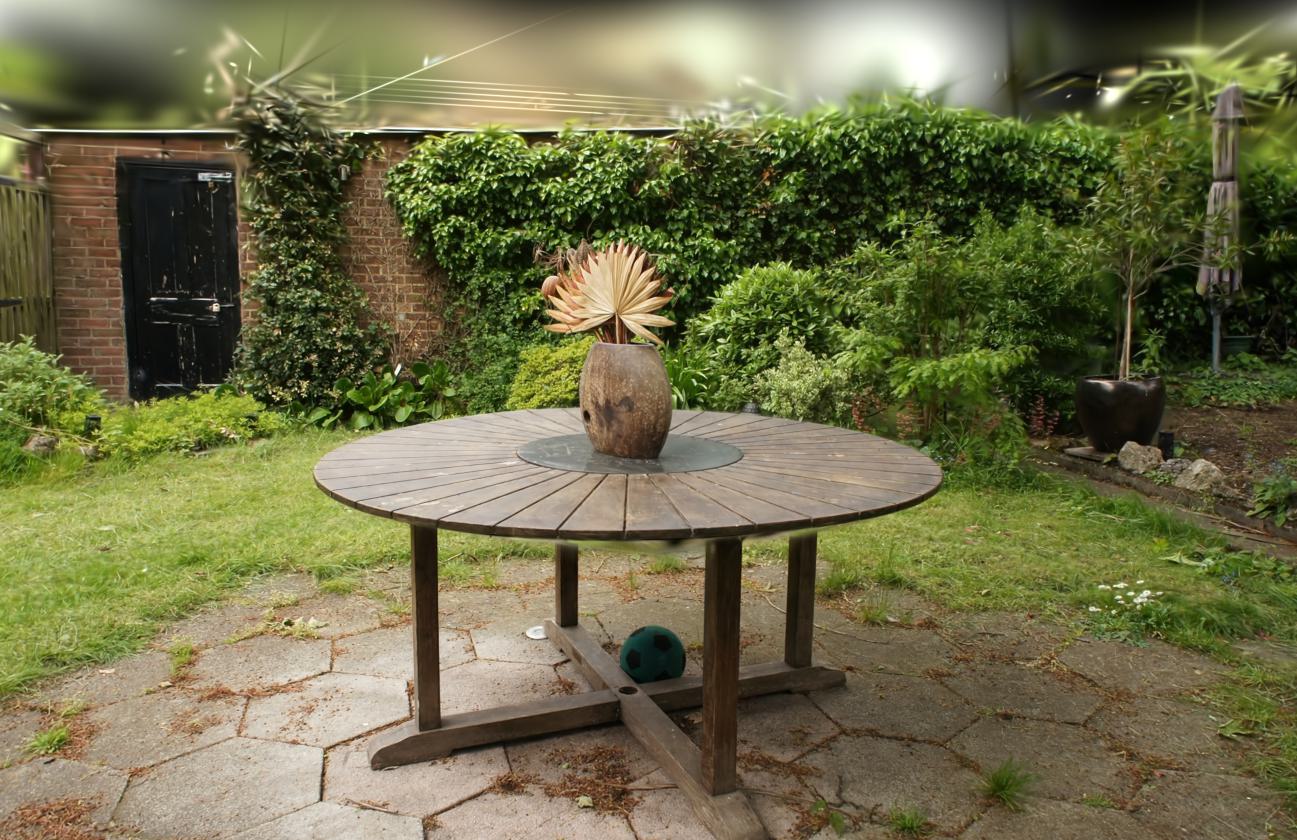} \\
                \includegraphics[width=0.5\columnwidth, keepaspectratio, trim={0pt 300pt 900pt 200pt}, clip]{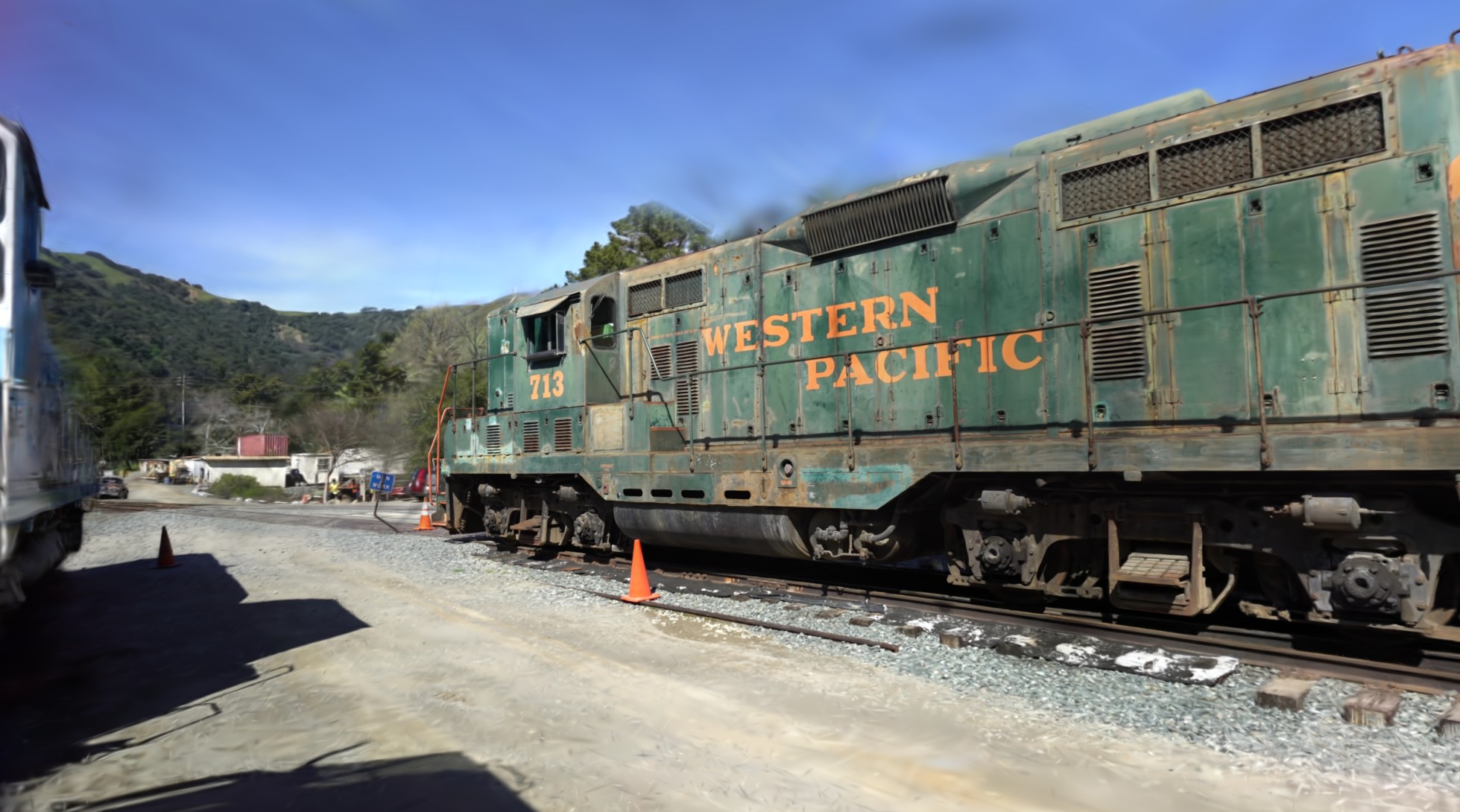} &
                \includegraphics[width=0.5\columnwidth, keepaspectratio, trim={0pt 300pt 900pt 200pt}, clip]{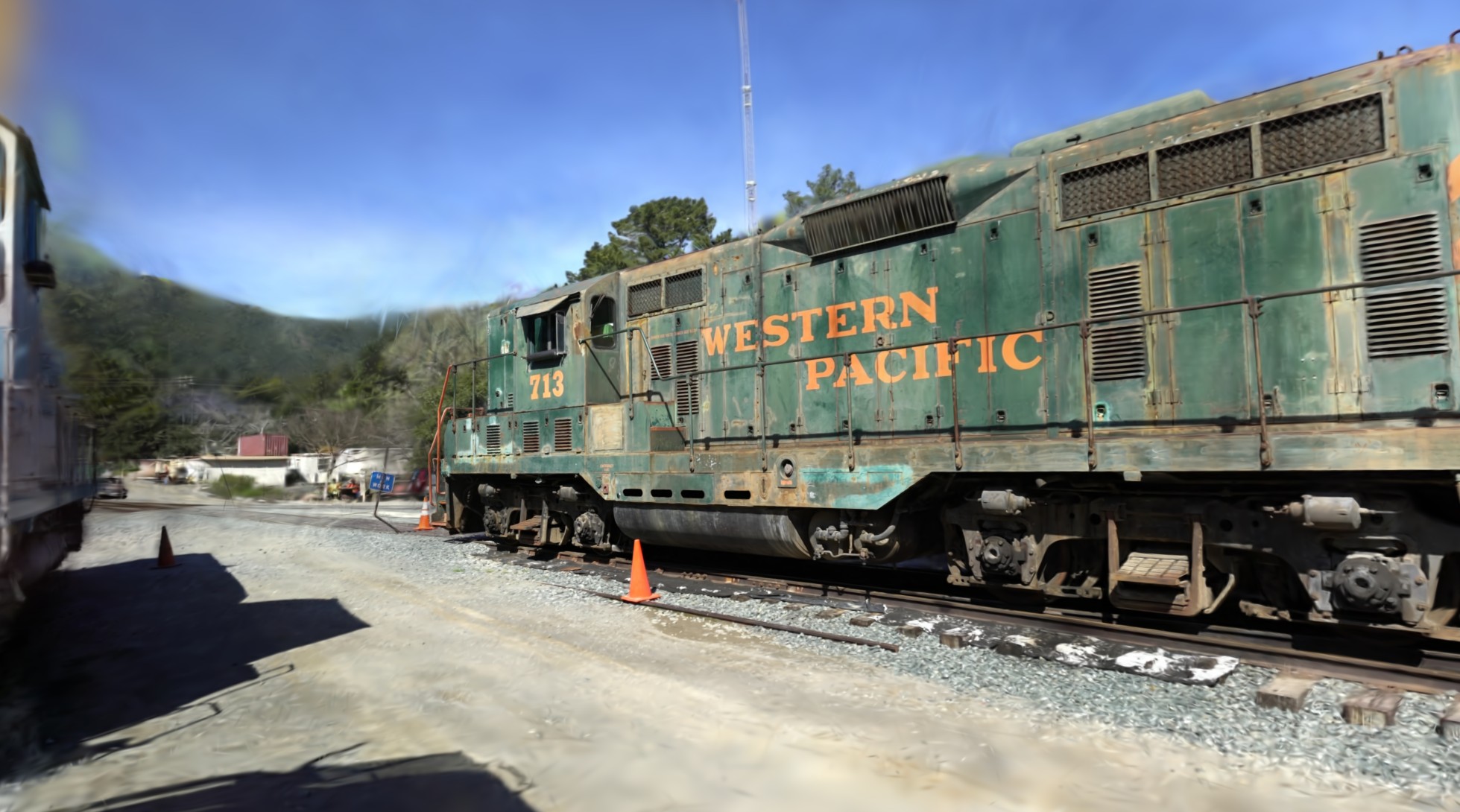} \\
                Ours & Mini-Splatting2~\cite{Fang2024MiniSplatting2B3} \\
            \end{tabular}
        \caption{\textbf{Qualitative comparison} between our approach and Mini-Splatting2~\cite{Fang2024MiniSplatting2B3} on the \texttt{garden} scene from Mip-NeRF360~\cite{barron2022mipnerf360} and on the \texttt{train} scene from Tanks \& Temples \cite{Knapitsch2017}.}
        \label{fig:mini_splatting_comparison}
\end{figure}

\tabref{table:no_limit_benchmark_appendix_mip} and \tabref{table:no_limit_benchmark_appendix_ommo} present results for the no-budget scenario, which were used for plots in \figref{fig:lpips_psnr_fps_plots_combined}. We use various values of the $\beta$ parameter without specifying a budget for our method, except for the last cell, which uses the budget set by the number of Gaussians generated by 3D Gaussian Splatting. For each cell, we also compare MCMC~\cite{kheradmand20243d} and EDGS~\cite{Kotovenko2025ARXIV_EDGS_Eliminating_Densification} where they are set to match the number of Gaussians produced in each of the cells.
The comparisons show that our method is able to produce high-quality results even without specifying a budget, instead adjusting the number of primitives based on the scene complexity, while still remaining sparse in the number of primitives. Notably, even when setting $\beta = 0$, which effectively disables densification, our method still performs well due to a dense initialization and effective filtering of unnecessary primitives enforced by the opacity penalty, consistent with findings in  ~\cite{Kotovenko2025ARXIV_EDGS_Eliminating_Densification}. However, this configuration does not allow explicit control over the number of primitives and may limit the achievable reconstruction quality.

If an even lower number of Gaussians is desired while still using a no-budget scenario, the number of initialized Gaussians can be reduced, effectively lowering the number of primitives generated at the end.

We show in~\tabref{tab:1M_results} the results on the selected benchmarks for the budget of 1M primitives. Additionally, we present per-scene results for 100k Gaussians in \tabref{table:expanded_all_results_100k}, 500k Gaussians in \tabref{table:expanded_all_results_500k}, 1M Gaussians in \tabref{table:expanded_all_results_1M}, and 2M Gaussians in \tabref{table:expanded_all_results_2M}. We also show additional qualitative results with Mini-Splatting2~\cite{Fang2024MiniSplatting2B3} in \figref{fig:mini_splatting_comparison}.

\section{Additional ablations and comparisons}
\label{sec:additional_ablations}

\begin{table}
        \centering
        \vspace{2pt}
        \resizebox{\columnwidth}{!}{
        \begin{tabular}{l|ccccc}
        \toprule
          & Ours & 3DGS (SfM) & MCMC (SfM) & EDGS & Perceptual-GS \\
        \midrule
        
        \# Memory (MiB) & \colorbox{tabthird}{9545} & \colorbox{tabsecond}{9049} & \colorbox{tabfirst}{8671} & 14517 & 12403 \\
        \bottomrule
        \end{tabular}
        }
        \caption{\textbf{Peak GPU memory usage} on the \texttt{15} scene from the OMMO dataset~\cite{lu2023largescaleoutdoormultimodaldataset} on the maximum budget of 500k primitives. We highlight the \colorbox{tabfirst}{best}, \colorbox{tabsecond}{second best} and \colorbox{tabthird}{third best} results among all.}
        \label{table:memory_usage}
\end{table}

\begin{table}
    \centering

    \vspace{2pt}
    \resizebox{\columnwidth}{!}{%
    \begin{tabular}{l|ccccc}
    \toprule
      & \textbf{PSNR} $\uparrow$ & \textbf{SSIM} $\uparrow$ & \textbf{LPIPS} $\downarrow$ & \textbf{FPS} $\uparrow$ & \textbf{Train (min)} $\downarrow$  \\
    \midrule
    
    ConeGS & \colorbox{tabsecond}{29.14} & \colorbox{tabsecond}{0.892} & \colorbox{tabsecond}{0.170} & \colorbox{tabfirst}{217} & \colorbox{tabfirst}{25} \\
    \makecell{ConeGS (iNGP retrain)} & \colorbox{tabfirst}{29.27} & \colorbox{tabfirst}{0.893} & \colorbox{tabfirst}{0.168} & \colorbox{tabsecond}{212} & \colorbox{tabsecond}{29} \\
    \bottomrule
    \end{tabular}
    }
    \caption{\textbf{Ablation on additional iNGP training} on the OMMO~\cite{lu2023largescaleoutdoormultimodaldataset} dataset with 500k primitives, evaluating the effect of continuing training iNGP in parallel with the full 3DGS optimization. We highlight the \colorbox{tabfirst}{best}, \colorbox{tabsecond}{second best} and \colorbox{tabthird}{third best} results among all.
    }
    \label{table:ingp_retraining_ommo}
\end{table}

\begin{figure}
        \centering
        \resizebox{1\columnwidth}{!}{%
            \input{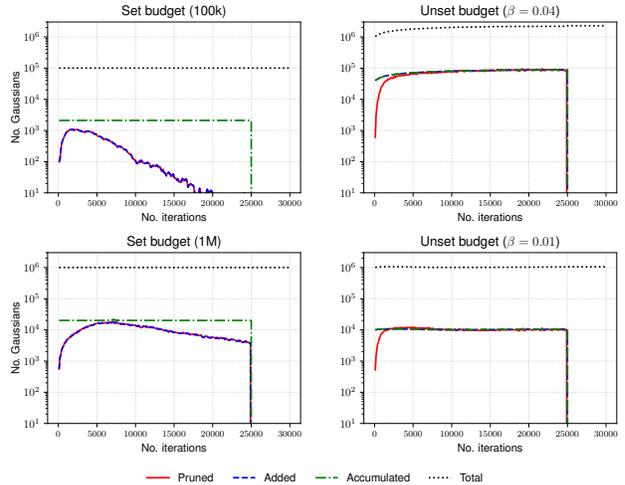}
        }
        \caption{\textbf{Number of primitives} pruned, added, accumulated, as well as the total number, during each densification, on different budget scenarios. Results obtained from the \texttt{garden} scene from Mip-NeRF360~\cite{barron2022mipnerf360}.
        }
        \label{fig:number_primitives_plots}
\end{figure}

\begin{table*}
\centering
\resizebox{\linewidth}{!}{%
\begin{tabular}{c|ccc|ccc|ccc|ccc}
\toprule
 & \multicolumn{3}{c|}{Mip-NeRF360 \cite{barron2022mipnerf360}} & \multicolumn{3}{c|}{OMMO \cite{lu2023largescaleoutdoormultimodaldataset}} & \multicolumn{3}{c|}{Tanks \& Temples \cite{Knapitsch2017}} & \multicolumn{3}{c}{DeepBlending \cite{DeepBlending2018}} \\
 & \textbf{PSNR $\uparrow$} & \textbf{SSIM $\uparrow$} & \textbf{LPIPS $\downarrow$} & \textbf{PSNR $\uparrow$} & \textbf{SSIM $\uparrow$} & \textbf{LPIPS $\downarrow$} & \textbf{PSNR $\uparrow$} & \textbf{SSIM $\uparrow$} & \textbf{LPIPS $\downarrow$} & \textbf{PSNR $\uparrow$} & \textbf{SSIM $\uparrow$} & \textbf{LPIPS $\downarrow$} \\
\midrule
\multicolumn{13}{c}{Number of Gaussians limited to $1$M} \\
\midrule
%
Ours & \colorbox{tabfirst}{29.37} & \colorbox{tabfirst}{0.880} & \colorbox{tabfirst}{0.168} & \colorbox{tabsecond}{29.44} & \colorbox{tabsecond}{0.899} & \colorbox{tabsecond}{0.154} & \colorbox{tabsecond}{23.70} & \colorbox{tabsecond}{0.835} & \colorbox{tabfirst}{0.207} & \colorbox{tabsecond}{29.69} & \colorbox{tabfirst}{0.889} & \colorbox{tabfirst}{0.273} \\
Ours, iNGP training during 3DGS & \colorbox{tabsecond}{29.25} & \colorbox{tabsecond}{0.879} & \colorbox{tabfirst}{0.168} & \colorbox{tabfirst}{29.54} & \colorbox{tabfirst}{0.900} & \colorbox{tabfirst}{0.152} & \colorbox{tabfirst}{23.72} & \colorbox{tabfirst}{0.836} & \colorbox{tabfirst}{0.207} & \colorbox{tabfirst}{29.73} & \colorbox{tabfirst}{0.889} & \colorbox{tabfirst}{0.273} \\

\bottomrule
\end{tabular}
}
\caption{\textbf{Ablation on additional iNGP training} for the budget of 1M primitives. We highlight the \colorbox{tabfirst}{best}, \colorbox{tabsecond}{second best} and \colorbox{tabthird}{third best} results among methods with comparable numbers of Gaussians.}
\label{tab:1M_ablation_with_training}
\end{table*}

We evaluate the peak GPU memory usage during optimization for several methods in \tabref{table:memory_usage}. The results show that our method is very close to 3DGS~\cite{kerbl3Dgaussians} and MCMC~\cite{kheradmand20243d} in terms of memory consumption, while remaining considerably lower than EDGS~\cite{Kotovenko2025ARXIV_EDGS_Eliminating_Densification} and Perceptual-GS~\cite{Zhou2025PerceptualGSSP}. This efficiency allows our method to run on a wider range of GPUs, making it more accessible to devices with limited memory.

We expand on ablation \textbf{(c)} from the main paper, where iNGP continues training in parallel with the full 3DGS optimization. Since iNGP training, like densification, is guided by the L1 error from 3DGS, the iNGP model can better focus on regions where 3DGS reconstruction may fall short, potentially improving densification. The original ablation was tested on a budget of 100k primitives, which may not fully reveal this effect. To explore further, we run experiments with larger budgets. On the 500k budget for challenging OMMO~\cite{lu2023largescaleoutdoormultimodaldataset} scenes (\tabref{table:ingp_retraining_ommo}), we observe additional performance gains, though at the cost of longer training time. We also test this setup with a budget of 1M primitives across all datasets (\tabref{tab:1M_ablation_with_training}), finding minimal improvements on difficult scenes and slight decreases on Mip-NeRF360, which may be caused by overfitting to certain areas.

\figref{fig:number_primitives_plots} illustrates the number of primitives added and removed during each densification and pruning step. It also tracks the accumulation buffer of primitives. When no budget is specified, all accumulated primitives are added to the scene (see~\eqnref{eq:accumulated_no_budget}). Under a fixed budget, however, not all of them are used. This is because accumulation happens every iteration and must remain available for pruning and densification, while the exact number that can be added is unknown beforehand. As a result, the system accumulates more primitives than are usually required to keep the total count close to the budget after pruning (see~\eqnref{eq:accumulated_budget}).

\begin{figure}
        \centering
        \resizebox{1\columnwidth}{!}{%
               \input{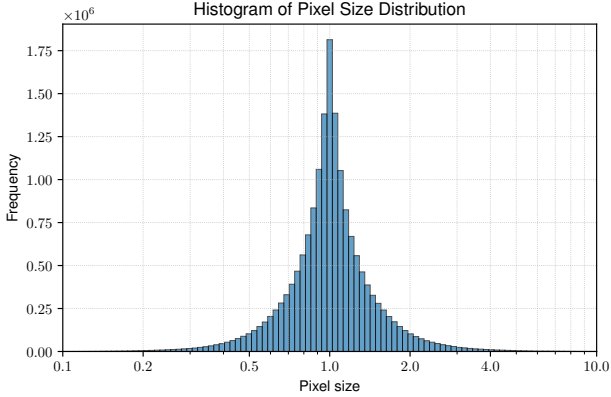}
           }
        \caption{\textbf{The histogram of perceived image-space size} of pixel-sized Gaussians rendered from different viewpoints. Size expressed in pixel widths. Analysis doesn't include the low-pass filter from 3DGS rasterization.}
        \label{fig:histograms_pixel_size}
\end{figure}

The primitive size is defined to be approximately one pixel from a single viewpoint. From other viewpoints, this size is not exactly one pixel, but should remain close. To confirm this, we measure the size of newly added pixel-sized primitives from multiple viewpoints and report the results in \figref{fig:histograms_pixel_size}. The distribution shows that the apparent size remains near one pixel, with very few cases above four pixels or below 0.2 pixels.

\section{Scene structure}
\label{sec:scene_structure}
\begin{figure}
        \centering
        \resizebox{1\columnwidth}{!}{%
               \input{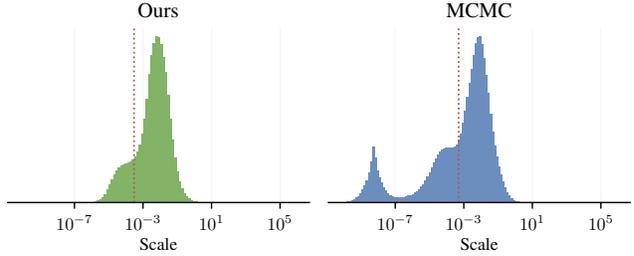}
           }
        \caption{\textbf{Histograms of the Gaussian scaling values} for our method and MCMC with SfM initialization on 2M Gaussians. The red lines indicate the minimum scaling required for a Gaussian to cover at least one pixel, disregarding the low-pass filter.}
        \label{fig:scaling_collapse}
\end{figure}

\begin{figure}
    \centering

    \begin{minipage}{0.495\columnwidth}
        \includegraphics[width=\columnwidth]{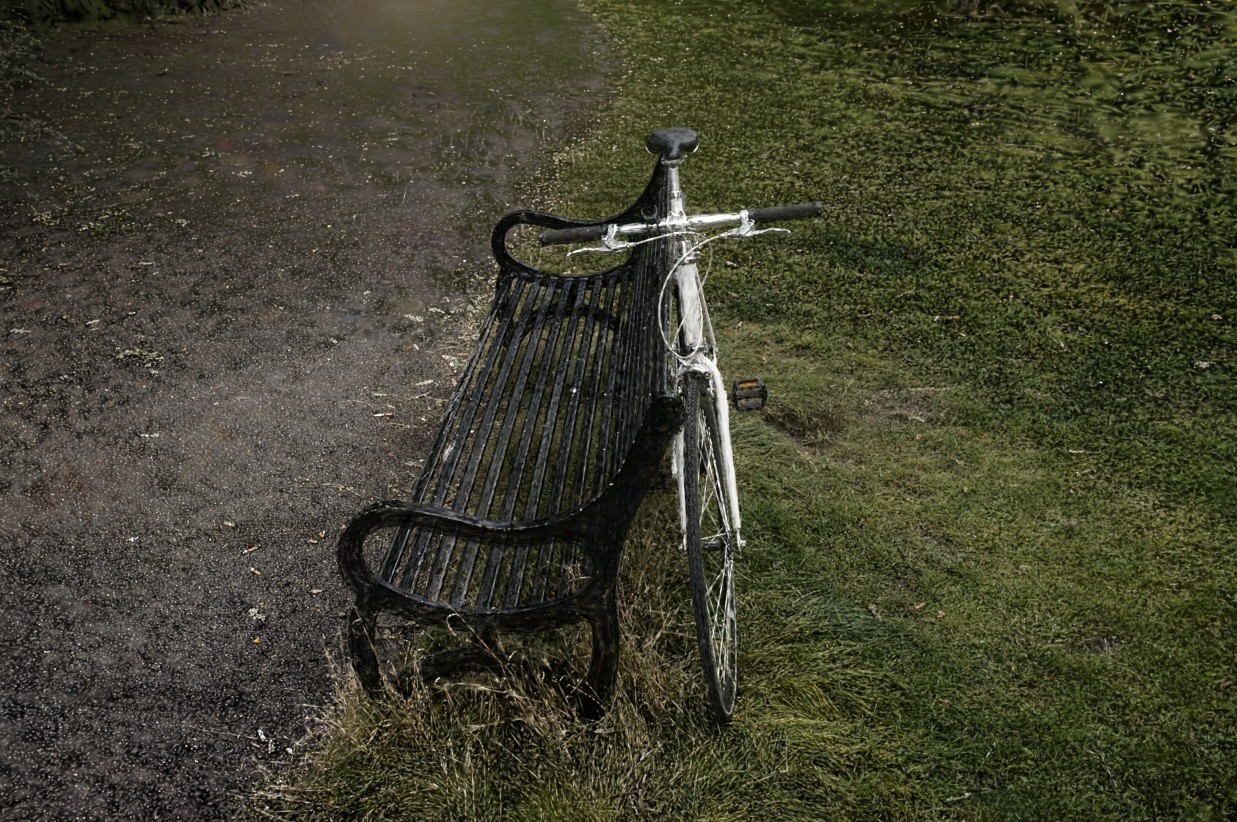}
    \end{minipage}
    \hfill
    \begin{minipage}{0.495\columnwidth}
        \includegraphics[width=\columnwidth]{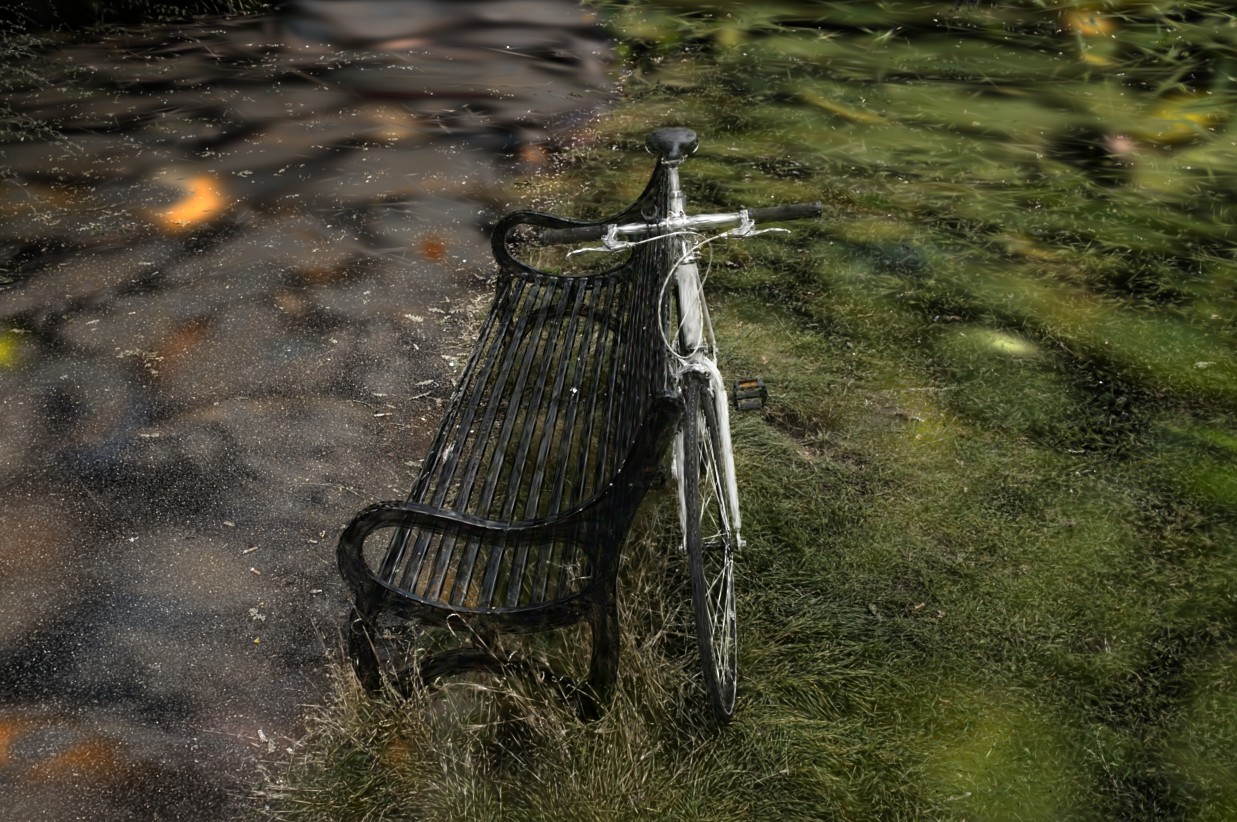}
    \end{minipage}

    \vspace{0.1em}

    \begin{minipage}{0.495\columnwidth}
        \includegraphics[width=\columnwidth]{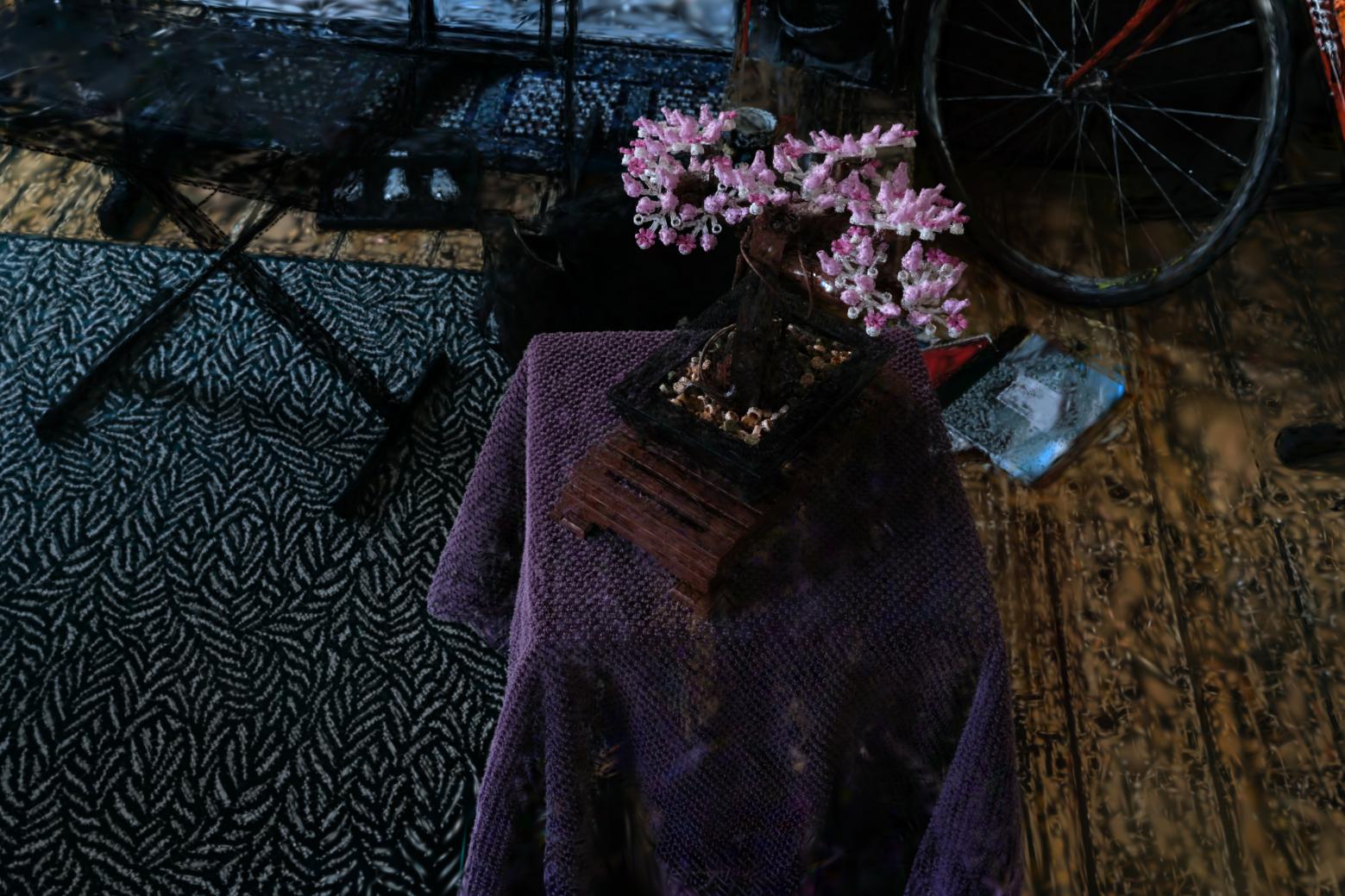}
    \end{minipage}
    \hfill
    \begin{minipage}{0.495\columnwidth}
        \includegraphics[width=\columnwidth]{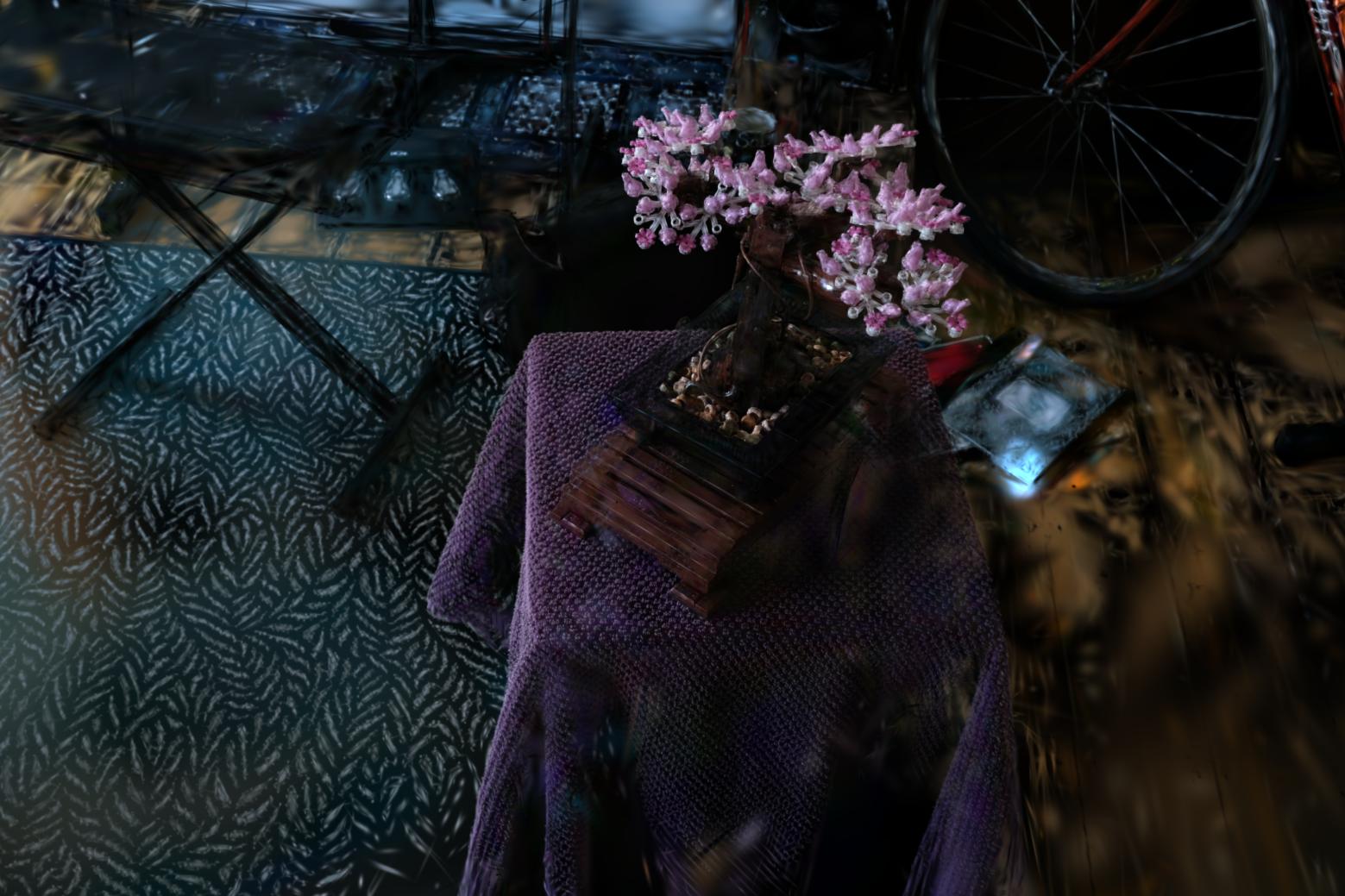}
    \end{minipage}

    \vspace{0.1em}

    \begin{minipage}{0.495\columnwidth}
        \includegraphics[width=\columnwidth]{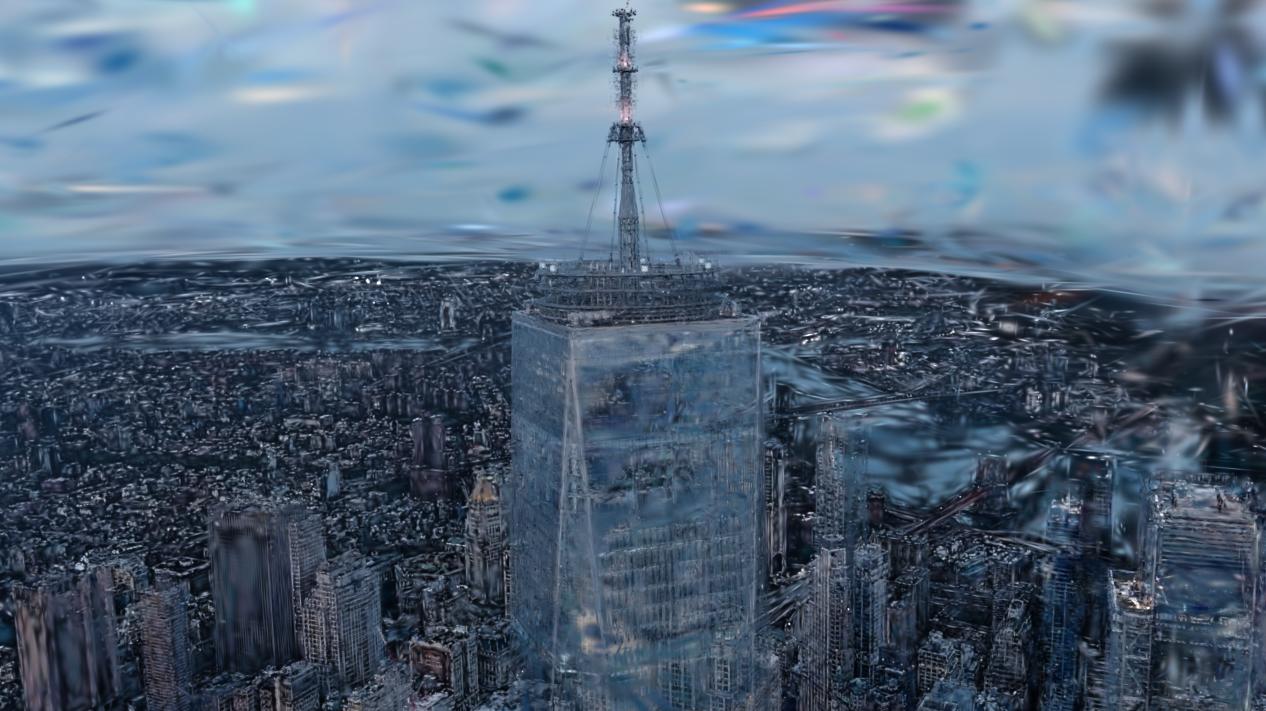}
    \end{minipage}
    \hfill
    \begin{minipage}{0.495\columnwidth}
        \includegraphics[width=\columnwidth]{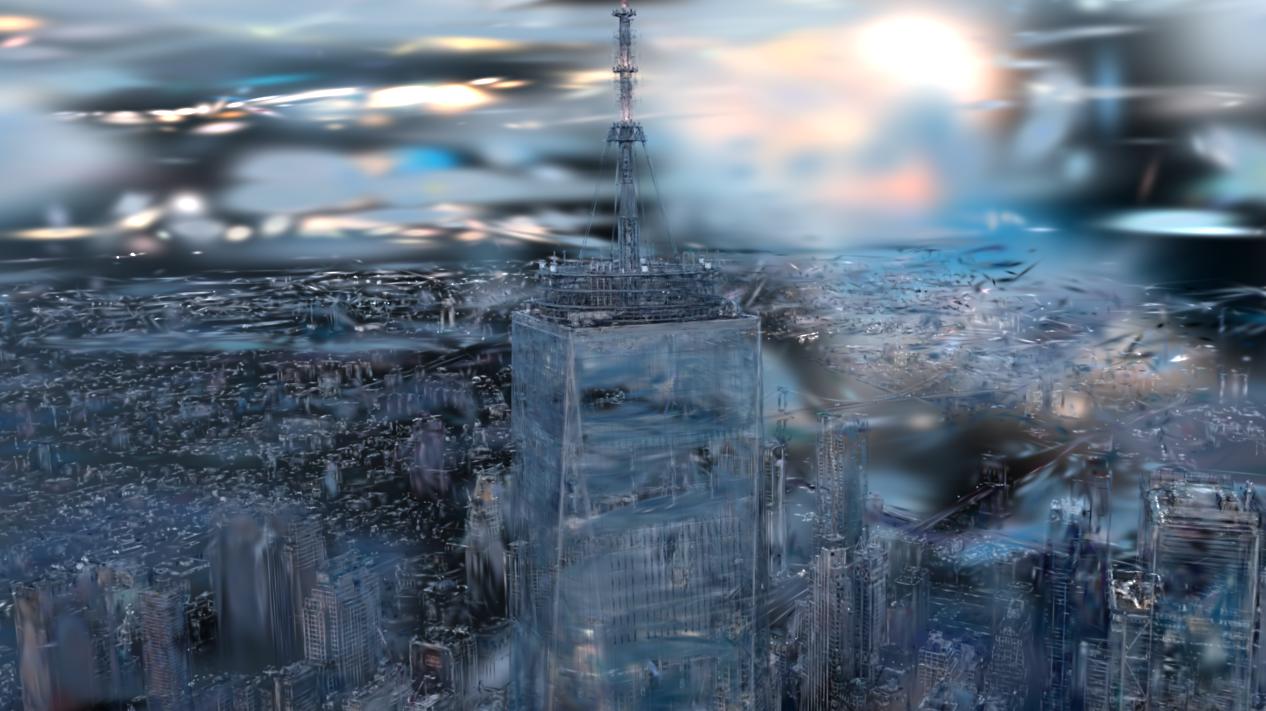}
    \end{minipage}

    \vspace{0.5em}

    \begin{minipage}{0.495\columnwidth}
        \centering
        Ours
    \end{minipage}
    \hfill
    \begin{minipage}{0.495\columnwidth}
        \centering
        MCMC
    \end{minipage}

    \caption{\textbf{Shrunk Gaussians.} Visual comparison between the proposed method and MCMC, rendered using Gaussians scaled to half their original size. Both methods were trained with a limit of 1 million primitives on the \texttt{bicycle} and \texttt{bonsai} scenes from Mip-NeRF360~\cite{barron2022mipnerf360}, and the \texttt{10} scene from OMMO~\cite{lu2023largescaleoutdoormultimodaldataset}.
    }
    \label{fig:size_limit_comparison}
\end{figure}

\begin{table}
        \centering
        \vspace{2pt}
        \resizebox{\columnwidth}{!}{
        \begin{tabular}{l|cccc}
        \toprule
          & Ours & MCMC (SfM) & MCMC & 3DGS (SfM) \\
        \midrule
        
        \# Gaussians per pixel & \colorbox{tabfirst}{30.72} & 49.55 & \colorbox{tabthird}{48.45} & \colorbox{tabsecond}{34.74} \\
        \bottomrule
        \end{tabular}
        }
        \caption{\textbf{Blending analysis.} Mean number of Gaussians contributing to alpha blending per pixel, averaged across the Mip-NeRF360~\cite{barron2022mipnerf360} dataset using a budget of 1M Gaussians. We highlight the \colorbox{tabfirst}{best}, \colorbox{tabsecond}{second best} and \colorbox{tabthird}{third best} results among all.}
        \label{table:gaussians_per_pixel}
\end{table}

To confirm that our method produces reconstructions with desirable characteristics, such as a balanced distribution of scaling values and placement close to surfaces, which are often important for downstream tasks, we analyze the scenes created with our method in comparison to MCMC~\cite{kheradmand20243d}.

Although Gaussian Splatting with the MCMC densification strategy explores the scene effectively, it also has clear shortcomings. Its cloning strategy and scaling penalty often cause Gaussians to shrink strongly along certain dimensions. The low-pass filter used by 3DGS renderer can hide this effect visually, but it prevents Gaussians from expanding in those directions and interferes with tasks that depend on accurate scales, such as MCMC position error. As shown in \figref{fig:scaling_collapse}, many of its Gaussians shrink below the pixel radius, in contrast to the more balanced distribution produced by our method.

\figref{fig:size_limit_comparison} shows that our approach produces Gaussians that are more uniform in size and placed closer to object surfaces. This avoids an overreliance on oversized background primitives located far from the surface. This primitive distribution not only improves geometric alignment but also increases rendering speed by reducing blending and sorting overhead during rasterization. To support this hypothesis, \tabref{table:gaussians_per_pixel} reports the mean number of Gaussians blended per pixel. Our method consistently requires less blending compared to the selected benchmarks. The difference is especially large compared to MCMC, which blends over $60\%$ more Gaussians per pixel on average.

\section{Failure cases}
\label{sec:failure_cases}

\begin{figure}
    \centering
    \renewcommand{\arraystretch}{0.5} 
    \setlength{\tabcolsep}{1pt}     
    \resizebox{\linewidth}{!}{%
    \begin{tabular}{%
        >{\centering\arraybackslash}m{0.33\linewidth}
        >{\centering\arraybackslash}m{0.33\linewidth}
        >{\centering\arraybackslash}m{0.33\linewidth}
    }
         GT & ConeGS & iNGP \\[5pt]
        \adjincludegraphics[width=\linewidth, keepaspectratio, trim={600pt 0pt 0pt 0pt}, clip]{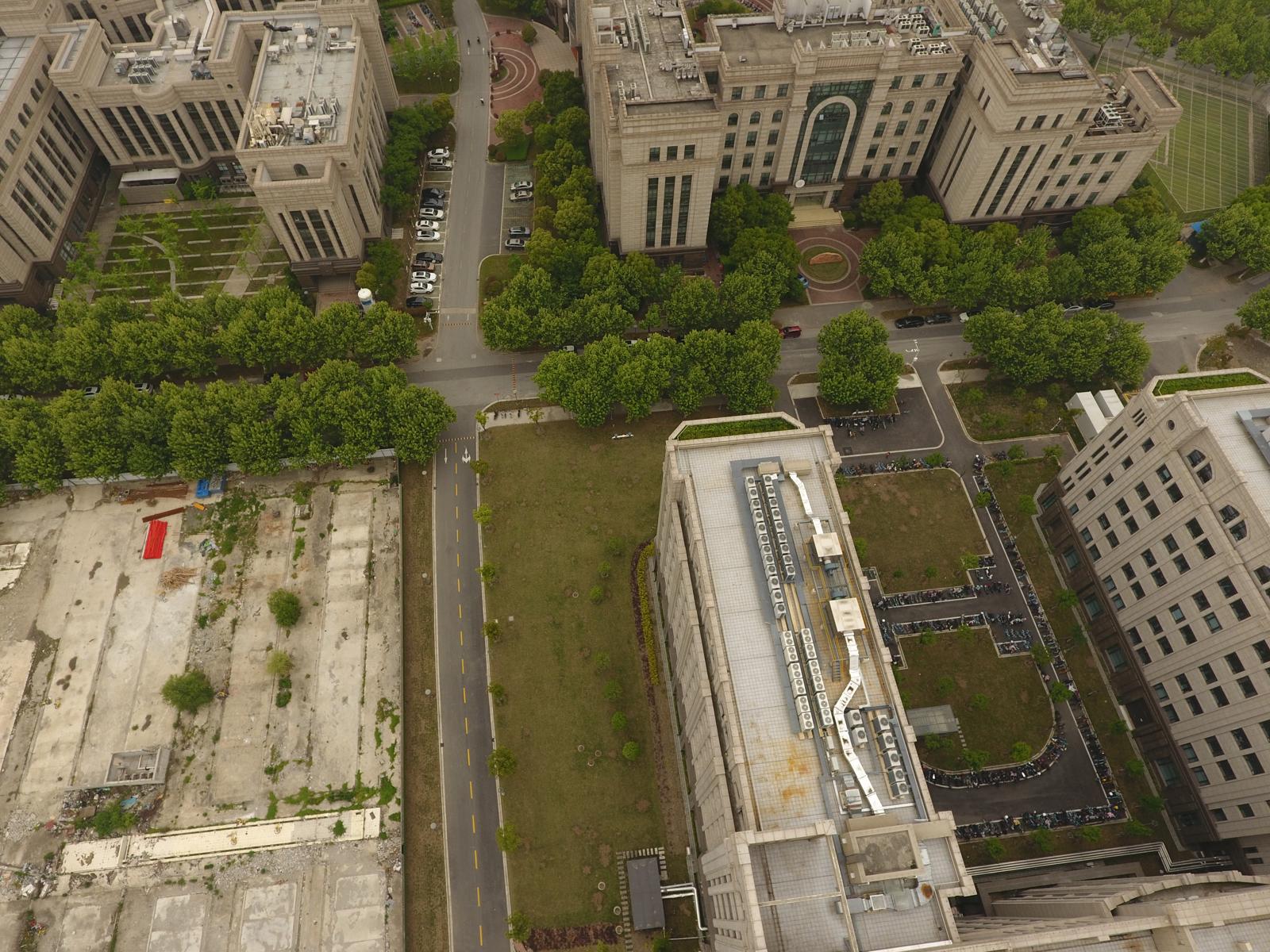} &
        \adjincludegraphics[width=\linewidth, keepaspectratio, trim={600pt 0pt 0pt 0pt}, clip]{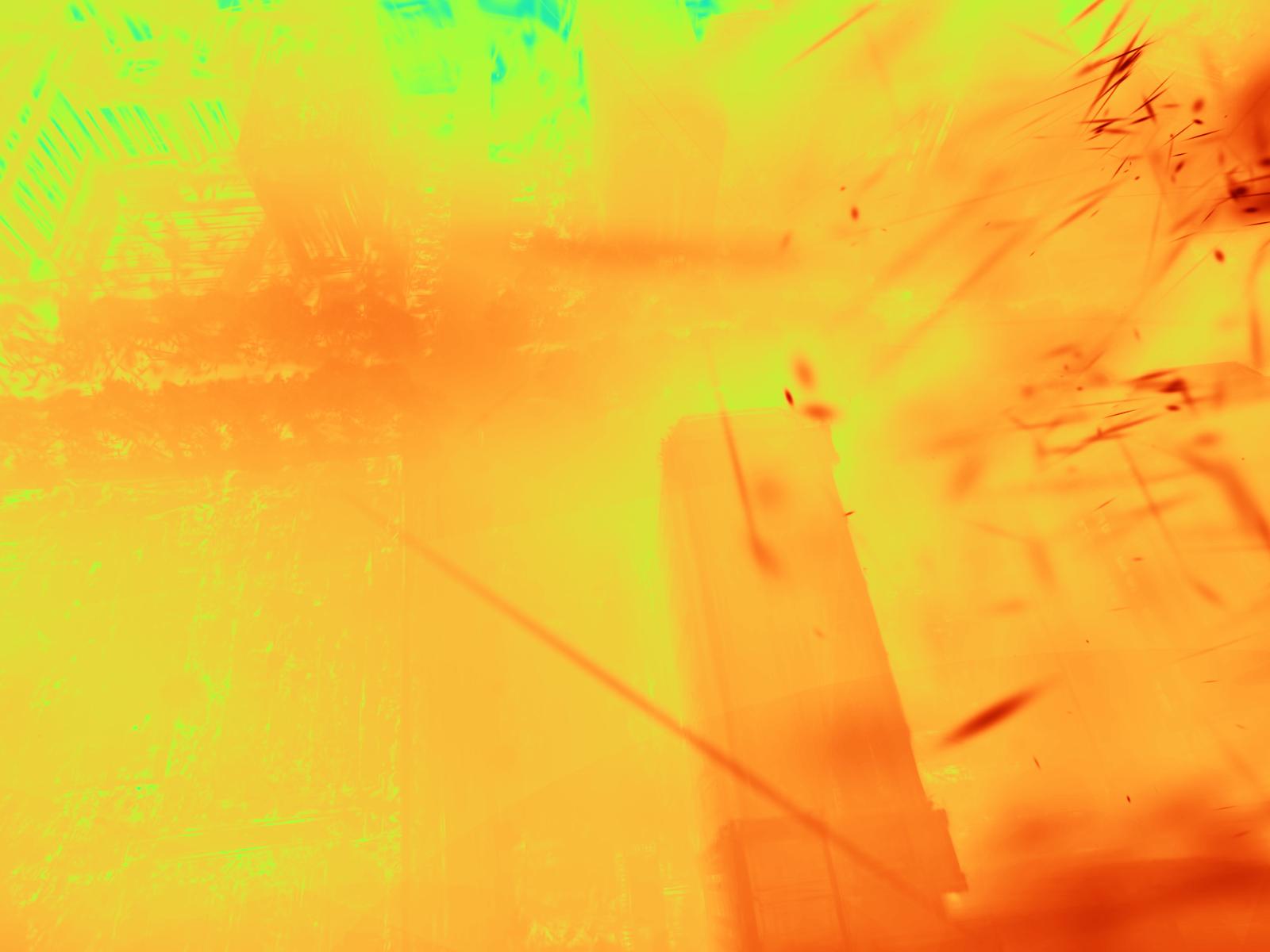} &
        \adjincludegraphics[width=\linewidth, keepaspectratio, trim={600pt 0pt 0pt 0pt}, clip]{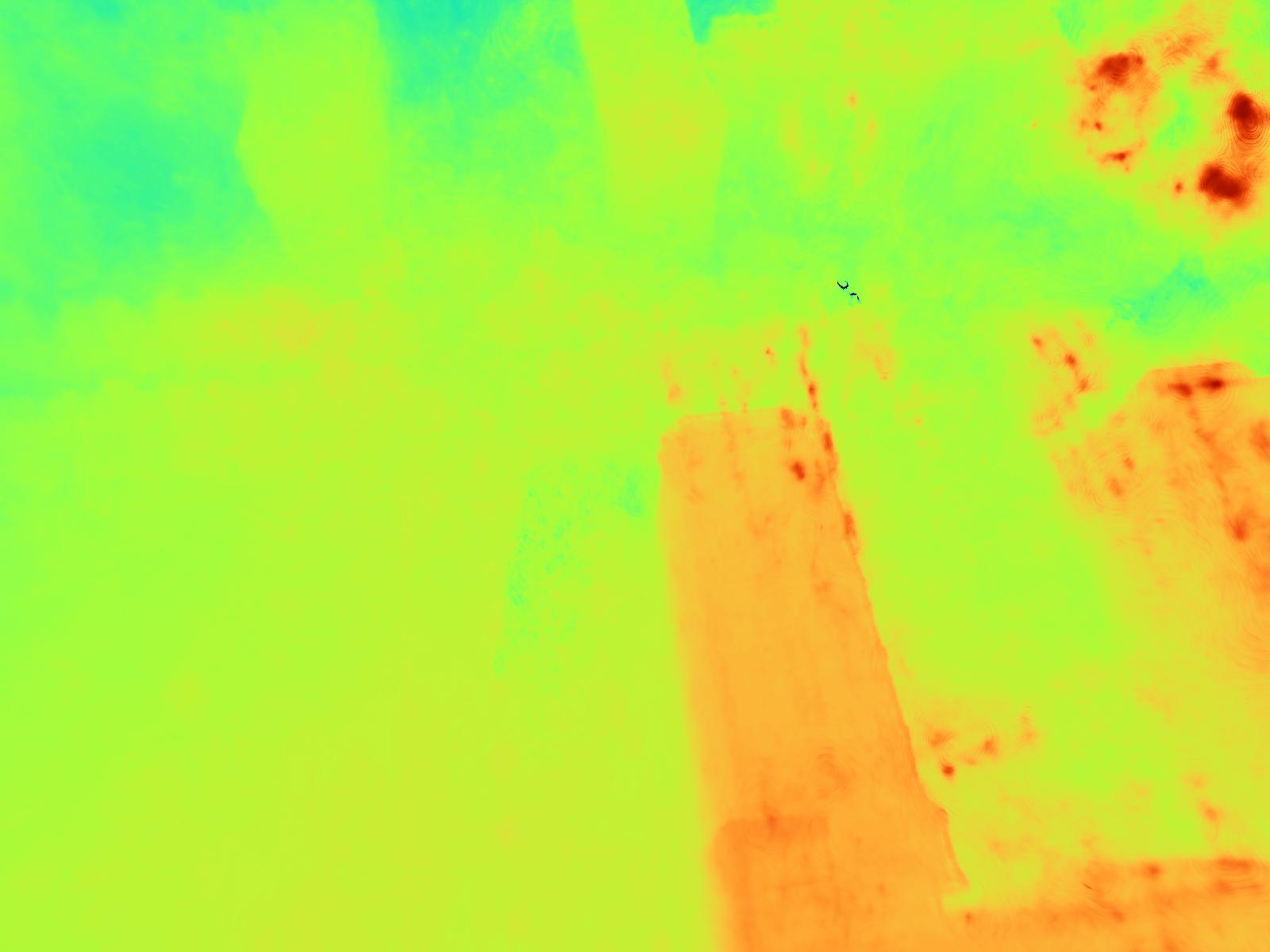}\\
    \end{tabular}
    }
    \caption{\textbf{Depth maps.} Depth maps for ConeGS and the iNGP reconstruction model on the \texttt{01} scene from the OMMO~\cite{lu2023largescaleoutdoormultimodaldataset} dataset. Since the Gaussians are generated from an iNGP reconstruction, the presence of floaters in it leads to floaters also appearing in the Gaussian Splatting results.
}
    \label{fig:depth_issues}
\end{figure}

While our method generally produces strong reconstructions on almost all tested scenes, its reliance on iNGP can make it susceptible to floaters in challenging scenarios such as noisy poses, very large or sparse-view scenes, or other cases where reliable iNGP reconstruction is difficult. One such example is shown in \figref{fig:depth_issues}, where the scene contains distortions and lacks sufficient viewpoint coverage near the cameras. This leads to spurious high-density regions in iNGP close to the cameras, which in turn degrades densification quality by placing Gaussians in incorrect regions. Although this can reduce performance, the method still produces high-quality reconstructions overall (see per-scene results).

\section{Initialization visualizations}
\label{sec:initializations}

\begin{figure}
    \centering

    \begin{minipage}{0.49\columnwidth}
        \centering
        SfM point cloud init. \\[0.5em]
        \includegraphics[width=\columnwidth]{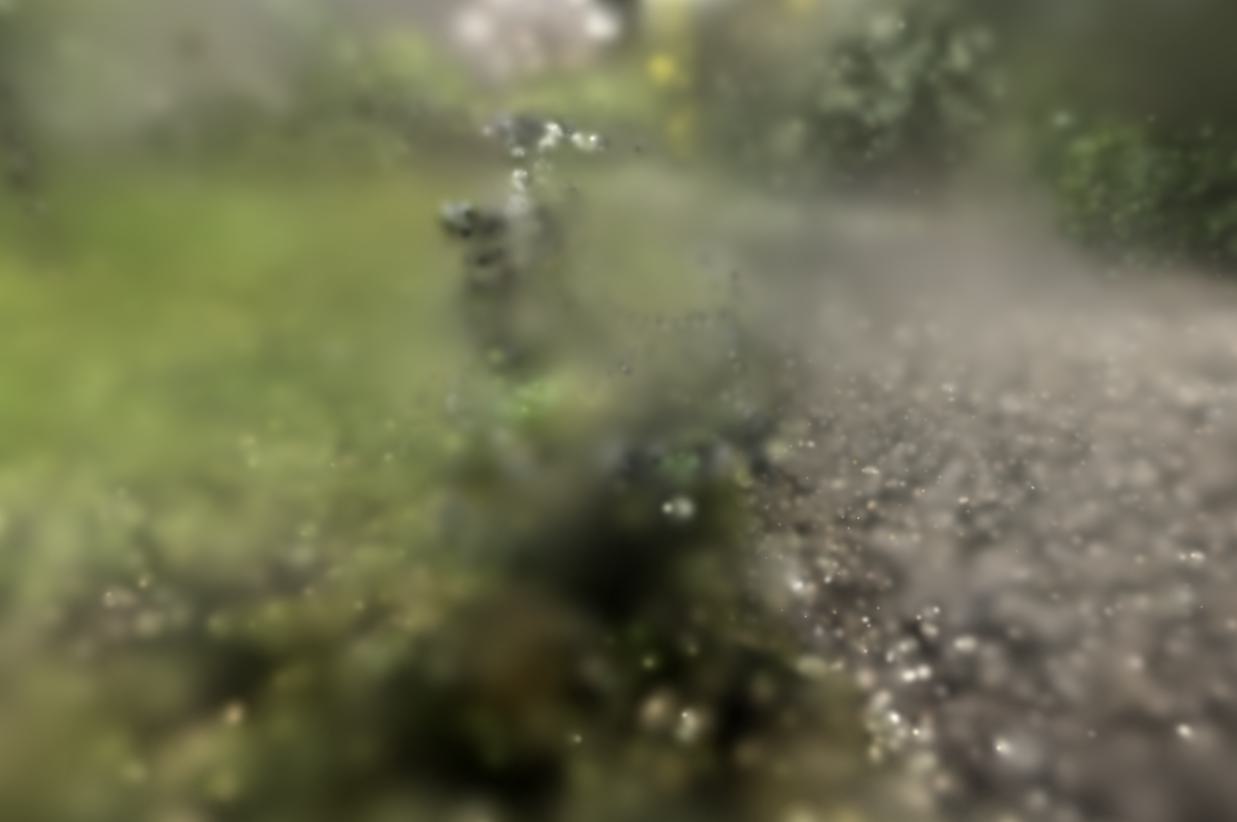}
    \end{minipage}
    \hfill
    \begin{minipage}{0.49\columnwidth}
        \centering
        Our init. \\[0.5em]
        \includegraphics[width=\columnwidth]{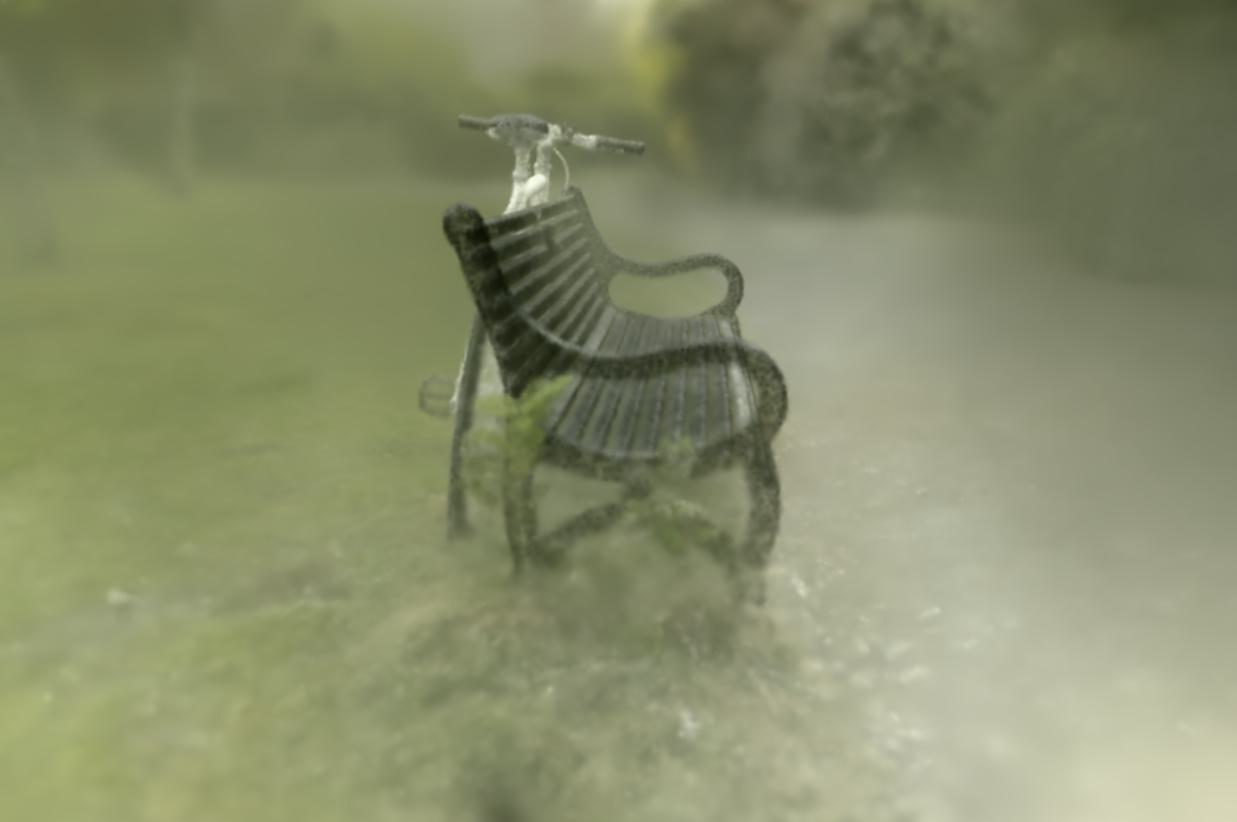}
    \end{minipage}

    \vspace{0.5em}

    \begin{minipage}{0.49\columnwidth}
        \centering
        1× pixel size init. \\[0.5em]
        \includegraphics[width=\columnwidth]{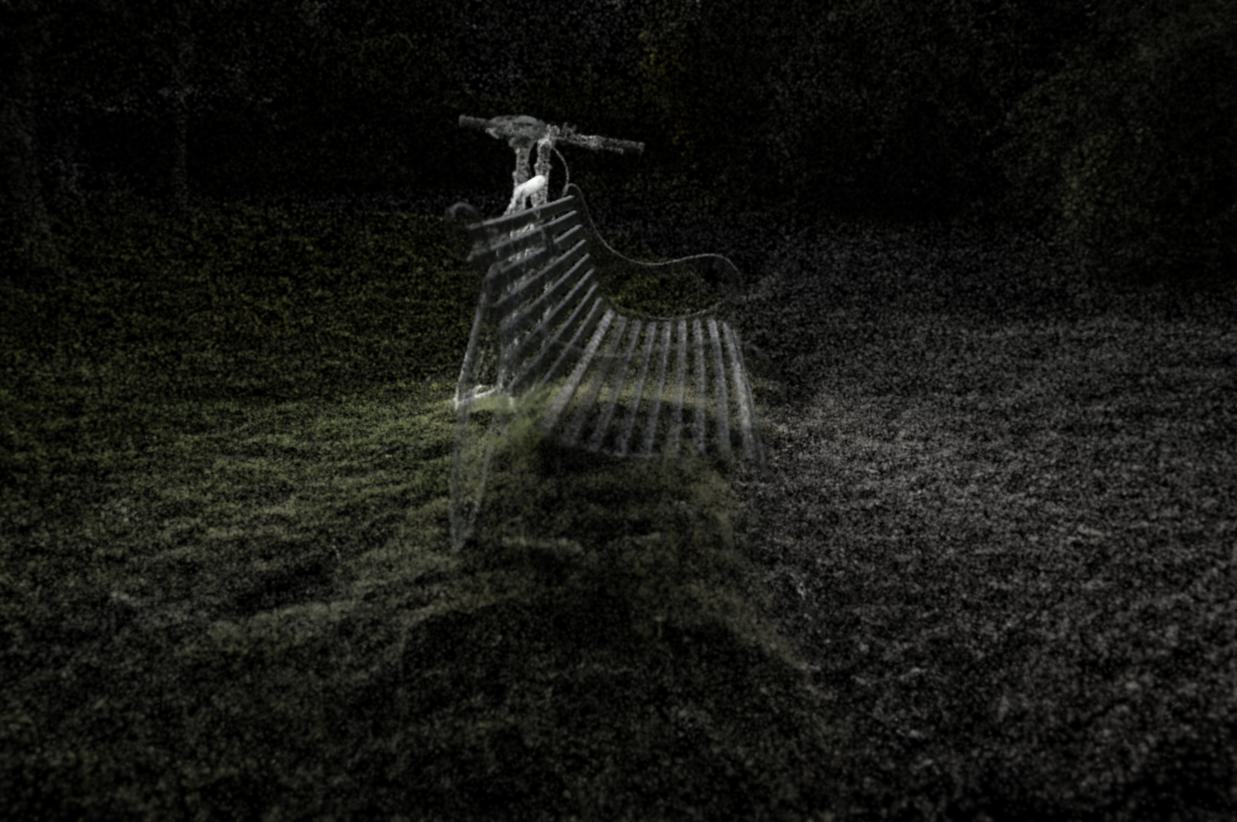}
    \end{minipage}
    \hfill
    \begin{minipage}{0.49\columnwidth}
        \centering
        10× pixel size init. \\[0.5em]
        \includegraphics[width=\columnwidth]{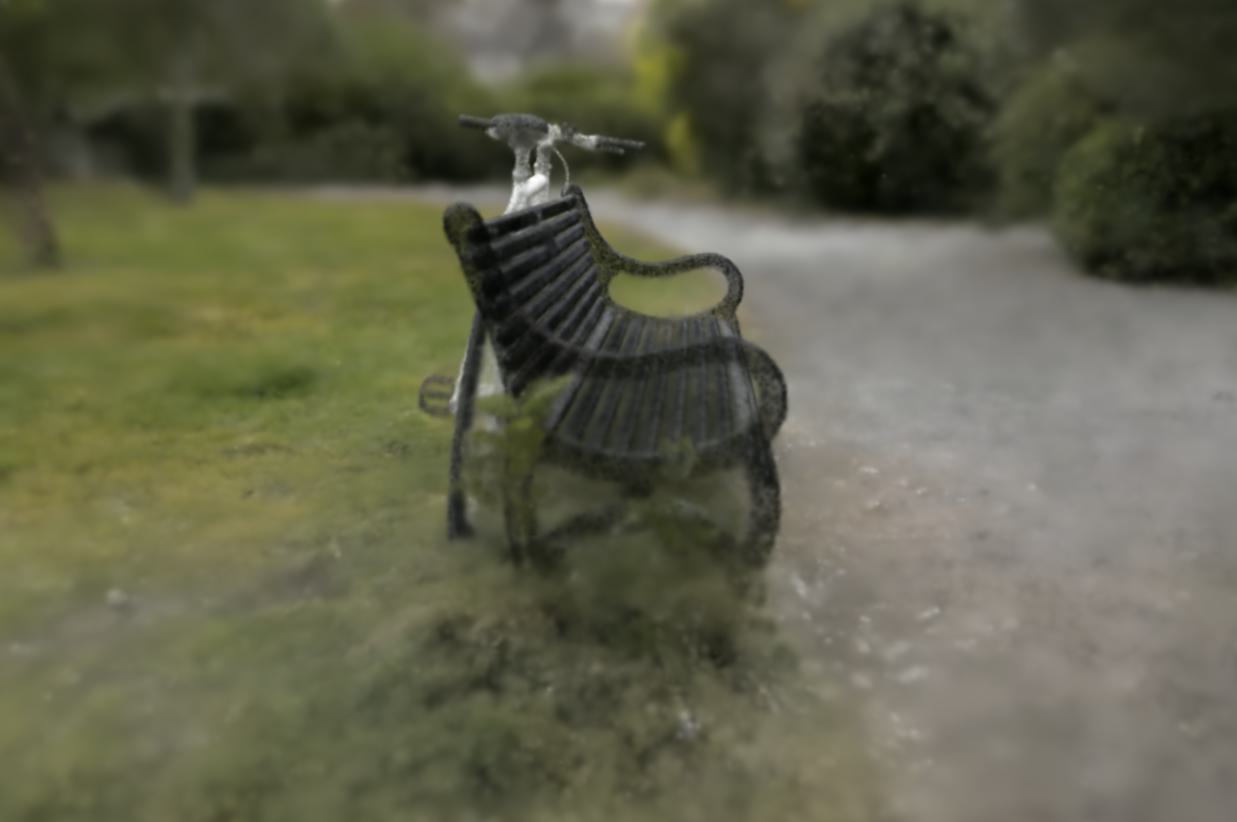}
    \end{minipage}

    \caption{\textbf{Initialization Methods Comparison.} Visual comparison of four initialization methods for 3DGS reconstruction on the \texttt{bicycle} scene from Mip-NeRF 360 dataset~\cite{barron2022mipnerf360}: (1) SfM point cloud initialization, (2) iNGP initialization with kNN-based sizing, (3) iNGP initialization with 1× sizing, and (4) iNGP initialization using the smaller of 10× pixel size or kNN-based sizing. 
}
    \label{fig:init_comparison}
\end{figure}

The choice of initialization has a strong impact on the performance of 3DGS reconstruction. A sufficiently good initialization can even remove the need for additional densification (\cite{Kotovenko2025ARXIV_EDGS_Eliminating_Densification}, \tabref{table:no_limit_benchmark_appendix_mip}, \tabref{table:no_limit_benchmark_appendix_ommo}). We visualize different initialization types in \figref{fig:init_comparison}. Initialization from a sparse SfM point cloud often leaves large gaps that are difficult to fill correctly, while our initialization produces a more uniform coverage. Another important strategy is initialization based on pixel width. This approach does not provide a useful inductive bias of larger primitives and can lead to more gaps in unobserved regions, although it may offer faster rendering due to reduced blending (\secref{subsec:ablations}). Finally, our formulation allows scaling the initial primitives in image space, producing a balance between these two types of initialization.

\section{Gaussian-Based Radiance Field training}
\label{subsec:mapping}

To establish a more direct connection between the volumetric ray marching procedure in Neural Radiance Fields (NeRF) and the rendering in 3D Gaussian Splatting (3DGS), the set of conical frustums sampled along a cone in Mip-NeRF and its derivatives~\cite{barron2021mipnerf, barron2022mipnerf360, barron2023zipnerf} can be reinterpreted as a set of 3D Gaussian primitives, covering approximately the same area. The purpose of this reformulation is to enable training a neural radiance field model using only the 3DGS renderer, without using the traditional volumetric ray integration.

\begin{figure}
    \centering
    \includegraphics[width=1\columnwidth]{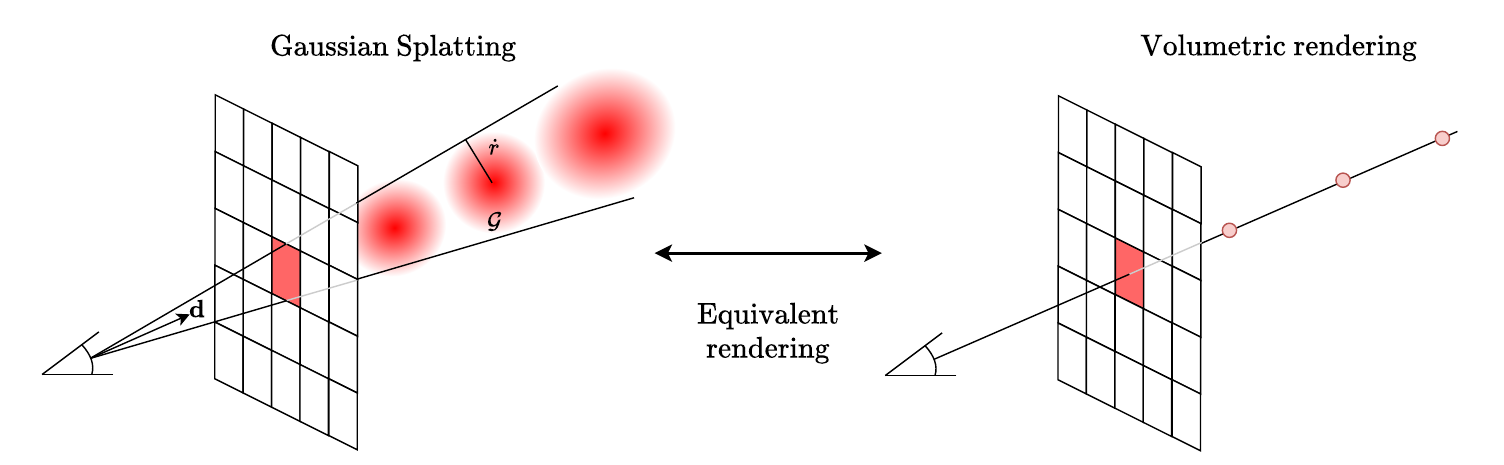}
    \caption{\textbf{Training equivalence.} Illustration of the equivalence between training an implicit radiance field model using NeRF-style ray marching and 3D Gaussian Splatting rasterization.}
    \label{fig:rendering_equivalence}
\end{figure}

To align the 3DGS and NeRF rendering formulations, it is necessary to demonstrate that an entire ray can be equivalently represented as a sequence of Gaussians such that, when rendered, the result is equal to the volumetric integration process. This is achieved by casting cones along pixel directions and subdividing each into a set of conical frustums. Each frustum is then mapped to a Gaussian primitive, where the midpoint
\begin{equation}
t_{\mu,i} = \frac{t_i + t_{i+1}}{2}
\end{equation}
is used to define the Gaussian center \( \mathbf{p}_i \). The view-dependent RGB color \( \mathbf{c}_i \) and scalar density \( \sigma_i \), which is converted to opacity \( o_i \) using~\eqnref{eq:opacity_formula}, are predicted by a neural network. As detailed in \secref{subsec:initialization}, the predicted RGB color \( \mathbf{c}_i \) is encoded as spherical harmonics coefficients, which are compatible with the 3DGS rasterizer.  

Since the Gaussians lie along the cone and their centers are co-linear with the pixel center and the camera origin, their projected 2D means fall directly on the target pixel. Consequently, the maximum opacity contribution aligns with the pixel center, and the kernel evaluation satisfies
\begin{equation}
K(\mathbf{p}_c, \boldsymbol{\mu}^{\text{2D}}_i, \boldsymbol{\Sigma}^{\text{2D}}_i) = 1.
\end{equation}
This condition ensures that the Gaussian opacity \( o_i \) corresponds directly to the opacity \( \alpha_i \) used in blending operations, as defined in~\eqnref{eq:opacity_formula} and~\eqnref{eq:3dgs_alpha}. The opacity is therefore determined by the predicted density and the length of the corresponding frustum.  

To maintain consistent 2D projection footprints across different depths, each Gaussian’s 3D covariance must be adjusted based on its location along the cone. Because elongating Gaussians along the ray direction does not affect the resulting 2D projection, the scaling can be derived using the pixel footprint size and set isotropically using the formulation in~\eqnref{eq:gaussian_scale_isotropic}, while the quaternion can be set to the identity quaternion.

Given these assignments for position, color, opacity, scale, and orientation, each Gaussian can be rendered using the 3DGS renderer to produce the same pixel color as that produced by volumetric rendering. Assuming no low-pass filtering is applied and that only a single image is rendered at a time, it becomes possible to train an Instant-NGP-style model using only the 3DGS rendering pipeline. Although this approach is marginally slower than traditional ray marching, it yields equivalent radiance field representations. A visual comparison of both approaches is shown in \figref{fig:rendering_equivalence}.  

This reinterpretation also provides a more principled foundation for the proposed densification strategy. Instead of using depth, the strategy can be understood as selecting a Gaussian that corresponds to a surface-level conical frustum of pixels with high photometric error. This establishes a clearer theoretical link between cone-based densification and neural implicit models such as NeRF.

\begin{table*}
\centering
\resizebox{0.95\linewidth}{!}{%
\begin{tabular}{cp{1.5cm}|ccccc}
\toprule
& \textit{Scene} & Ours & 3DGS (SfM) & MCMC (SfM) & EDGS \\ 
\midrule
\multirow{8}{*}{\rotatebox{90}{Mip-NeRF360 \cite{barron2022mipnerf360}}} 
 & Bicycle & \colorbox{tabfirst}{24.05} / \colorbox{tabfirst}{0.652} / \colorbox{tabfirst}{0.379} / \colorbox{tabthird}{368} & 20.99 / 0.470 / 0.543 / \colorbox{tabfirst}{466} & \colorbox{tabthird}{23.34} / \colorbox{tabthird}{0.630} / \colorbox{tabthird}{0.400} / 339 & \colorbox{tabsecond}{23.68} / \colorbox{tabsecond}{0.635} / \colorbox{tabsecond}{0.393} / \colorbox{tabsecond}{444} \\
 & Garden & \colorbox{tabfirst}{24.69} / \colorbox{tabsecond}{0.709} / \colorbox{tabfirst}{0.336} / \colorbox{tabthird}{431} & 22.87 / 0.589 / 0.445 / 430 & \colorbox{tabsecond}{24.51} / \colorbox{tabfirst}{0.710} / \colorbox{tabthird}{0.350} / \colorbox{tabfirst}{569} & \colorbox{tabthird}{24.47} / \colorbox{tabthird}{0.706} / \colorbox{tabsecond}{0.341} / \colorbox{tabsecond}{493} \\
 & Stump & \colorbox{tabfirst}{25.71} / \colorbox{tabfirst}{0.707} / \colorbox{tabfirst}{0.351} / \colorbox{tabthird}{456} & 22.79 / 0.547 / 0.498 / \colorbox{tabsecond}{506} & \colorbox{tabsecond}{25.16} / \colorbox{tabsecond}{0.680} / \colorbox{tabsecond}{0.370} / 402 & \colorbox{tabthird}{25.08} / \colorbox{tabthird}{0.673} / \colorbox{tabthird}{0.377} / \colorbox{tabfirst}{588} \\
 & Room & \colorbox{tabfirst}{31.09} / \colorbox{tabfirst}{0.903} / \colorbox{tabfirst}{0.254} / \colorbox{tabsecond}{275} & 27.02 / 0.859 / 0.328 / \colorbox{tabthird}{267} & \colorbox{tabsecond}{30.37} / \colorbox{tabsecond}{0.900} / \colorbox{tabthird}{0.270} / 224 & \colorbox{tabthird}{30.08} / \colorbox{tabthird}{0.896} / \colorbox{tabsecond}{0.265} / \colorbox{tabfirst}{317} \\
 & Counter & \colorbox{tabfirst}{28.22} / \colorbox{tabsecond}{0.879} / \colorbox{tabfirst}{0.251} / \colorbox{tabthird}{229} & 25.64 / 0.845 / 0.307 / \colorbox{tabfirst}{252} & \colorbox{tabsecond}{27.83} / \colorbox{tabfirst}{0.880} / \colorbox{tabthird}{0.260} / 198 & \colorbox{tabthird}{27.77} / \colorbox{tabthird}{0.877} / \colorbox{tabfirst}{0.251} / \colorbox{tabsecond}{245} \\
 & Kitchen & \colorbox{tabfirst}{29.73} / \colorbox{tabfirst}{0.896} / \colorbox{tabfirst}{0.183} / 239 & 20.53 / 0.671 / 0.440 / \colorbox{tabfirst}{297} & \colorbox{tabthird}{28.37} / \colorbox{tabthird}{0.890} / \colorbox{tabthird}{0.210} / \colorbox{tabthird}{246} & \colorbox{tabsecond}{28.84} / \colorbox{tabsecond}{0.891} / \colorbox{tabsecond}{0.186} / \colorbox{tabsecond}{269} \\
 & Bonsai & \colorbox{tabfirst}{30.67} / \colorbox{tabfirst}{0.920} / \colorbox{tabfirst}{0.241} / 276 & 25.41 / 0.866 / 0.331 / \colorbox{tabfirst}{326} & \colorbox{tabsecond}{29.84} / \colorbox{tabsecond}{0.910} / \colorbox{tabthird}{0.260} / \colorbox{tabthird}{281} & \colorbox{tabthird}{29.73} / \colorbox{tabthird}{0.906} / \colorbox{tabsecond}{0.256} / \colorbox{tabsecond}{298} \\
 & \textbf{Average} & \colorbox{tabfirst}{27.74} / \colorbox{tabfirst}{0.809} / \colorbox{tabfirst}{0.285} / \colorbox{tabthird}{325} & 23.61 / 0.692 / 0.413 / \colorbox{tabsecond}{363} & \colorbox{tabthird}{27.06} / \colorbox{tabsecond}{0.800} / \colorbox{tabthird}{0.303} / 323 & \colorbox{tabsecond}{27.09} / \colorbox{tabthird}{0.798} / \colorbox{tabsecond}{0.296} / \colorbox{tabfirst}{379} \\

\midrule
\multirow{9}{*}{\rotatebox{90}{OMMO \cite{lu2023largescaleoutdoormultimodaldataset}}}
 & 01 & \colorbox{tabsecond}{22.58} / \colorbox{tabfirst}{0.625} / \colorbox{tabfirst}{0.458} / \colorbox{tabsecond}{212} & 20.53 / 0.543 / 0.575 / 159 & \colorbox{tabfirst}{22.66} / \colorbox{tabsecond}{0.620} / \colorbox{tabsecond}{0.470} / \colorbox{tabthird}{180} & \colorbox{tabthird}{22.50} / \colorbox{tabthird}{0.608} / \colorbox{tabthird}{0.475} / \colorbox{tabfirst}{222} \\
 & 03 & \colorbox{tabfirst}{25.39} / \colorbox{tabfirst}{0.842} / \colorbox{tabfirst}{0.244} / \colorbox{tabfirst}{356} & 24.09 / 0.803 / 0.295 / 112 & \colorbox{tabthird}{24.30} / \colorbox{tabthird}{0.810} / \colorbox{tabthird}{0.290} / \colorbox{tabthird}{217} & \colorbox{tabsecond}{24.85} / \colorbox{tabsecond}{0.827} / \colorbox{tabsecond}{0.259} / \colorbox{tabsecond}{313} \\
 & 05 & \colorbox{tabfirst}{27.95} / \colorbox{tabfirst}{0.863} / \colorbox{tabfirst}{0.246} / \colorbox{tabfirst}{415} & 26.87 / 0.841 / 0.296 / \colorbox{tabthird}{354} & \colorbox{tabsecond}{27.70} / \colorbox{tabsecond}{0.860} / \colorbox{tabthird}{0.270} / 318 & \colorbox{tabthird}{27.69} / \colorbox{tabthird}{0.856} / \colorbox{tabsecond}{0.259} / \colorbox{tabsecond}{398} \\
 & 06 & \colorbox{tabfirst}{27.30} / \colorbox{tabfirst}{0.913} / \colorbox{tabsecond}{0.202} / \colorbox{tabthird}{321} & \colorbox{tabthird}{26.53} / 0.899 / 0.229 / 107 & 26.25 / \colorbox{tabsecond}{0.910} / \colorbox{tabthird}{0.210} / \colorbox{tabfirst}{339} & \colorbox{tabsecond}{26.60} / \colorbox{tabsecond}{0.910} / \colorbox{tabfirst}{0.198} / \colorbox{tabfirst}{339} \\
 & 10 & \colorbox{tabfirst}{29.31} / \colorbox{tabfirst}{0.853} / \colorbox{tabfirst}{0.252} / \colorbox{tabsecond}{338} & \colorbox{tabsecond}{28.86} / 0.833 / 0.291 / 125 & 28.70 / \colorbox{tabsecond}{0.840} / \colorbox{tabthird}{0.280} / \colorbox{tabthird}{337} & \colorbox{tabthird}{28.78} / \colorbox{tabthird}{0.839} / \colorbox{tabsecond}{0.275} / \colorbox{tabfirst}{341} \\
 & 13 & \colorbox{tabfirst}{30.36} / \colorbox{tabfirst}{0.899} / \colorbox{tabfirst}{0.211} / \colorbox{tabsecond}{449} & \colorbox{tabsecond}{29.79} / \colorbox{tabsecond}{0.889} / \colorbox{tabthird}{0.247} / 171 & \colorbox{tabthird}{29.45} / \colorbox{tabthird}{0.880} / 0.250 / \colorbox{tabthird}{378} & 28.98 / 0.878 / \colorbox{tabsecond}{0.234} / \colorbox{tabfirst}{468} \\
 & 14 & \colorbox{tabfirst}{29.73} / \colorbox{tabfirst}{0.924} / \colorbox{tabfirst}{0.145} / \colorbox{tabfirst}{394} & 27.09 / 0.874 / 0.220 / 237 & \colorbox{tabsecond}{29.26} / \colorbox{tabsecond}{0.920} / \colorbox{tabsecond}{0.160} / \colorbox{tabthird}{360} & \colorbox{tabthird}{28.92} / \colorbox{tabthird}{0.909} / \colorbox{tabthird}{0.169} / \colorbox{tabsecond}{374} \\
 & 15 & \colorbox{tabfirst}{28.10} / \colorbox{tabfirst}{0.893} / \colorbox{tabfirst}{0.182} / \colorbox{tabthird}{405} & \colorbox{tabsecond}{27.80} / \colorbox{tabthird}{0.877} / \colorbox{tabthird}{0.213} / 153 & \colorbox{tabthird}{27.72} / \colorbox{tabsecond}{0.890} / \colorbox{tabsecond}{0.200} / \colorbox{tabfirst}{427} & 27.58 / \colorbox{tabthird}{0.877} / 0.215 / \colorbox{tabsecond}{420} \\
 & \textbf{Average} & \colorbox{tabfirst}{27.59} / \colorbox{tabfirst}{0.852} / \colorbox{tabfirst}{0.242} / \colorbox{tabfirst}{361} & 26.45 / 0.820 / 0.296 / 177 & \colorbox{tabsecond}{27.00} / \colorbox{tabsecond}{0.841} / \colorbox{tabthird}{0.266} / \colorbox{tabthird}{320} & \colorbox{tabthird}{26.99} / \colorbox{tabthird}{0.838} / \colorbox{tabsecond}{0.261} / \colorbox{tabsecond}{359} \\

\midrule
\multirow{3}{*}{\fontsize{6}{7}\selectfont  \rotatebox{90}{\makecell{Tanks \& \\ Temples \cite{Knapitsch2017}}}}
 & Truck & \colorbox{tabfirst}{24.54} / \colorbox{tabfirst}{0.822} / \colorbox{tabfirst}{0.291} / \colorbox{tabfirst}{184} & \colorbox{tabthird}{23.59} / \colorbox{tabthird}{0.803} / 0.318 / 128 & \colorbox{tabsecond}{23.86} / \colorbox{tabsecond}{0.810} / \colorbox{tabthird}{0.316} / \colorbox{tabthird}{155} & 23.33 / 0.799 / \colorbox{tabsecond}{0.314} / \colorbox{tabsecond}{183} \\
 & Train & \colorbox{tabfirst}{21.70} / \colorbox{tabfirst}{0.759} / \colorbox{tabfirst}{0.328} / \colorbox{tabfirst}{204} & \colorbox{tabthird}{21.17} / 0.744 / 0.349 / 118 & 21.13 / \colorbox{tabthird}{0.749} / \colorbox{tabthird}{0.347} / \colorbox{tabthird}{160} & \colorbox{tabsecond}{21.31} / \colorbox{tabsecond}{0.756} / \colorbox{tabsecond}{0.334} / \colorbox{tabsecond}{180} \\
 & \textbf{Average} & \colorbox{tabfirst}{23.12} / \colorbox{tabfirst}{0.790} / \colorbox{tabfirst}{0.309} / \colorbox{tabfirst}{194} & \colorbox{tabthird}{22.38} / 0.774 / 0.334 / 123 & \colorbox{tabsecond}{22.49} / \colorbox{tabsecond}{0.780} / \colorbox{tabthird}{0.332} / \colorbox{tabthird}{158} & 22.32 / \colorbox{tabthird}{0.778} / \colorbox{tabsecond}{0.324} / \colorbox{tabsecond}{182} \\

\midrule
\multirow{3}{*}{\fontsize{8}{8}\selectfont \rotatebox{90}{\makecell{Deep \\ Blending \cite{DeepBlending2018}}}} 
 & Dr Johnson & \colorbox{tabfirst}{28.78} / \colorbox{tabfirst}{0.875} / \colorbox{tabfirst}{0.338} / \colorbox{tabthird}{447} & 23.58 / 0.808 / 0.440 / \colorbox{tabsecond}{521} & \colorbox{tabsecond}{28.16} / \colorbox{tabsecond}{0.869} / \colorbox{tabsecond}{0.343} / \colorbox{tabfirst}{550} & \colorbox{tabthird}{27.66} / \colorbox{tabthird}{0.865} / \colorbox{tabthird}{0.347} / 444 \\
 & Playroom & \colorbox{tabfirst}{30.09} / \colorbox{tabfirst}{0.885} / \colorbox{tabfirst}{0.318} / \colorbox{tabsecond}{489} & 25.72 / 0.845 / 0.384 / \colorbox{tabsecond}{489} & \colorbox{tabsecond}{29.72} / \colorbox{tabsecond}{0.883} / \colorbox{tabsecond}{0.323} / \colorbox{tabfirst}{552} & \colorbox{tabthird}{29.19} / \colorbox{tabthird}{0.879} / \colorbox{tabthird}{0.327} / 486 \\
 & \textbf{Average} & \colorbox{tabfirst}{29.44} / \colorbox{tabfirst}{0.880} / \colorbox{tabfirst}{0.328} / \colorbox{tabthird}{468} & 24.65 / 0.827 / 0.412 / \colorbox{tabsecond}{505} & \colorbox{tabsecond}{28.94} / \colorbox{tabsecond}{0.876} / \colorbox{tabsecond}{0.333} / \colorbox{tabfirst}{551} & \colorbox{tabthird}{28.43} / \colorbox{tabthird}{0.872} / \colorbox{tabthird}{0.337} / 465 \\
\bottomrule
\end{tabular}}
\caption{\textbf{Detailed results} on a selection of datasets and methods with the number of Gaussians limited to 100k. Each field contains PSNR, SSIM, LPIPS and FPS respectively. We highlight the \colorbox{tabfirst}{best}, \colorbox{tabsecond}{second best} and \colorbox{tabthird}{third best} results among all. Slight discrepancies from the main table are due to rounding.}
\label{table:expanded_all_results_100k}
\end{table*}

\begin{table*}
\centering
\resizebox{0.95\linewidth}{!}{%
\begin{tabular}{cp{1.5cm}|ccccc}
\toprule
& \textit{Scene} & Ours & 3DGS (SfM) & MCMC (SfM) & EDGS \\ 
\midrule
\multirow{8}{*}{\rotatebox{90}{Mip-NeRF360 \cite{barron2022mipnerf360}}} 
& Bicycle & \colorbox{tabfirst}{25.25} / \colorbox{tabfirst}{0.758} / \colorbox{tabfirst}{0.242} / \colorbox{tabsecond}{234} & 24.06 / 0.651 / 0.370 / \colorbox{tabthird}{174} & \colorbox{tabthird}{24.90} / \colorbox{tabthird}{0.740} / \colorbox{tabthird}{0.282} / 162 & \colorbox{tabsecond}{24.95} / \colorbox{tabsecond}{0.751} / \colorbox{tabsecond}{0.251} / \colorbox{tabfirst}{248} \\
 & Garden & \colorbox{tabfirst}{26.87} / \colorbox{tabfirst}{0.837} / \colorbox{tabfirst}{0.157} / \colorbox{tabthird}{246} & 25.16 / 0.743 / 0.296 / \colorbox{tabfirst}{297} & \colorbox{tabthird}{26.36} / \colorbox{tabthird}{0.823} / \colorbox{tabthird}{0.189} / 221 & \colorbox{tabsecond}{26.49} / \colorbox{tabsecond}{0.833} / \colorbox{tabsecond}{0.161} / \colorbox{tabsecond}{252} \\
 & Stump & \colorbox{tabfirst}{27.03} / \colorbox{tabfirst}{0.793} / \colorbox{tabfirst}{0.215} / \colorbox{tabsecond}{255} & 25.27 / 0.688 / 0.348 / \colorbox{tabthird}{226} & \colorbox{tabsecond}{26.81} / \colorbox{tabsecond}{0.779} / \colorbox{tabthird}{0.250} / 175 & \colorbox{tabthird}{26.50} / \colorbox{tabthird}{0.768} / \colorbox{tabsecond}{0.237} / \colorbox{tabfirst}{296} \\
 & Room & \colorbox{tabfirst}{32.02} / \colorbox{tabfirst}{0.924} / \colorbox{tabsecond}{0.205} / \colorbox{tabfirst}{192} & \colorbox{tabthird}{31.47} / 0.913 / 0.234 / 112 & \colorbox{tabsecond}{31.72} / \colorbox{tabthird}{0.921} / \colorbox{tabthird}{0.221} / \colorbox{tabthird}{114} & 31.36 / \colorbox{tabsecond}{0.923} / \colorbox{tabfirst}{0.203} / \colorbox{tabsecond}{168} \\
 & Counter & \colorbox{tabthird}{28.87} / \colorbox{tabthird}{0.906} / \colorbox{tabsecond}{0.194} / \colorbox{tabfirst}{148} & 28.86 / 0.901 / 0.213 / \colorbox{tabthird}{113} & \colorbox{tabsecond}{28.95} / \colorbox{tabsecond}{0.907} / \colorbox{tabthird}{0.205} / 93 & \colorbox{tabfirst}{29.05} / \colorbox{tabfirst}{0.912} / \colorbox{tabfirst}{0.184} / \colorbox{tabsecond}{137} \\
 & Kitchen & \colorbox{tabfirst}{31.38} / \colorbox{tabfirst}{0.929} / \colorbox{tabsecond}{0.124} / \colorbox{tabfirst}{172} & 30.76 / 0.918 / 0.143 / \colorbox{tabthird}{111} & \colorbox{tabthird}{31.04} / \colorbox{tabthird}{0.921} / \colorbox{tabthird}{0.141} / \colorbox{tabthird}{111} & \colorbox{tabsecond}{31.35} / \colorbox{tabsecond}{0.928} / \colorbox{tabfirst}{0.122} / \colorbox{tabsecond}{154} \\
 & Bonsai & \colorbox{tabfirst}{32.16} / \colorbox{tabfirst}{0.944} / \colorbox{tabfirst}{0.191} / \colorbox{tabfirst}{190} & 31.95 / 0.936 / 0.218 / \colorbox{tabthird}{135} & \colorbox{tabthird}{31.97} / \colorbox{tabthird}{0.939} / \colorbox{tabthird}{0.210} / 128 & \colorbox{tabsecond}{32.02} / \colorbox{tabsecond}{0.942} / \colorbox{tabfirst}{0.191} / \colorbox{tabsecond}{170} \\
 & \textbf{Average} & \colorbox{tabfirst}{29.08} / \colorbox{tabfirst}{0.870} / \colorbox{tabfirst}{0.190} / \colorbox{tabfirst}{205} & 28.22 / 0.821 / 0.260 / \colorbox{tabthird}{167} & \colorbox{tabsecond}{28.82} / \colorbox{tabthird}{0.861} / \colorbox{tabthird}{0.214} / 143 & \colorbox{tabsecond}{28.82} / \colorbox{tabsecond}{0.865} / \colorbox{tabsecond}{0.193} / \colorbox{tabsecond}{204} \\

\midrule
\multirow{9}{*}{\rotatebox{90}{OMMO \cite{lu2023largescaleoutdoormultimodaldataset}}}
 & 01 & 23.46 / \colorbox{tabfirst}{0.690} / \colorbox{tabfirst}{0.356} / \colorbox{tabfirst}{140} & \colorbox{tabfirst}{23.81} / 0.682 / 0.385 / \colorbox{tabsecond}{102} & \colorbox{tabsecond}{23.73} / \colorbox{tabthird}{0.685} / \colorbox{tabthird}{0.378} / \colorbox{tabthird}{90} & \colorbox{tabthird}{23.63} / \colorbox{tabsecond}{0.686} / \colorbox{tabsecond}{0.363} / 84 \\
 & 03 & \colorbox{tabsecond}{27.01} / \colorbox{tabsecond}{0.887} / \colorbox{tabsecond}{0.181} / \colorbox{tabfirst}{197} & 26.18 / 0.868 / 0.220 / \colorbox{tabsecond}{111} & \colorbox{tabthird}{26.50} / \colorbox{tabthird}{0.875} / \colorbox{tabthird}{0.212} / \colorbox{tabthird}{105} & \colorbox{tabfirst}{27.16} / \colorbox{tabfirst}{0.892} / \colorbox{tabfirst}{0.176} / 93 \\
 & 05 & \colorbox{tabthird}{28.61} / \colorbox{tabsecond}{0.877} / \colorbox{tabsecond}{0.199} / \colorbox{tabfirst}{232} & 28.53 / 0.874 / 0.235 / \colorbox{tabsecond}{162} & \colorbox{tabsecond}{28.71} / \colorbox{tabsecond}{0.877} / \colorbox{tabthird}{0.233} / \colorbox{tabthird}{154} & \colorbox{tabfirst}{28.79} / \colorbox{tabfirst}{0.881} / \colorbox{tabfirst}{0.196} / 122 \\
 & 06 & \colorbox{tabfirst}{27.79} / \colorbox{tabthird}{0.932} / \colorbox{tabsecond}{0.159} / \colorbox{tabfirst}{195} & \colorbox{tabsecond}{27.59} / 0.931 / 0.161 / \colorbox{tabthird}{120} & 26.99 / \colorbox{tabsecond}{0.934} / \colorbox{tabsecond}{0.159} / \colorbox{tabsecond}{152} & \colorbox{tabthird}{27.38} / \colorbox{tabfirst}{0.941} / \colorbox{tabfirst}{0.135} / 105 \\
 & 10 & \colorbox{tabfirst}{31.16} / \colorbox{tabsecond}{0.905} / \colorbox{tabfirst}{0.165} / \colorbox{tabfirst}{235} & \colorbox{tabthird}{31.04} / \colorbox{tabthird}{0.894} / 0.194 / \colorbox{tabthird}{143} & 30.56 / 0.892 / \colorbox{tabthird}{0.192} / \colorbox{tabsecond}{168} & \colorbox{tabsecond}{31.11} / \colorbox{tabfirst}{0.906} / \colorbox{tabfirst}{0.165} / 129 \\
 & 13 & \colorbox{tabfirst}{33.02} / \colorbox{tabfirst}{0.949} / \colorbox{tabfirst}{0.115} / \colorbox{tabfirst}{280} & \colorbox{tabsecond}{32.40} / \colorbox{tabthird}{0.939} / \colorbox{tabthird}{0.149} / \colorbox{tabthird}{165} & \colorbox{tabthird}{32.18} / 0.934 / 0.155 / \colorbox{tabsecond}{175} & 31.95 / \colorbox{tabsecond}{0.944} / \colorbox{tabsecond}{0.123} / 151 \\
 & 14 & \colorbox{tabfirst}{31.57} / \colorbox{tabfirst}{0.950} / \colorbox{tabsecond}{0.096} / \colorbox{tabfirst}{225} & \colorbox{tabfirst}{31.57} / \colorbox{tabthird}{0.946} / \colorbox{tabthird}{0.107} / \colorbox{tabsecond}{135} & 31.14 / \colorbox{tabthird}{0.946} / 0.108 / \colorbox{tabthird}{130} & \colorbox{tabthird}{31.51} / \colorbox{tabfirst}{0.950} / \colorbox{tabfirst}{0.095} / 112 \\
 & 15 & \colorbox{tabthird}{30.50} / \colorbox{tabfirst}{0.944} / \colorbox{tabfirst}{0.090} / \colorbox{tabfirst}{231} & \colorbox{tabfirst}{30.59} / \colorbox{tabthird}{0.934} / 0.114 / \colorbox{tabthird}{149} & 29.92 / 0.933 / \colorbox{tabthird}{0.113} / \colorbox{tabsecond}{162} & \colorbox{tabsecond}{30.52} / \colorbox{tabsecond}{0.942} / \colorbox{tabsecond}{0.095} / 125 \\
 & \textbf{Average} & \colorbox{tabfirst}{29.14} / \colorbox{tabsecond}{0.892} / \colorbox{tabsecond}{0.170} / \colorbox{tabfirst}{217} & \colorbox{tabthird}{28.96} / 0.884 / 0.196 / \colorbox{tabthird}{136} & 28.72 / \colorbox{tabthird}{0.885} / \colorbox{tabthird}{0.194} / \colorbox{tabsecond}{142} & \colorbox{tabsecond}{29.01} / \colorbox{tabfirst}{0.893} / \colorbox{tabfirst}{0.169} / 115 \\
\midrule
\multirow{3}{*}{\fontsize{6}{7}\selectfont  \rotatebox{90}{\makecell{Tanks \& \\ Temples \cite{Knapitsch2017}}}}
 & Truck & \colorbox{tabfirst}{25.41} / \colorbox{tabfirst}{0.858} / \colorbox{tabfirst}{0.204} / \colorbox{tabfirst}{122} & \colorbox{tabthird}{24.80} / 0.842 / 0.253 / \colorbox{tabthird}{93} & \colorbox{tabsecond}{25.24} / \colorbox{tabsecond}{0.851} / \colorbox{tabthird}{0.242} / 68 & 24.66 / \colorbox{tabthird}{0.850} / \colorbox{tabsecond}{0.212} / \colorbox{tabsecond}{114} \\
 & Train & 21.97 / \colorbox{tabsecond}{0.799} / \colorbox{tabsecond}{0.253} / \colorbox{tabfirst}{140} & \colorbox{tabfirst}{22.28} / 0.790 / 0.277 / 72 & \colorbox{tabthird}{22.12} / \colorbox{tabthird}{0.798} / \colorbox{tabthird}{0.275} / \colorbox{tabthird}{86} & \colorbox{tabsecond}{22.27} / \colorbox{tabfirst}{0.812} / \colorbox{tabfirst}{0.246} / \colorbox{tabsecond}{104} \\
 & \textbf{Average} & \colorbox{tabfirst}{23.69} / \colorbox{tabsecond}{0.829} / \colorbox{tabfirst}{0.229} / \colorbox{tabfirst}{131} & \colorbox{tabthird}{23.54} / 0.816 / 0.265 / \colorbox{tabthird}{82} & \colorbox{tabsecond}{23.68} / \colorbox{tabthird}{0.825} / \colorbox{tabthird}{0.259} / 77 & 23.47 / \colorbox{tabfirst}{0.831} / \colorbox{tabfirst}{0.229} / \colorbox{tabsecond}{109} \\
\midrule
\multirow{3}{*}{\fontsize{8}{8}\selectfont \rotatebox{90}{\makecell{Deep \\ Blending \cite{DeepBlending2018}}}} 
 & Dr Johnson & \colorbox{tabfirst}{29.20} / \colorbox{tabfirst}{0.890} / \colorbox{tabfirst}{0.291} / \colorbox{tabfirst}{270} & \colorbox{tabsecond}{28.85} / 0.881 / \colorbox{tabthird}{0.307} / 151 & \colorbox{tabthird}{28.71} / \colorbox{tabthird}{0.884} / 0.313 / \colorbox{tabthird}{180} & 28.60 / \colorbox{tabsecond}{0.888} / \colorbox{tabsecond}{0.292} / \colorbox{tabsecond}{269} \\
 & Playroom & \colorbox{tabfirst}{30.51} / \colorbox{tabfirst}{0.892} / \colorbox{tabfirst}{0.279} / \colorbox{tabfirst}{287} & 29.66 / 0.883 / \colorbox{tabthird}{0.298} / 161 & \colorbox{tabsecond}{30.16} / \colorbox{tabthird}{0.889} / 0.303 / \colorbox{tabthird}{209} & \colorbox{tabthird}{30.06} / \colorbox{tabsecond}{0.890} / \colorbox{tabsecond}{0.281} / \colorbox{tabsecond}{277} \\
 & \textbf{Average} & \colorbox{tabfirst}{29.86} / \colorbox{tabfirst}{0.891} / \colorbox{tabfirst}{0.285} / \colorbox{tabfirst}{278} & 29.26 / 0.882 / \colorbox{tabthird}{0.303} / 156 & \colorbox{tabsecond}{29.44} / \colorbox{tabthird}{0.887} / 0.308 / \colorbox{tabthird}{194} & \colorbox{tabthird}{29.33} / \colorbox{tabsecond}{0.889} / \colorbox{tabsecond}{0.287} / \colorbox{tabsecond}{273} \\
 
\bottomrule
\end{tabular}}
\caption{\textbf{Detailed results} on a selection of datasets and methods with the number of Gaussians limited to 500k. Each field contains PSNR, SSIM, LPIPS and FPS respectively. We highlight the \colorbox{tabfirst}{best}, \colorbox{tabsecond}{second best} and \colorbox{tabthird}{third best} results among all. Slight discrepancies from the main table are due to rounding.}
\label{table:expanded_all_results_500k}
\end{table*}

\begin{table*}
\centering
\resizebox{0.95\linewidth}{!}{%
\begin{tabular}{cp{1.5cm}|ccccc}
\toprule
& \textit{Scene} & Ours & 3DGS (SfM) & MCMC (SfM) & EDGS \\ 
\midrule
\multirow{8}{*}{\rotatebox{90}{Mip-NeRF360 \cite{barron2022mipnerf360}}} 
 & Bicycle & \colorbox{tabfirst}{25.45} / \colorbox{tabfirst}{0.778} / \colorbox{tabfirst}{0.198} / \colorbox{tabsecond}{156} & 24.40 / 0.688 / 0.325 / \colorbox{tabthird}{129} & \colorbox{tabsecond}{25.34} / \colorbox{tabthird}{0.768} / \colorbox{tabthird}{0.238} / 106 & \colorbox{tabthird}{25.26} / \colorbox{tabfirst}{0.778} / \colorbox{tabsecond}{0.199} / \colorbox{tabfirst}{161} \\
 & Garden & \colorbox{tabfirst}{27.42} / \colorbox{tabfirst}{0.861} / \colorbox{tabfirst}{0.115} / \colorbox{tabfirst}{162} & 26.70 / 0.827 / 0.172 / \colorbox{tabthird}{142} & \colorbox{tabthird}{26.83} / \colorbox{tabthird}{0.847} / \colorbox{tabthird}{0.147} / 135 & \colorbox{tabsecond}{27.05} / \colorbox{tabsecond}{0.857} / \colorbox{tabsecond}{0.118} / \colorbox{tabsecond}{158} \\
 & Stump & \colorbox{tabfirst}{27.23} / \colorbox{tabfirst}{0.802} / \colorbox{tabfirst}{0.184} / \colorbox{tabsecond}{164} & 25.64 / 0.719 / 0.301 / \colorbox{tabthird}{162} & \colorbox{tabsecond}{27.21} / \colorbox{tabsecond}{0.798} / \colorbox{tabthird}{0.213} / 110 & \colorbox{tabthird}{26.71} / \colorbox{tabthird}{0.783} / \colorbox{tabsecond}{0.203} / \colorbox{tabfirst}{179} \\
 & Room & \colorbox{tabfirst}{32.30} / \colorbox{tabfirst}{0.928} / \colorbox{tabsecond}{0.196} / \colorbox{tabfirst}{139} & 31.62 / 0.918 / 0.221 / \colorbox{tabthird}{82} & \colorbox{tabsecond}{32.09} / \colorbox{tabthird}{0.926} / \colorbox{tabthird}{0.208} / 79 & \colorbox{tabthird}{31.71} / \colorbox{tabfirst}{0.928} / \colorbox{tabfirst}{0.191} / \colorbox{tabsecond}{111} \\
 & Counter & \colorbox{tabthird}{29.21} / \colorbox{tabthird}{0.911} / \colorbox{tabsecond}{0.181} / \colorbox{tabfirst}{107} & 29.11 / 0.907 / 0.201 / \colorbox{tabthird}{75} & \colorbox{tabsecond}{29.24} / \colorbox{tabsecond}{0.913} / \colorbox{tabthird}{0.191} / 61 & \colorbox{tabfirst}{29.27} / \colorbox{tabfirst}{0.917} / \colorbox{tabfirst}{0.171} / \colorbox{tabsecond}{94} \\
 & Kitchen & \colorbox{tabsecond}{31.66} / \colorbox{tabsecond}{0.932} / \colorbox{tabsecond}{0.117} / \colorbox{tabfirst}{118} & 31.18 / 0.924 / 0.131 / \colorbox{tabthird}{78} & \colorbox{tabthird}{31.44} / \colorbox{tabthird}{0.927} / \colorbox{tabthird}{0.130} / 72 & \colorbox{tabfirst}{31.83} / \colorbox{tabfirst}{0.933} / \colorbox{tabfirst}{0.114} / \colorbox{tabsecond}{101} \\
 & Bonsai & 32.33 / \colorbox{tabsecond}{0.945} / \colorbox{tabsecond}{0.183} / \colorbox{tabfirst}{132} & \colorbox{tabthird}{32.34} / 0.941 / 0.205 / \colorbox{tabthird}{95} & \colorbox{tabfirst}{32.46} / \colorbox{tabthird}{0.944} / \colorbox{tabthird}{0.199} / 81 & \colorbox{tabsecond}{32.44} / \colorbox{tabfirst}{0.947} / \colorbox{tabfirst}{0.181} / \colorbox{tabsecond}{115} \\
 & \textbf{Average} & \colorbox{tabfirst}{29.37} / \colorbox{tabfirst}{0.880} / \colorbox{tabfirst}{0.168} / \colorbox{tabfirst}{140} & 28.71 / 0.846 / 0.222 / \colorbox{tabthird}{109} & \colorbox{tabsecond}{29.23} / \colorbox{tabthird}{0.875} / \colorbox{tabthird}{0.189} / 92 & \colorbox{tabthird}{29.18} / \colorbox{tabsecond}{0.878} / \colorbox{tabfirst}{0.168} / \colorbox{tabsecond}{131} \\

\midrule
\multirow{9}{*}{\rotatebox{90}{OMMO \cite{lu2023largescaleoutdoormultimodaldataset}}}
& 01 & 23.73 / \colorbox{tabfirst}{0.711} / \colorbox{tabfirst}{0.315} / \colorbox{tabfirst}{104} & \colorbox{tabfirst}{24.18} / 0.703 / 0.350 / \colorbox{tabthird}{73} & \colorbox{tabfirst}{24.18} / \colorbox{tabsecond}{0.709} / \colorbox{tabthird}{0.341} / 63 & \colorbox{tabthird}{24.02} / \colorbox{tabsecond}{0.709} / \colorbox{tabsecond}{0.321} / \colorbox{tabsecond}{84} \\
 & 03 & \colorbox{tabsecond}{27.39} / \colorbox{tabsecond}{0.896} / \colorbox{tabsecond}{0.167} / \colorbox{tabfirst}{127} & 26.96 / 0.886 / 0.197 / 64 & \colorbox{tabthird}{27.20} / \colorbox{tabthird}{0.890} / \colorbox{tabthird}{0.189} / \colorbox{tabthird}{67} & \colorbox{tabfirst}{27.72} / \colorbox{tabfirst}{0.905} / \colorbox{tabfirst}{0.157} / \colorbox{tabsecond}{85} \\
 & 05 & 28.56 / \colorbox{tabthird}{0.877} / \colorbox{tabsecond}{0.185} / \colorbox{tabfirst}{157} & \colorbox{tabthird}{28.72} / \colorbox{tabthird}{0.877} / 0.227 / \colorbox{tabsecond}{115} & \colorbox{tabsecond}{28.98} / \colorbox{tabsecond}{0.883} / \colorbox{tabthird}{0.217} / 98 & \colorbox{tabfirst}{29.15} / \colorbox{tabfirst}{0.885} / \colorbox{tabfirst}{0.181} / \colorbox{tabthird}{101} \\
 & 06 & \colorbox{tabfirst}{27.87} / \colorbox{tabthird}{0.936} / \colorbox{tabthird}{0.150} / \colorbox{tabfirst}{136} & \colorbox{tabthird}{27.68} / 0.933 / 0.158 / \colorbox{tabsecond}{110} & 27.46 / \colorbox{tabsecond}{0.939} / \colorbox{tabsecond}{0.146} / \colorbox{tabthird}{103} & \colorbox{tabsecond}{27.75} / \colorbox{tabfirst}{0.945} / \colorbox{tabfirst}{0.127} / 102 \\
 & 10 & \colorbox{tabsecond}{31.60} / \colorbox{tabsecond}{0.914} / \colorbox{tabsecond}{0.147} / \colorbox{tabfirst}{170} & \colorbox{tabthird}{31.47} / \colorbox{tabthird}{0.906} / 0.171 / 104 & 30.89 / 0.904 / \colorbox{tabthird}{0.169} / \colorbox{tabsecond}{120} & \colorbox{tabfirst}{31.82} / \colorbox{tabfirst}{0.920} / \colorbox{tabfirst}{0.138} / \colorbox{tabthird}{116} \\
 & 13 & \colorbox{tabfirst}{33.55} / \colorbox{tabfirst}{0.957} / \colorbox{tabfirst}{0.098} / \colorbox{tabfirst}{190} & \colorbox{tabsecond}{33.13} / \colorbox{tabthird}{0.949} / \colorbox{tabthird}{0.129} / 116 & \colorbox{tabthird}{32.83} / 0.945 / 0.131 / \colorbox{tabthird}{124} & \colorbox{tabthird}{32.83} / \colorbox{tabsecond}{0.955} / \colorbox{tabsecond}{0.100} / \colorbox{tabsecond}{125} \\
 & 14 & \colorbox{tabthird}{31.80} / \colorbox{tabsecond}{0.953} / \colorbox{tabsecond}{0.090} / \colorbox{tabfirst}{146} & \colorbox{tabsecond}{31.93} / \colorbox{tabthird}{0.951} / \colorbox{tabthird}{0.097} / 89 & 31.44 / 0.950 / 0.099 / \colorbox{tabsecond}{96} & \colorbox{tabfirst}{32.08} / \colorbox{tabfirst}{0.955} / \colorbox{tabfirst}{0.086} / \colorbox{tabthird}{92} \\
 & 15 & \colorbox{tabthird}{30.99} / \colorbox{tabfirst}{0.950} / \colorbox{tabsecond}{0.081} / \colorbox{tabfirst}{151} & \colorbox{tabsecond}{31.13} / \colorbox{tabthird}{0.943} / \colorbox{tabthird}{0.096} / 100 & 30.42 / 0.942 / 0.097 / \colorbox{tabsecond}{104} & \colorbox{tabfirst}{31.25} / \colorbox{tabfirst}{0.950} / \colorbox{tabfirst}{0.079} / \colorbox{tabsecond}{104} \\
 & \textbf{Average} & \colorbox{tabsecond}{29.44} / \colorbox{tabsecond}{0.899} / \colorbox{tabsecond}{0.154} / \colorbox{tabfirst}{148} & \colorbox{tabthird}{29.40} / 0.894 / 0.178 / 96 & 29.18 / \colorbox{tabthird}{0.895} / \colorbox{tabthird}{0.174} / \colorbox{tabthird}{97} & \colorbox{tabfirst}{29.58} / \colorbox{tabfirst}{0.903} / \colorbox{tabfirst}{0.149} / \colorbox{tabsecond}{101} \\

\midrule
\multirow{3}{*}{\fontsize{6}{7}\selectfont  \rotatebox{90}{\makecell{Tanks \& \\ Temples \cite{Knapitsch2017}}}}
 & Truck & \colorbox{tabsecond}{25.34} / \colorbox{tabfirst}{0.864} / \colorbox{tabfirst}{0.181} / \colorbox{tabfirst}{97} & \colorbox{tabthird}{25.06} / 0.851 / 0.234 / \colorbox{tabthird}{70} & \colorbox{tabfirst}{25.57} / \colorbox{tabsecond}{0.862} / \colorbox{tabthird}{0.217} / 45 & 25.02 / \colorbox{tabsecond}{0.862} / \colorbox{tabfirst}{0.181} / \colorbox{tabsecond}{82} \\
 & Train & 22.05 / \colorbox{tabthird}{0.805} / \colorbox{tabsecond}{0.232} / \colorbox{tabfirst}{109} & \colorbox{tabthird}{22.13} / 0.801 / 0.255 / \colorbox{tabthird}{51} & \colorbox{tabfirst}{22.29} / \colorbox{tabsecond}{0.811} / \colorbox{tabthird}{0.249} / 47 & \colorbox{tabfirst}{22.29} / \colorbox{tabfirst}{0.823} / \colorbox{tabfirst}{0.220} / \colorbox{tabsecond}{78} \\
 & \textbf{Average} & \colorbox{tabsecond}{23.70} / \colorbox{tabthird}{0.835} / \colorbox{tabsecond}{0.207} / \colorbox{tabfirst}{103} & 23.60 / 0.826 / 0.245 / \colorbox{tabthird}{60} & \colorbox{tabfirst}{23.93} / \colorbox{tabsecond}{0.837} / \colorbox{tabthird}{0.233} / 46 & \colorbox{tabthird}{23.66} / \colorbox{tabfirst}{0.843} / \colorbox{tabfirst}{0.201} / \colorbox{tabsecond}{80} \\

\midrule
\multirow{3}{*}{\fontsize{8}{8}\selectfont \rotatebox{90}{\makecell{Deep \\ Blending \cite{DeepBlending2018}}}} 
 & Dr Johnson & \colorbox{tabsecond}{28.98} / \colorbox{tabsecond}{0.886} / \colorbox{tabsecond}{0.282} / \colorbox{tabfirst}{194} & \colorbox{tabthird}{28.85} / \colorbox{tabthird}{0.884} / 0.293 / 106 & \colorbox{tabfirst}{29.06} / \colorbox{tabthird}{0.884} / \colorbox{tabthird}{0.291} / \colorbox{tabthird}{164} & 28.75 / \colorbox{tabfirst}{0.890} / \colorbox{tabfirst}{0.277} / \colorbox{tabsecond}{180} \\
 & Playroom & \colorbox{tabfirst}{30.39} / \colorbox{tabfirst}{0.891} / \colorbox{tabfirst}{0.264} / \colorbox{tabfirst}{200} & 29.02 / 0.877 / 0.290 / 112 & \colorbox{tabsecond}{30.28} / \colorbox{tabsecond}{0.890} / \colorbox{tabthird}{0.284} / \colorbox{tabthird}{173} & \colorbox{tabthird}{30.11} / \colorbox{tabthird}{0.889} / \colorbox{tabsecond}{0.265} / \colorbox{tabsecond}{182} \\
 & \textbf{Average} & \colorbox{tabfirst}{29.69} / \colorbox{tabsecond}{0.889} / \colorbox{tabsecond}{0.273} / \colorbox{tabfirst}{197} & 28.94 / 0.881 / 0.292 / 109 & \colorbox{tabsecond}{29.67} / \colorbox{tabthird}{0.887} / \colorbox{tabthird}{0.288} / \colorbox{tabthird}{168} & \colorbox{tabthird}{29.43} / \colorbox{tabfirst}{0.890} / \colorbox{tabfirst}{0.271} / \colorbox{tabsecond}{181} \\

\bottomrule
\end{tabular}}
\caption{\textbf{Detailed results} on a selection of datasets and methods with the number of Gaussians limited to 1M. Each field contains PSNR, SSIM, LPIPS and FPS respectively. We highlight the \colorbox{tabfirst}{best}, \colorbox{tabsecond}{second best} and \colorbox{tabthird}{third best} results among all. Slight discrepancies from the main table are due to rounding.}
\label{table:expanded_all_results_1M}
\end{table*}

\begin{table*}
\centering
\resizebox{0.95\linewidth}{!}{%
\begin{tabular}{cp{1.5cm}|ccccc}
\toprule
& \textit{Scene} & Ours & 3DGS (SfM) & MCMC (SfM) & EDGS \\ 
\midrule

\multirow{8}{*}{\rotatebox{90}{Mip-NeRF360 \cite{barron2022mipnerf360}}} 
& Bicycle & \colorbox{tabthird}{25.43} / \colorbox{tabthird}{0.781} / \colorbox{tabsecond}{0.175} / \colorbox{tabfirst}{99} & 24.85 / 0.729 / 0.267 / \colorbox{tabthird}{79} & \colorbox{tabfirst}{25.56} / \colorbox{tabsecond}{0.786} / \colorbox{tabthird}{0.205} / 71 & \colorbox{tabsecond}{25.48} / \colorbox{tabfirst}{0.792} / \colorbox{tabfirst}{0.168} / \colorbox{tabsecond}{93} \\
 & Garden & \colorbox{tabfirst}{27.59} / \colorbox{tabfirst}{0.870} / \colorbox{tabfirst}{0.097} / \colorbox{tabfirst}{97} & 27.22 / 0.852 / 0.129 / \colorbox{tabthird}{84} & \colorbox{tabthird}{27.33} / \colorbox{tabthird}{0.862} / \colorbox{tabthird}{0.122} / 82 & \colorbox{tabsecond}{27.46} / \colorbox{tabfirst}{0.870} / \colorbox{tabfirst}{0.097} / \colorbox{tabsecond}{89} \\
 & Stump & \colorbox{tabsecond}{27.03} / \colorbox{tabsecond}{0.796} / \colorbox{tabfirst}{0.177} / \colorbox{tabfirst}{100} & 26.18 / 0.749 / 0.255 / \colorbox{tabfirst}{100} & \colorbox{tabfirst}{27.39} / \colorbox{tabfirst}{0.808} / \colorbox{tabthird}{0.189} / 69 & \colorbox{tabthird}{26.79} / \colorbox{tabthird}{0.788} / \colorbox{tabsecond}{0.186} / \colorbox{tabfirst}{100} \\
 & Room & \colorbox{tabfirst}{32.32} / \colorbox{tabsecond}{0.929} / \colorbox{tabsecond}{0.189} / \colorbox{tabfirst}{92} & 31.80 / 0.919 / 0.218 / \colorbox{tabthird}{63} & \colorbox{tabsecond}{32.14} / \colorbox{tabsecond}{0.929} / \colorbox{tabthird}{0.199} / 53 & \colorbox{tabthird}{31.89} / \colorbox{tabfirst}{0.930} / \colorbox{tabfirst}{0.184} / \colorbox{tabsecond}{81} \\
 & Counter & \colorbox{tabthird}{29.23} / \colorbox{tabthird}{0.912} / \colorbox{tabsecond}{0.175} / \colorbox{tabfirst}{70} & 29.07 / 0.907 / 0.201 / \colorbox{tabsecond}{68} & \colorbox{tabfirst}{29.41} / \colorbox{tabsecond}{0.917} / \colorbox{tabthird}{0.181} / 42 & \colorbox{tabsecond}{29.37} / \colorbox{tabfirst}{0.918} / \colorbox{tabfirst}{0.164} / \colorbox{tabthird}{65} \\
 & Kitchen & \colorbox{tabthird}{31.80} / \colorbox{tabsecond}{0.934} / \colorbox{tabsecond}{0.114} / \colorbox{tabfirst}{72} & 31.60 / 0.927 / 0.126 / \colorbox{tabthird}{54} & \colorbox{tabsecond}{32.00} / \colorbox{tabthird}{0.931} / \colorbox{tabthird}{0.122} / 47 & \colorbox{tabfirst}{32.06} / \colorbox{tabfirst}{0.935} / \colorbox{tabfirst}{0.111} / \colorbox{tabsecond}{65} \\
 & Bonsai & \colorbox{tabsecond}{32.63} / \colorbox{tabfirst}{0.948} / \colorbox{tabsecond}{0.177} / \colorbox{tabsecond}{84} & 32.34 / 0.941 / 0.204 / \colorbox{tabsecond}{84} & \colorbox{tabfirst}{32.80} / \colorbox{tabfirst}{0.948} / \colorbox{tabthird}{0.189} / 53 & \colorbox{tabthird}{32.62} / \colorbox{tabthird}{0.947} / \colorbox{tabfirst}{0.175} / \colorbox{tabfirst}{85} \\
 & \textbf{Average} & \colorbox{tabsecond}{29.43} / \colorbox{tabthird}{0.881} / \colorbox{tabsecond}{0.158} / \colorbox{tabfirst}{88} & 29.01 / 0.861 / 0.200 / \colorbox{tabthird}{76} & \colorbox{tabfirst}{29.52} / \colorbox{tabfirst}{0.883} / \colorbox{tabthird}{0.172} / 60 & \colorbox{tabthird}{29.38} / \colorbox{tabfirst}{0.883} / \colorbox{tabfirst}{0.155} / \colorbox{tabsecond}{83} \\

\midrule
\multirow{9}{*}{\rotatebox{90}{OMMO \cite{lu2023largescaleoutdoormultimodaldataset}}}
 & 01 & 23.92 / \colorbox{tabsecond}{0.725} / \colorbox{tabfirst}{0.283} / \colorbox{tabfirst}{73} & \colorbox{tabfirst}{24.51} / 0.721 / 0.321 / \colorbox{tabthird}{49} & \colorbox{tabfirst}{24.51} / \colorbox{tabfirst}{0.731} / \colorbox{tabthird}{0.303} / 45 & \colorbox{tabthird}{24.24} / \colorbox{tabsecond}{0.725} / \colorbox{tabsecond}{0.285} / \colorbox{tabsecond}{64} \\
 & 03 & \colorbox{tabsecond}{27.73} / \colorbox{tabsecond}{0.902} / \colorbox{tabsecond}{0.155} / \colorbox{tabsecond}{70} & 27.05 / 0.888 / 0.193 / \colorbox{tabthird}{47} & \colorbox{tabsecond}{27.73} / \colorbox{tabthird}{0.900} / \colorbox{tabthird}{0.171} / 45 & \colorbox{tabfirst}{27.77} / \colorbox{tabfirst}{0.909} / \colorbox{tabfirst}{0.149} / \colorbox{tabfirst}{82} \\
 & 05 & 28.65 / \colorbox{tabthird}{0.877} / \colorbox{tabsecond}{0.176} / \colorbox{tabsecond}{91} & \colorbox{tabthird}{28.69} / \colorbox{tabthird}{0.877} / 0.227 / \colorbox{tabfirst}{112} & \colorbox{tabsecond}{29.20} / \colorbox{tabfirst}{0.887} / \colorbox{tabthird}{0.201} / 67 & \colorbox{tabfirst}{29.31} / \colorbox{tabfirst}{0.887} / \colorbox{tabfirst}{0.169} / \colorbox{tabthird}{90} \\
 & 06 & \colorbox{tabfirst}{27.93} / \colorbox{tabthird}{0.937} / \colorbox{tabthird}{0.143} / \colorbox{tabthird}{79} & \colorbox{tabthird}{27.67} / 0.933 / 0.157 / \colorbox{tabfirst}{109} & 27.58 / \colorbox{tabsecond}{0.942} / \colorbox{tabsecond}{0.138} / 63 & \colorbox{tabfirst}{27.93} / \colorbox{tabfirst}{0.947} / \colorbox{tabfirst}{0.123} / \colorbox{tabsecond}{97} \\
 & 10 & \colorbox{tabsecond}{31.81} / \colorbox{tabsecond}{0.920} / \colorbox{tabsecond}{0.135} / \colorbox{tabfirst}{104} & \colorbox{tabthird}{31.77} / \colorbox{tabthird}{0.915} / 0.153 / 70 & 31.59 / 0.913 / \colorbox{tabthird}{0.151} / \colorbox{tabthird}{80} & \colorbox{tabfirst}{32.26} / \colorbox{tabfirst}{0.928} / \colorbox{tabfirst}{0.123} / \colorbox{tabsecond}{89} \\
 & 13 & \colorbox{tabfirst}{33.88} / \colorbox{tabfirst}{0.961} / \colorbox{tabsecond}{0.088} / \colorbox{tabfirst}{112} & \colorbox{tabsecond}{33.79} / \colorbox{tabthird}{0.957} / 0.112 / 73 & 33.30 / 0.953 / \colorbox{tabthird}{0.111} / \colorbox{tabthird}{80} & \colorbox{tabthird}{33.55} / \colorbox{tabfirst}{0.961} / \colorbox{tabfirst}{0.087} / \colorbox{tabsecond}{90} \\
 & 14 & \colorbox{tabfirst}{32.02} / \colorbox{tabsecond}{0.955} / \colorbox{tabsecond}{0.086} / \colorbox{tabsecond}{83} & \colorbox{tabsecond}{31.95} / 0.951 / 0.096 / \colorbox{tabthird}{82} & 31.75 / \colorbox{tabthird}{0.953} / \colorbox{tabthird}{0.093} / 62 & \colorbox{tabthird}{31.76} / \colorbox{tabfirst}{0.956} / \colorbox{tabfirst}{0.084} / \colorbox{tabfirst}{86} \\
 & 15 & \colorbox{tabthird}{31.25} / \colorbox{tabsecond}{0.953} / \colorbox{tabsecond}{0.075} / \colorbox{tabsecond}{86} & \colorbox{tabsecond}{31.35} / \colorbox{tabthird}{0.947} / 0.090 / \colorbox{tabthird}{83} & 30.80 / \colorbox{tabthird}{0.947} / \colorbox{tabthird}{0.086} / 67 & \colorbox{tabfirst}{31.63} / \colorbox{tabfirst}{0.954} / \colorbox{tabfirst}{0.072} / \colorbox{tabfirst}{87} \\
 & \textbf{Average} & \colorbox{tabsecond}{29.65} / \colorbox{tabsecond}{0.904} / \colorbox{tabsecond}{0.143} / \colorbox{tabfirst}{87} & \colorbox{tabthird}{29.60} / 0.899 / 0.169 / \colorbox{tabthird}{78} & 29.56 / \colorbox{tabthird}{0.903} / \colorbox{tabthird}{0.157} / 64 & \colorbox{tabfirst}{29.81} / \colorbox{tabfirst}{0.908} / \colorbox{tabfirst}{0.137} / \colorbox{tabsecond}{86} \\
\midrule

\multirow{3}{*}{\fontsize{6}{7}\selectfont  \rotatebox{90}{\makecell{Tanks \& \\ Temples \cite{Knapitsch2017}}}}
& Truck & \colorbox{tabsecond}{25.43} / \colorbox{tabsecond}{0.865} / \colorbox{tabsecond}{0.167} / \colorbox{tabfirst}{69} & \colorbox{tabthird}{25.39} / 0.859 / 0.217 / \colorbox{tabthird}{53} & \colorbox{tabfirst}{25.86} / \colorbox{tabfirst}{0.869} / \colorbox{tabthird}{0.193} / 38 & 25.09 / \colorbox{tabsecond}{0.865} / \colorbox{tabfirst}{0.162} / \colorbox{tabsecond}{55} \\
 & Train & \colorbox{tabthird}{21.95} / \colorbox{tabthird}{0.809} / \colorbox{tabsecond}{0.217} / \colorbox{tabfirst}{77} & \colorbox{tabsecond}{22.29} / 0.802 / 0.253 / \colorbox{tabthird}{55} & \colorbox{tabfirst}{22.35} / \colorbox{tabsecond}{0.822} / \colorbox{tabthird}{0.228} / 40 & 21.84 / \colorbox{tabfirst}{0.827} / \colorbox{tabfirst}{0.204} / \colorbox{tabsecond}{58} \\
 & \textbf{Average} & \colorbox{tabthird}{23.69} / \colorbox{tabthird}{0.837} / \colorbox{tabsecond}{0.192} / \colorbox{tabfirst}{73} & \colorbox{tabsecond}{23.84} / 0.831 / 0.235 / \colorbox{tabthird}{54} & \colorbox{tabfirst}{24.11} / \colorbox{tabfirst}{0.846} / \colorbox{tabthird}{0.211} / 39 & 23.47 / \colorbox{tabfirst}{0.846} / \colorbox{tabfirst}{0.183} / \colorbox{tabsecond}{56} \\
 
\midrule
\multirow{3}{*}{\fontsize{8}{8}\selectfont \rotatebox{90}{\makecell{Deep \\ Blending \cite{DeepBlending2018}}}} 
 & Dr Johnson & \colorbox{tabfirst}{29.01} / \colorbox{tabthird}{0.884} / \colorbox{tabsecond}{0.274} / \colorbox{tabfirst}{127} & \colorbox{tabthird}{28.75} / \colorbox{tabsecond}{0.886} / \colorbox{tabthird}{0.282} / 77 & \colorbox{tabsecond}{28.83} / 0.882 / 0.285 / \colorbox{tabthird}{104} & 28.65 / \colorbox{tabfirst}{0.887} / \colorbox{tabfirst}{0.268} / \colorbox{tabsecond}{113} \\
 & Playroom & \colorbox{tabfirst}{30.50} / \colorbox{tabsecond}{0.887} / \colorbox{tabfirst}{0.247} / \colorbox{tabfirst}{127} & 29.32 / 0.880 / 0.279 / 83 & \colorbox{tabsecond}{30.22} / \colorbox{tabfirst}{0.890} / \colorbox{tabthird}{0.272} / \colorbox{tabthird}{105} & \colorbox{tabthird}{30.11} / \colorbox{tabthird}{0.886} / \colorbox{tabsecond}{0.248} / \colorbox{tabsecond}{112} \\
 & \textbf{Average} & \colorbox{tabfirst}{29.76} / \colorbox{tabsecond}{0.886} / \colorbox{tabsecond}{0.261} / \colorbox{tabfirst}{127} & 29.04 / 0.883 / 0.281 / 80 & \colorbox{tabsecond}{29.53} / \colorbox{tabsecond}{0.886} / \colorbox{tabthird}{0.279} / \colorbox{tabthird}{104} & \colorbox{tabthird}{29.38} / \colorbox{tabfirst}{0.887} / \colorbox{tabfirst}{0.258} / \colorbox{tabsecond}{112} \\
\bottomrule
\end{tabular}}
\caption{\textbf{Detailed results} on a selection of datasets and methods with the number of Gaussians limited to 2M. Each field contains PSNR, SSIM, LPIPS and FPS respectively. We highlight the \colorbox{tabfirst}{best}, \colorbox{tabsecond}{second best} and \colorbox{tabthird}{third best} results among all. Slight discrepancies from the main table are due to rounding.}
\label{table:expanded_all_results_2M}
\end{table*}


\end{document}